\documentclass[10pt,journal,compsoc]{IEEEtran}
\usepackage{footmisc}
\usepackage{cite}
\usepackage{amsmath}
\usepackage{epsfig}
\usepackage{graphicx}
\usepackage{multirow}
\usepackage{multicol}
\usepackage{amssymb}
\usepackage{rotating}
\usepackage{mathtools}
\usepackage[table]{xcolor}
\usepackage{ragged2e}
\usepackage{tikz}
\usepackage{hhline}
\usepackage{booktabs}
\usepackage{bbding}
\usepackage{wasysym}
\usepackage{threeparttable}
\usepackage{mathrsfs}
\usepackage{dcolumn}
\usepackage{float}
\usepackage{widetext}
\usepackage{wrapfig}
\usepackage{color}
\usepackage{pifont}
\usepackage{booktabs}
\usepackage{microtype}
\usepackage{soul}
\usepackage{color, xcolor}
\usepackage{cite}
\usepackage{pifont}
\usepackage{enumitem}
\usepackage{footmisc}
\usepackage[pagebackref=false,breaklinks=true,colorlinks,bookmarks=false]{hyperref}
\usepackage[capitalize]{cleveref}

\definecolor{lightblue}{RGB}{54,46,204}
\definecolor{ouryellow}{RGB}{149,146,0}
\definecolor{O1yellow}{RGB}{255,199,143}
\definecolor{A1yellow}{RGB}{255,153,51}
\definecolor{O1blue}{RGB}{83,130,53}
\definecolor{A1blue}{RGB}{56,86,34}
\DeclareGraphicsExtensions{.pdf,.jpg,.png}

\crefname{section}{Sec.}{Secs.}
\Crefname{section}{Section}{Sections}
\Crefname{table}{Table}{Tables}
\crefname{table}{Tab.}{Tabs.}

\def\ourData{PanopticVideo-300}
\def\ourApproach{WinDB}
\def\OurNet{FishNet}

\begin{document}
\title{WinDB: HMD-free and Distortion-free Panoptic Video Fixation Learning}
\author{Guotao~Wang,~
        Chenglizhao~Chen*,~
        Aimin~Hao,~
        Hong~Qin,~
        and~Deng-Ping~Fan\\

        %\emph{Codes \& Data \url{https://github.com/guotaowang/STANet}}.

\IEEEcompsocitemizethanks{
\IEEEcompsocthanksitem
Guotao~Wang is with the State Key Laboratory of Virtual Reality Technology and Systems, Beihang University, China.
(E-mail: qduwgt@163.com)
\IEEEcompsocthanksitem
Chenglizhao Chen is with the College of Computer Science and Technology, China University of Petroleum, China.
(E-mail: cclz123@163.com).
\IEEEcompsocthanksitem
Aimin Hao is with the State Key Laboratory of Virtual Reality Technology and Systems, Beihang University,
Research Unit of Virtual Human and Virtual Surgery, Chinese Academy of Medical Sciences, and
with Pengcheng Laboratory, China.
(Email: ham@buaa.edu.cn)
\IEEEcompsocthanksitem
Hong Qin is with the Computer Science Department, Stony Brook University, USA.
(E-mail: qin@cs.stonybrook.edu).
\IEEEcompsocthanksitem
Deng-Ping Fan is with ETH Zurich, Zurich, Switzerland
(Email: dengpfan@gmail.com)
\IEEEcompsocthanksitem
Corresponding author: Chenglizhao Chen.}}
\markboth{IEEE TRANSACTIONS ON PATTERN ANALYSIS AND MACHINE INTELLIGENCE}%
{Wang \MakeLowercase{\textit{\textit{et al.}}}:WinDB: HMD-free and Distortion-free Panoptic
Video Fixation Learning}
\IEEEtitleabstractindextext{%
\begin{abstract}
To date, the widely adopted way to perform fixation collection in panoptic video is based on a head-mounted display (HMD), where users' fixations are collected while wearing an HMD to explore the given panoptic scene freely. However, this widely-used data collection method is insufficient for training deep models to accurately predict which regions in a given panoptic are most important when it contains intermittent salient events. The main reason is that there always exist ``blind zooms'' when using HMD to collect fixations since the users cannot keep spinning their heads to explore the entire panoptic scene all the time. Consequently, the collected fixations tend to be trapped in some local views, leaving the remaining areas to be the ``blind zooms''. Therefore, fixation data collected using HMD-based methods that accumulate local views cannot accurately represent the overall global importance --- the main purpose of fixations --- of complex panoptic scenes. To conquer, this paper introduces the auxiliary \textbf{win}dow with a \textbf{d}ynamic \textbf{b}lurring (\textbf{\ourApproach}) fixation collection approach for panoptic video, which doesn't need HMD and is able to well reflect the regional-wise importance degree. Using our WinDB approach, we have released a new \textbf{\ourData} dataset, containing 300 panoptic clips covering over 225 categories. Specifically, since using \ourApproach~to collect fixations is blind zoom free, there exists frequent and intensive ``fixation shifting" --- a very special phenomenon that has long been overlooked by the previous research --- in our new set. Thus, we present an effective \textbf{fi}xation \textbf{sh}ifting \textbf{net}work (\textbf{\OurNet}) to conquer it. All these new fixation collection tool, dataset, and network could be very potential to open a new age for fixation-related research and applications in $360^{\rm o}$ environments.
\end{abstract}
}

\maketitle
\IEEEdisplaynontitleabstractindextext
\IEEEpeerreviewmaketitle
\ifCLASSOPTIONcompsoc

\vspace{-16pt}
\section{Introduction and Motivation}
\label{sec:introduction}
Given a panoptic scene, the primary goal of panoptic fixation prediction is to perceive the regional-wise importance, which reflects the varying degrees of attention that different areas of the entire scene. This objective enables the rapid localization of the most ``important regions" in the scene.
Generally, the localized important regions have large variants of applications.
As shown in Fig.~\ref{fig:APPNew}, it can drive dynamic regional-wise compression ratio and rendering quality~\cite{hu2017deep}.
When wearing a \textbf{h}ead-\textbf{m}ounted \textbf{d}isplay (HMD), we can use the localized regions to strike a better balance between computation cost and rendering quality and thus have better graphics and smoother performance, ultimately increasing immersion~\cite{sitzmann2018saliency}.

\begin{figure}[!t]
\centering
\includegraphics[width=1\linewidth]{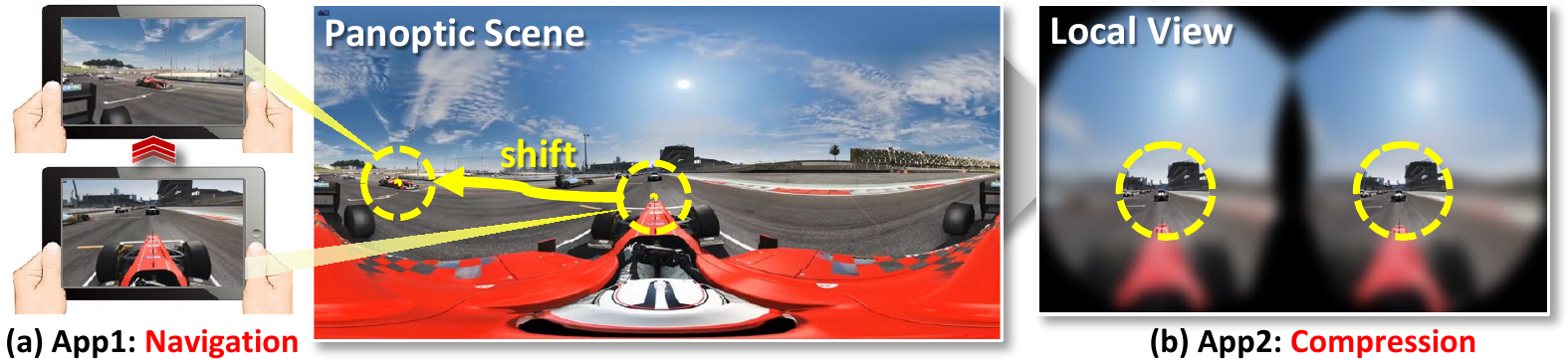}%{OurHMD.pdf}
\vspace{-0.4cm}
\caption{Panoptic fixations that can well reflect the scene's regional-wise importance can be applied to multiple applications. Here we illustrate the two most representative ones: (a) panoptic video navigation~\cite{hu2017deep}, which facilitates the localizing of intermittent ``salient events" in blind zoom, and (b) panoptic video compression~\cite{sitzmann2018saliency}.}
\label{fig:APPNew}
\vspace{-17pt}
\end{figure}

\begin{figure*}[!t]
\centering
\includegraphics[width=1\linewidth]{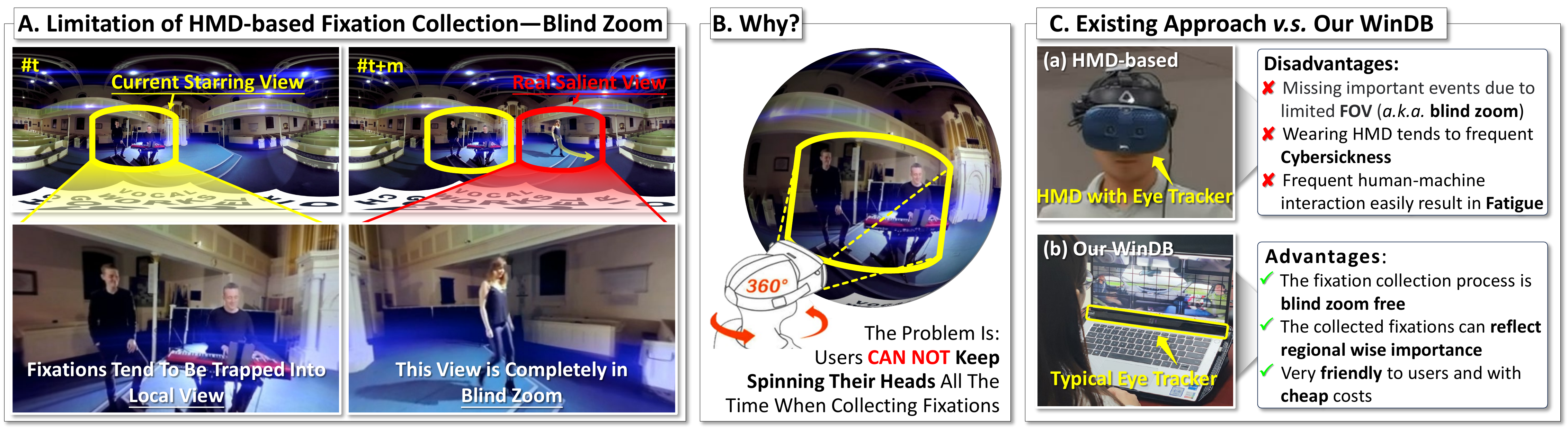}
\vspace{-25pt}
\caption{
The existing HMD-based fixation collection method
\cite{zhang2018saliency,zhang2022pav,xu2018gaze,xu2018predicting} for panoptic data has a critical limitation --- \textbf{blind zoom},
results in the collected fixations being insufficient to train deep models to accurately predict which regions in a given panoptic are most important (A).
The reason causing this drawback has been shown in (B), where users wearing an HMD tend to become ``retard'' after the early scene exploring stage, resulting in missing important events that occurred in blind zooms.
(C) summarizes the advantages of our novel WinDB against the existing HMD-based fixation collection, where advantages and disadvantages are respectively remarked as~\textcolor{green}{\ding{51}} and~\textcolor{red}{\ding{55}}. See Sec.~\ref{sec:introduction} for details.
}
\label{fig:motivation}
\vspace{-9pt}
\end{figure*}

Different from the conventional 2D fixation prediction~\cite{Aberman_2022_CVPR,Jiang_2022_CVPR,wang2021semantic,tsiami2020stavis}, which has received extensive research attention, the panoptic fixation prediction~\cite{djilali2021rethinking,zhu2021viewing,ECCV2022,hu2017deep,nguyen2018your,zhu2019prediction} is currently in its infancy.
The major problem causing such slow progress is the shortage of large-scale datasets~\cite{cohen2018spherical,jiang2019spherical,su2019kernel,djilali2021rethinking}, because collecting human-eye fixations in panoptic scene is much more challenging than that in conventional 2D scene~\cite{sitzmann2018saliency,TCSVT1,li2022spherical,lee2020spherephd,esteves2018learning,weiler2018learning}.
Also, the panoptic fixation prediction is much more complex than the conventional fixation prediction in 2D images, where 2D data only has one fixed view, yet, panoptic data allows users to explore $\rm 360^{\rm o}$ panoptic video freely~\cite{qiao2020viewport,su2017learning,lee2018memory,su2017making}.
Thus, our panoptic fixation prediction research community is currently facing a dilemma --- using tiny small-scale training data\footnote{There are only 208 video clips most commonly used in our research community, and most of them are simple scenes.} to beat a complex problem.

To date, the HMD-based human-eye fixation collection~\cite{zhang2018saliency, zhang2022pav, xu2018gaze, xu2018predicting} is the most popular approach, where users wear an HMD to explore the given panoptic scene freely~\cite{xu2020viewport,Tome_2019_ICCV,li2018bridge}, and, at the same time, fixations are collected.
Albeit its broad application,
the HMD-based fixation collection~\cite{zhang2018saliency,zhang2022pav,xu2018gaze,xu2018predicting,rondon2021track} has two problems, and one of them is extremely critical. \textbf{First}, there always exist ``blind zooms'' when using HMD to collect fixations since the users cannot keep spinning their heads to explore the entire panoptic scene all the time.
The blind zoom problem makes the HMD-based fixations inconsistent with the real regional-wise importance degrees in the given panoptic scene.
Thus, a salient event occurring in ``blind zoom'' might receive zero fixation.
To facilitate a better understanding, we have further demonstrated a vivid example in Fig.~\ref{fig:motivation}-A and B.
\textbf{Second}, the HMD-based fixation collection is relatively ``expensive''\footnote{Our concern is not about the high cost of the equipment itself but rather the expensiveness of the annotation process.}, and users usually feel very uncomfortable (\emph{e.g.}, cybersickness~\cite{kim2018vrsa}) when wearing an HMD to explore the panoptic scene.
In a word, the methodology of HMD-based fixation collection is ill-posed, and its annotation process is extremely expensive.

Besides the above-mentioned HMD-based fixation prediction approach, it is worth mentioning that plain \textbf{E}qui-\textbf{R}ectangular \textbf{P}rojection (ERP)~\cite{Sun_2021_CVPR,Xu_2021_CVPR,yang2018automatic,Zhang_2019_ICCV,su2016pano2vid,yu2018deep}, which projects a panoptic scene, a typical spherical data, to 2D formulation, is also generally not suitable to serve as the platform for human-eye fixation collection.
The primary reason is that the ERP-based 2D form suffers severe visual distortions~\cite{zhuang2023spdet, taneja2015geometric,jin2020geometric,Pintore_2021_CVPR}, especially for those regions around the poles (see the top row of Fig.~\ref{fig:motivation}-A).
As a result, the distorted regions --- being irregular to their surroundings, may occasionally draw users' fixations even if they are not salient~\cite{song20233d,cheng2018cube,xiong2018snap,ma2020stage}.
In short, compared with the HMD-based method, ERP is a double sword; its advantage (\emph{e.g.}, no blind zoom) cannot outweigh its disadvantage (\emph{e.g.}, severe visual distortions), making it unsuitable in real $360^{\rm o}$ fixation collection.

Given the abovementioned aspects, this paper presents a novel approach, named as auxiliary \textbf{win}dow with a \textbf{d}ynamic \textbf{b}lurring (\textbf{WinDB}), to panoptic fixation collection.
Our WinDB approach has considered all the advantages and disadvantages of both HMD-based and ERP-based methods.
The key idea is to take full advantage of the ERP-based fixation collection (\emph{i.e.}, blind zoom free) and resort to a series of tailored designs to suppress the visual distortions, and we can preview its overview in Fig.~\ref{fig:pipeline}.
Moreover, since our WinDB approach is HMD-free, enabling users to explore a panoptic scene in front of a computer and collect fixations via an eye tracker, which is more comfortable than the HMD-based method (see Fig.~\ref{fig:motivation}-C-b).
The fixations collected by our WinDB approach can well indicate each region's importance degree in the given panoptic scene.

\begin{figure*}[!t]
\centering
\includegraphics[width=1\linewidth]{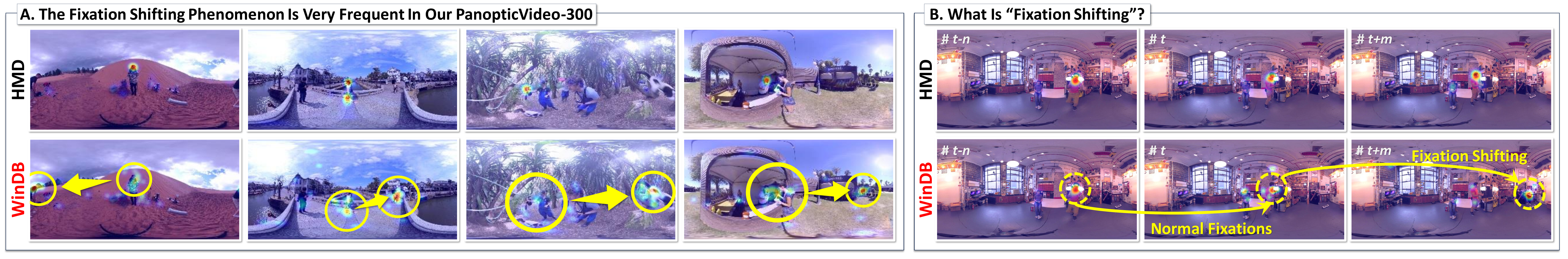}%{OurHMD.pdf}
\vspace{-14pt}
\caption{
Qualitative demonstration of the differences between the datasets collected by our WinDB method and the HMD method (\emph{i.e.}, VR-Eye Tracking~\cite{xu2018gaze}). \textbf{Sub-figure A} illustrates the fixation shifting phenomenon in \ourData~dataset. Since our WinDB method does not have blind zoom, our WinDB is able to capture salient events that are ignored by the HMD method, \emph{i.e.}, the sudden human and animal events within the blind zoom. \textbf{Sub-figure B} shows the phenomenon of ``fixation shifting", in which the fixation shifts from the talking person to the person pushing the door. However, since the HMD method has a blind zoom, it has focused on the person talking.
}
\label{fig:newdemo}
\vspace{-11pt}
\end{figure*}

%As shown in Fig.\ref{fig:OurHMD}-B, in racing scenes, our WinDB method is capable of detecting lateral overtaking maneuvers, enabling the panoptic video navigation to switch to a lateral perspective.
%However, using the HMD approach may lead to missing the exciting moments of overtaking due to blind zoom (see Fig.\ref{fig:OurHMD}-B-a).
%Additionally, our method can be applied for compressing panoptic videos during HMD playback (see Fig.~\ref{fig:OurHMD}-B-b).

In addition, using our proposed WinDB approach, we have constructed a large panoptic video fixation prediction dataset, PanopticVideo-300\footnote{The PanopticVideo-300 is now publicly available at \url{https://github.com/guotaowang/PanopticVideo-300}.}, which is the most challenging dataset in 360$^{\rm o}$ video fixation prediction due to its significant inclusion of scenes with blind zoom.
Since the blind-zoom issue has been nicely addressed, the fixations collected by our proposed WinDB approach can well reflect the regional-wise importance, making PanopticVideo-300 the first comprehensive dataset in our research field.

\emph{W.r.t} our PanopticVideo-300, we have found an interesting phenomenon --- there exists frequent ``fixation shifting''.
Because of the limitation of HMD (\emph{i.e.}, blind zoom), the fixations collected by HMD tend to be generally smooth in the conventional panoptic fixation sets (\emph{e.g.}, VR-EyeTracking~\cite{xu2018gaze} and Wild360~\cite{cheng2018cube}).
As we have stated, HMD-based fixations tend to be trapped in local view due to blind zoom, and thus, these fixations are generally smooth.
In sharp contrast, our WinDB is blind-zoom free; thus, those salient events that were supposed to be neglected by the HMD-based method can now be fully discovered.
As a result, in our PanopticVideo-300, fixations might shift to another long-distance position in a very short period (see Fig.~\ref{fig:newdemo}-A), and we call this phenomenon ``fixation shifting''.
This phenomenon is not bad at all; instead, it can further verify the solidness of our WinDB in collecting fixations.
For a better reading, we have provided a vivid example in Fig.~\ref{fig:newdemo}-B, where the fixations are shifted from the talking person to the man who pushed the door in. The corresponding technical details of WinDB and more in-depth analysis will be provided in Sec.~\ref{Sec:WinDB}.

Further, we face another dilemma, \emph{i.e.}, none of the previous fixation prediction networks can well handle the ``fixation shifting'' phenomenon.
The main reasons for such deficiency are generally two-fold: \textbf{1)} their networks' designs are over emphasized on pursuing the spatiotemporal smoothness towards their predicted fixations, and \textbf{2)} their sensing scope for fixations is basically local, making them theoretically impossible to perceive long-range fixation shifting.
So, this paper also provides a new network (\emph{i.e.}, FishNet) to handle the fixation shifting, whose technical rationale is very inspiring and has the potential to guide future works, and this novel content will be discussed and detailed in Sec.~\ref{Sec:FishNet}.
In summary, the key contributions include the following:

\vspace{-1pt}
\begin{itemize}[leftmargin=*]
	\setlength{\itemsep}{1pt}
	\setlength{\parsep}{-2pt}
	\setlength{\parskip}{0pt}
	\setlength{\leftmargin}{-15pt}
	\vspace{-4pt}
\vspace{2pt} % human-eye
\item We introduce a \textbf{new fixation collection approach (\textbf{\ourApproach})} for panoptic data, which is the first one that has truly conquered the HMD-induced blind zoom limitation;
\vspace{2pt}
\item Built upon \ourApproach, we have released a new set, the \textbf{Panoptic Video-300}, which is the first solid and most challenging dataset for 360$^{\rm o}$ video fixation prediction, whose fixations can well reflect the regional-wise importance;
\vspace{2pt}
\item As the first attempt, we have devised a novel paradigm of network design to handle the ``fixation shifting'' challenge, and we coin the new network as \textbf{FishNet};
\vspace{2pt}
\item This paper has conducted a whole package work, including a new data collection way (WinDB\footnote{The WinDB tool is publicly available at\! \url{https://github.com/guotaowang/WinDB}.}), a new network (FishNet\footnote{The codes and results are also publicly available at \url{https://github.com/guotaowang/FishNet}.}), and a new dataset (PanopticVideo-300\footnote{The PanopticVideo-300 is now publicly available at \url{https://github.com/guotaowang/PanopticVideo-300}.}), whose methodologies, new findings, in-depth analysis, and conclusion can jointly contribute our research community.
\vspace*{-6pt}
\end{itemize}

\begin{figure*}[!t]
\centering
\includegraphics[width=1\linewidth]{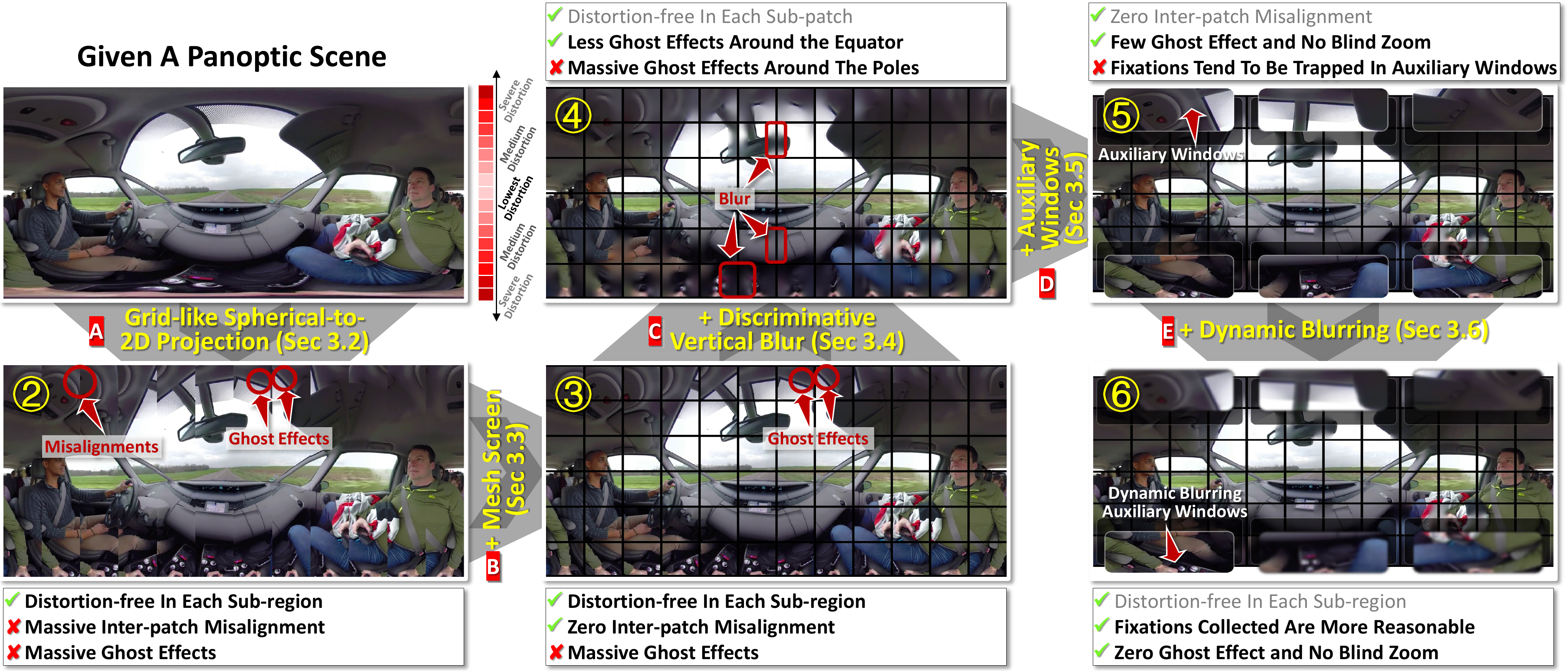}
\vspace{-23pt}
\caption{
The overall pipeline of our new HMD-free fixation collection approach for panoptic data. Compared to the widely-used HMD-based method, our \ourApproach~approach is more economical, comfortable, and reasonable.
The most \textbf{severe} distortion in the given panoptic video scene will be solved by \textbf{D} and \textbf{E}, while the \textbf{medium} distortion will be solved by \textbf{A}, \textbf{B}, and \textbf{C}.
See Sec.~\ref{Sec:WinDB} for details.
}
\label{fig:pipeline}
\vspace{-11pt}
\end{figure*}

\vspace{-9pt}
\section{Related Work}
\subsection{Panoptic Fixation Collection Approach}
\vspace{-2pt}
The fixation collection method of panoptic video fixation collection has two main research branches, \emph{i.e.}, the methods based on HMD~\cite{zhang2018saliency, zhang2022pav, xu2018gaze, xu2018predicting} and the methods based on ERP~\cite{cheng2018cube,rana2019towards}.
HMD-based methods~\cite{zhang2018saliency, zhang2022pav, xu2018gaze, xu2018predicting} require the users to wear the HMD to watch the panoptic video, and the users' fixation is collected.
The advantage of this method is that there is no distortion problem.
However, in this method, users can only observe the content within the local view of HMD.

In contrast, ERP-based methods~\cite{cheng2018cube,rana2019towards} only require the users to watch the video content played on a computer screen and get the users' fixation using eye-tracking~\cite{cheng2018cube} or mouse input~\cite{rana2019towards}.
These methods have the advantage of ensuring users are blind-zoom-free.

\vspace{-11pt}
\subsection{Panoptic Fixation Learning Network}
\vspace{-2pt}
Panoptic video fixation learning aims to rapidly locate important regions,
but it's more challenging for panoptic video compared to traditional 2D methods~\cite{Aberman_2022_CVPR,Jiang_2022_CVPR,wang2021semantic,tsiami2020stavis}.
In 2D video, fixation learning extracts important regions within a fixed view.
In panoptic video~\cite{zhang2018saliency,zhang2022pav,xu2018gaze,xu2018predicting}, all views must be considered.
However, due to bind zoom, HMD-based methods only capture the locally important region, not across the entire panorama.
This leads to inaccuracies in predicting the most important region.
Moreover, there are other issues in panoptic fixation learning networks, \emph{e.g.}, problems with loading pre-trained model parameters, and feature distortion. Panoptic fixation learning networks have three main research branches. Next, we review recent advancements.

\textbf{(1) ERP and CMP bi-stream fusion scheme.}
In this approach~\cite{wang2022bifuse,dahou2021atsal,qiao2020viewport,cheng2018cube}, 2D models were employed to capture both panoptic ERP information and detailed local information from cube maps (CMP). The fusion of these ERP and CMP bi-streams was achieved through different techniques, such as dynamic weighted fusion~\cite{cong2023multi}, bi-projection fusion~\cite{wang2022bifuse}, and multiplication ~\cite{dahou2021atsal}. It's worth noting that this fusion process lacked explicit alignments, essentially making it an additional step for feature embedding.

\textbf{(2) Sphere convolution-based network.}
Spherical convolution~\cite{djilali2021rethinking,zhu2019prediction,li2023spherical,su2021learning,xu2021spherical,zhang2018saliency} involves several key steps. Firstly, it identifies the central position of the convolution kernel on the ERP.
Then, it calculates the kernel's surface area on the sphere using ERP-to-sphere coordinate mapping.
Next, it maps the sphere's coordinates to their corresponding ERP pixels, which are then used as input for the CNN layers to embed semantic features.
Finally, the CNN outputs are re-mapped back to the ERP.
Spherical convolution offers the advantage of performing operations directly on the sphere's surface, eliminating ERP distortion.
However, it presents a challenge as it differs from the standard CNN model, making all available feature backbones for semantic feature embedding unavailable.

\textbf{(3) Transformer-based method.}
This method~\cite{yun2022panoramic} addresses ERP distortion by incorporating local 2D projection layers into a Transformer model~\cite{liu2021visual,wang2021pyramid}.
Each local projection layer employs deformable convolution~\cite{dai2017deformable} to map spherical information to local 2D patches.
The spherical to 2D patch coordinate mapping determines the projection location.
While it significantly reduces ERP distortion, it generates intermediate features from ERP with some visual distortions, limiting the utilization of pre-trained network parameters.

\vspace{-9pt}
\section{Novel 360$^{\rm o}$ Fixation Collection (WinDB)}
\label{Sec:WinDB}
\subsection{WinDB Overview}
\label{sub:methodo}
\vspace{-2pt}
To leverage the advantages of ERP and HMD, particularly in eliminating blind zoom, WinDB employs ERP as the foundational computational platform for fixation collection.
Therefore, the key problem is to eliminate the negative impact of distortion as much as possible.
We have illustrated the WinDB's overview in Fig.~\ref{fig:pipeline}, which consists of five steps.

To address the distortion issue, we employ the step $\textbf{\textcolor[RGB]{255,255,255}{\sethlcolor{red}{\hl{\textbf{\rm A}}}}}$ --- grid-like spherical-to-2D projection (see Fig.~\ref{fig:30deg}-B) --- to equally divide the input ERP image into sub-regions.
Then, each spherical sub-region is projected onto plain 2D image patches.
Thus, after the step $\textbf{\textcolor[RGB]{255,255,255}{\sethlcolor{red}{\hl{\textbf{\rm A}}}}}$, all local sub-patches are distortion-free, see the yellow mark ${{\normalsize{\textcircled{\scriptsize{2}}}\normalsize}}$ in Fig.~\ref{fig:pipeline}.
However, there are noticeable ``ghost effects'' with massive ``inter-patch misalignment'' (see the red arrows in Fig.~\ref{fig:30deg}-B).
Clearly, these artifacts are unacceptable when collecting fixations since they tend to attract human fixation.

To improve, we have devised a very tricky step (the step $\textbf{\textcolor[RGB]{255,255,255}{\sethlcolor{red}{\hl{\textbf{\rm B}}}}}$ of Fig.~\ref{fig:pipeline}) --- we use black narrow lines to block those intersections of neighbored patches, making the entire image to be covered with a black ``mesh screen''.
This step is inspired by a unique mechanism of our human visual system, called Persistence of Vision (POV~\cite{anderson1978myth}, see Fig.~\ref{fig:mesh}), which implies that our human brain can automatically restore those blocked narrow regions.
Thus, using such ``mesh screen tricky'', we can easily handle the ``inter-patch misalignment'' (see the yellow mark ${{\normalsize{\textcircled{\scriptsize{3}}}\normalsize}}$ in Fig.~\ref{fig:pipeline}). More technical details will be provided in Sec.~\ref{sub:MS}.

Next, we shall handle the ``ghost effects''. Here we come up with a simple yet effective solution (the step $\textbf{\textcolor[RGB]{255,255,255}{\sethlcolor{red}{\hl{\textbf{\rm C}}}}}$ of Fig.~\ref{fig:pipeline}), \emph{i.e.}, we perform ``discriminative vertical blur\footnote{Vertical blur: only the left and right sides of a rectangle patch are blurred, because the adopted spherical-to-2D projection method performs equal distance sampling over the sphere's longitude. As a result, the ghost effects only exist between vertically neighbored patches. For a better understanding, please see Fig.~\ref{fig:30deg}-B.}'' on each local patch.
And the blur degree is dynamically controlled according to the patch's position, where patches near the poles are blurred more.
By this step, the ghost effects can be alleviated significantly; see the yellow mark ${{\normalsize{\textcircled{\scriptsize{4}}}\normalsize}}$ in Fig.~\ref{fig:pipeline}.
More technical details will be given in Sec.~\ref{sub:BOR}.

However, noticeable ghost effects still exist in some patches, especially those near the poles (\emph{i.e.}, the top and bottom rows).
Hence, we present a new step (the step $\textbf{\textcolor[RGB]{255,255,255}{\sethlcolor{red}{\hl{\textbf{\rm D}}}}}$ of Fig.~\ref{fig:pipeline}), where ``auxiliary windows'' are placed around the poles.
These windows are essentially large-size sub-patches, which are, of course, distortion-free.
By assigning an appropriate window coverage, the ``ghost effects'' can be solved completely; see the yellow mark ${{\normalsize{\textcircled{\scriptsize{5}}}\normalsize}}$ in Fig.~\ref{fig:pipeline}. More technical details will be given in Sec.~\ref{sub:UAW}.

Further, due to the fact that the auxiliary windows are distortion-free and more informative than local patches, users are naturally more likely to pay attention to them.
We shall avoid this tendency to ensure an objective fixation collection.
So, we adopt the step $\textbf{\textcolor[RGB]{255,255,255}{\sethlcolor{red}{\hl{\textbf{\rm E}}}}}$ --- dynamic blurring, which gradually blurs those auxiliary windows that have received human-eye fixations.
More technical details will be given in Sec.~\ref{sub:ABT}.

All the steps mentioned above consist of our WinDB, which is distortion-free, has no ghost effects, no visual artifacts, no blind zoom, and is user-friendly.

\vspace{-8pt}
\subsection{Grid-like Spherical-to-2D Projection}
\label{sub:GSP}
\vspace{-2pt}
Generally, representing a panoptic image via ERP can retain the image's overall information.
However, as shown in Fig.~\ref{fig:pipeline}-${\normalsize{\textcircled{\scriptsize{1}}}\normalsize}$, there exists ERP visual distortions, where the distortion level becomes more dramatic when closing to the poles.

Inspired by the limited focal range of our HVS\footnote{The HVS has a relatively large range field of view (FOV)~\cite{Yoon_2022_CVPR}, \emph{i.e.}, about $\rm 110^o$, however, the focal range is only about $\rm 25^o$~\cite{AR} (see Fig.~\ref{fig:30deg}-A).}, we present the ``grid-like spherical-to-2D projection (Fig.~\ref{fig:30deg}-B)'' to let the ERP stays ``local distortion-free'' --- each sub-patch has zero distortions.
Although this ``grid-like spherical-to-2D projection'' unavoidably brings noticeable inter-patch misalignments over the vertical direction, we believe that the ERP after the projection can also be regarded as distortion-free if we can achieve ``\emph{\textbf{a precondition}}'' --- let our HVS focus on the sub-patches' inside regions.
Here, we shall leave the question of how to reach this ``precondition'' to the following subsections and begin detailing the proposed ``grid-like spherical-to-2D projection''.

As shown in Fig.~\ref{fig:30deg}-B, suppose a typical ERP frame (with a degree of freedom: horizontal 360$^{\rm o}$ and vertical 180$^{\rm o}$) has been projected on a sphere, we divide the sphere to slices, where the vertical dividing interval is fixed to $\rm 30^{\rm o}$\footnote{The dividing interval is $\rm 30^{\rm o}$ to cover HVS focus range ($\rm 25^{\rm o}$)~\cite{AR}.}. Yet the horizontal interval is dynamic to ensure the obtained spherical slices have the same grid topology as the ERP grid.
As a result, each ``spherical slice (light blue)'' strictly correlates to an ``ERP sub-patch (yellow)'', and they have the same vertical size but have different horizontal sizes, \emph{i.e.}, spherical slices near the poles are squeezed horizontally, and this is the main reason causing the visual distortions in ERP.

\begin{figure}[!t]
\centering
\includegraphics[width=1\linewidth]{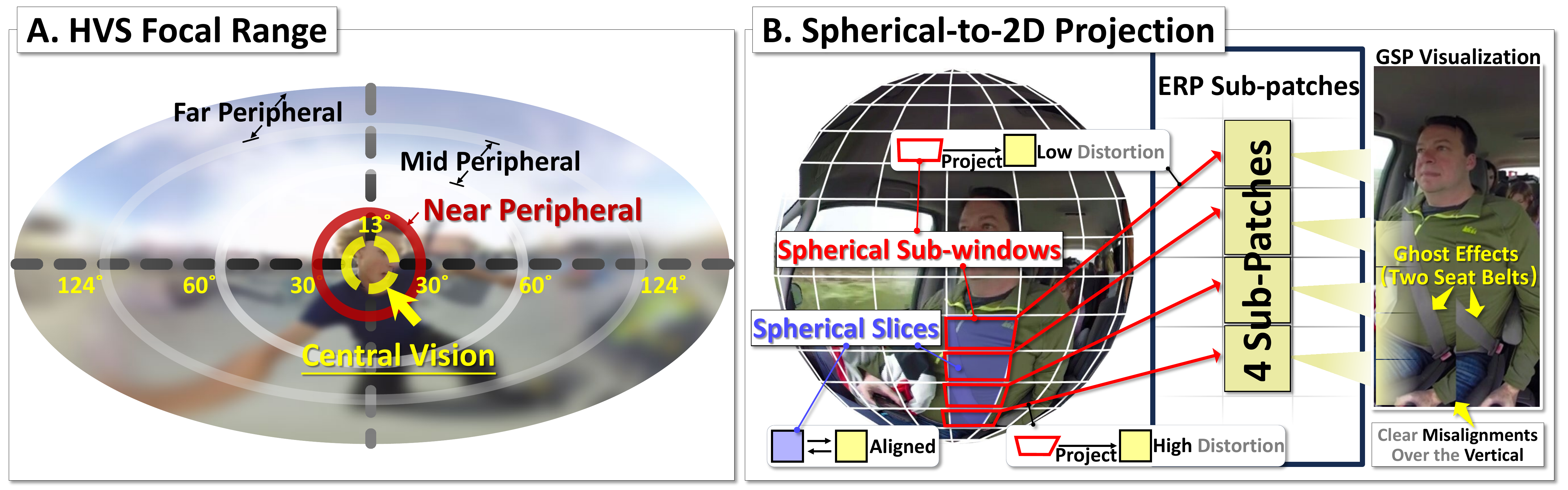}
\vspace{-24pt}
\caption{Detailed illustration of our grid-like spherical-to-2D projection. \textbf{Sub-figure A} is the HVS focal range; the human eye mainly focuses in a range of 30 degrees. \textbf{Sub-figure B} is the technical detail and rationale for adopting the spherical-to-2D projection.
See section~\ref{sub:GSP} for details.
}
\label{fig:30deg}
\vspace{-18pt}
\end{figure}

To make ERP distortion-free, we resort to ``spherical sub-windows'', which have uniform size (\emph{i.e.}, the red box of Fig.~\ref{fig:30deg}-B).
Due to the limited focal range of HVS, this size shall be the same as the largest ``spherical slice'' --- those slices around the sphere's equator.
To be more specific, the sizes of ``spherical slices'' range between (horizontal: 0$\sim$30$^{\rm o}$, vertical: 30$^{\rm o}$), and the sizes of ``spherical sub-windows'' are uniformly assigned to (horizontal: 30$^{\rm o}$, vertical: 30$^{\rm o}$).
Therefore, the ``spherical sub-windows'' are generally larger than ``spherical slices'', especially near the poles.
So, we can easily achieve distortion-free projection using ``spherical sub-windows'', which project each ``spherical sub-window'' to ``ERP sub-patch''.
Since we already have the mapping information between ``spherical slices'' and ``ERP sub-patches'',
our ``grid-like spherical-to-2D projection'' can be realized as:
\begin{equation}
\begin{aligned}
\label{eq:mapping}
\hspace{-0.58em}\rm Sphere\leftarrow \mathcal{P}_{\rm E2S}(ERP),\\
\hspace{-0.58em}\rm \{SSlices, EPats, \mathbb{M}_{E\rightleftharpoons S}\} = \emph{SGrid}(Sphere, ERP),\\
\hspace{-0.58em}\rm \{SWindows, \mathbb{M}_{S\rightleftharpoons W}\} = \emph{SWindow}(Sphere, SSlices),\\
\hspace{-0.58em}\rm ERP^{\star}\leftarrow \emph{Fill}\big(EPats,\underbrace{\rm \mathcal{P}_{S2E}(SWindows)}_{\color{gray}{Distortion\ Free}}, \mathbb{M}_{S\rightleftharpoons W}, \mathbb{M}_{E\rightleftharpoons S}\big),
\end{aligned}
\end{equation}
\noindent where ``ERP'' is ERP image, ``Sphere'' is ERP's spherical form, ``SSlices'' denotes spherical slices, ``SWindows'' denotes spherical sub-windows, and ``EPats'' represents the topology of ERP sub-patches (see Fig.~\ref{fig:30deg}-B);
$\rm \mathbb{M}_{E\rightleftharpoons S}$ is the mapping between ``ERP sub-patches'' and ``spherical sub-slices'', $\rm \mathbb{M}_{S\rightleftharpoons W}$ denotes the mapping between ``spherical sub-slices'' and ``spherical sub-windows''; $\mathcal{P}_{\rm E2S}$ is a typical projection from ERP to sphere, and $\mathcal{P}_{\rm S2E}$ projects sphere back to ERP; \emph{SGrid}$(\cdot)$ performs grid-like dividing on sphere, \emph{SWindow}$(\cdot)$ divides the sphere to uniform sub-windows; \emph{Fill}$(\cdot)$ uses the EPats as the indicator to reformulate a distortion-free EPR, \emph{i.e.}, ERP$^{\star}$ (the final result), which can be detailed as:
\begin{equation}
\begin{aligned}
\label{eq:fill}
&\rm \color{gray}{Step\ 1:} &\rm \emph{f}_\emph{i} = \emph{PAlign}\big(\mathcal{P}_{S2E}(\rm SWindows), \mathbb{M}_{S\rightleftharpoons W}\big),\\
&\rm \color{gray}{Step\ 2:} &\rm \emph{f}_\emph{i} = \emph{PAlign}\big(\emph{f}_\emph{i}, \mathbb{M}_{\rm E\rightleftharpoons S}\big),\\
&\rm \color{gray}{Step\ 3:} &\rm ERP^{\star} = \emph{Reform}\big(\underbrace{\{\cdots,\ \emph{f}_\emph{i},\ \cdots\}}_{\color{gray}{{\rm All}\ \emph{f}}}, \rm EPats\big), \\
%ERP^{\star} = Refo(PA(PA(\mathcal{P}_{S2E}(SWindows), \mathbb{M}_{S\rightleftharpoons W}), \mathbb{M}_{E\rightleftharpoons S}), EPats),
\end{aligned}
\end{equation}
where \emph{f} is a temporary container, \emph{PAlign}$(\cdot)$ performs pixel-wise projection according to the given mappings (\emph{i.e.}, $\mathbb{M}$), \emph{Reform}$(\cdot)$ reformulate all distortion-free patches into a complete ERP map; the step 1 projects ``distortion-free spherical sub-windows'' to ``spherical slices'', the step 2 projects ``spherical slices'' to ``ERP sub-patches'', and the step 3 reformulates the obtained ``ERP sub-patches'' as the ERP$^\star$. The detailed definitions of ``spherical sub-windows'', ``spherical slices'', and ``ERP sub-patches'' can be reviewed from Fig.~\ref{fig:30deg}-B.

Qualitative demonstrations regarding ERP$^{\star}$ can be found in the ${{\normalsize{\textcircled{\scriptsize{2}}}\normalsize}}$ of Fig.~\ref{fig:pipeline} and Fig.~\ref{fig:30deg}-B.
Despite the advantage of ERP$^{\star}$ that all local patches are distortion-free, massive visual artifacts can be easily noticed, \emph{i.e.}, inter-patch misalignments, and ghost effects (see ${{\normalsize{\textcircled{\scriptsize{2}}}\normalsize}}$ of Fig.~\ref{fig:pipeline}).

\vspace{-9pt}
\subsection{Mesh Screen}
\label{sub:MS}
\vspace{-2pt}
Recall that we have mentioned ``\emph{\textbf{a precondition}}'' in Sec.~\ref{sub:GSP} that we shall let HVS focus on inside regions of ERP sub-patches.
Here, we present a very tricky solution to reach this precondition.
Besides, this solution can well alleviate the inter-patch misalignments in ERP$^{\star}$ (Eq.~\ref{eq:mapping}).
That is, we propose applying an additional ``mesh screen'' on ERP$^{\star}$, see the black mesh screen in the ${\normalsize{\textcircled{\scriptsize{3}}}\normalsize}$ of Fig.~\ref{fig:pipeline}.

Our technical rationales are two-fold.
\textbf{First}, the HVS tends to focus on visual artifacts (see Fig.~\ref{fig:mesh}-A), and using the proposed ``mesh screen'' can shift HVS's focal to the inside contents of each ERP sub-patch (see Fig.~\ref{fig:mesh}-B).
The HVS usually pays less attention to regular patterns~\cite{borst2015common}, and the proposed mesh screen is a regular pattern, which could be automatically omitted. Also, we designed the mesh screen to have the same grid size as the ERP sub-patches, and thus, all inter-patch misalignments can be fully blocked. So, mesh screen can eliminate misalignments and focus HVS on ERP's center regions.
\textbf{Second}, though the mesh screen inevitably results in some information lost --- the mesh screen fully blocks the inter-patch regions, our brain can automatically restore the overall ERP context.
This phenomenon is automatically achieved by the POV (Persistence of Vision)~\cite{anderson1978myth} mechanism, which says that the visual clues echo in our brain for a while after the visual signals have been lost.
Since our task is in a video environment, the mesh screen won't actually cause information loss due to the POV mechanism.

\begin{figure}[!t]
\centering
\includegraphics[width=1\linewidth]{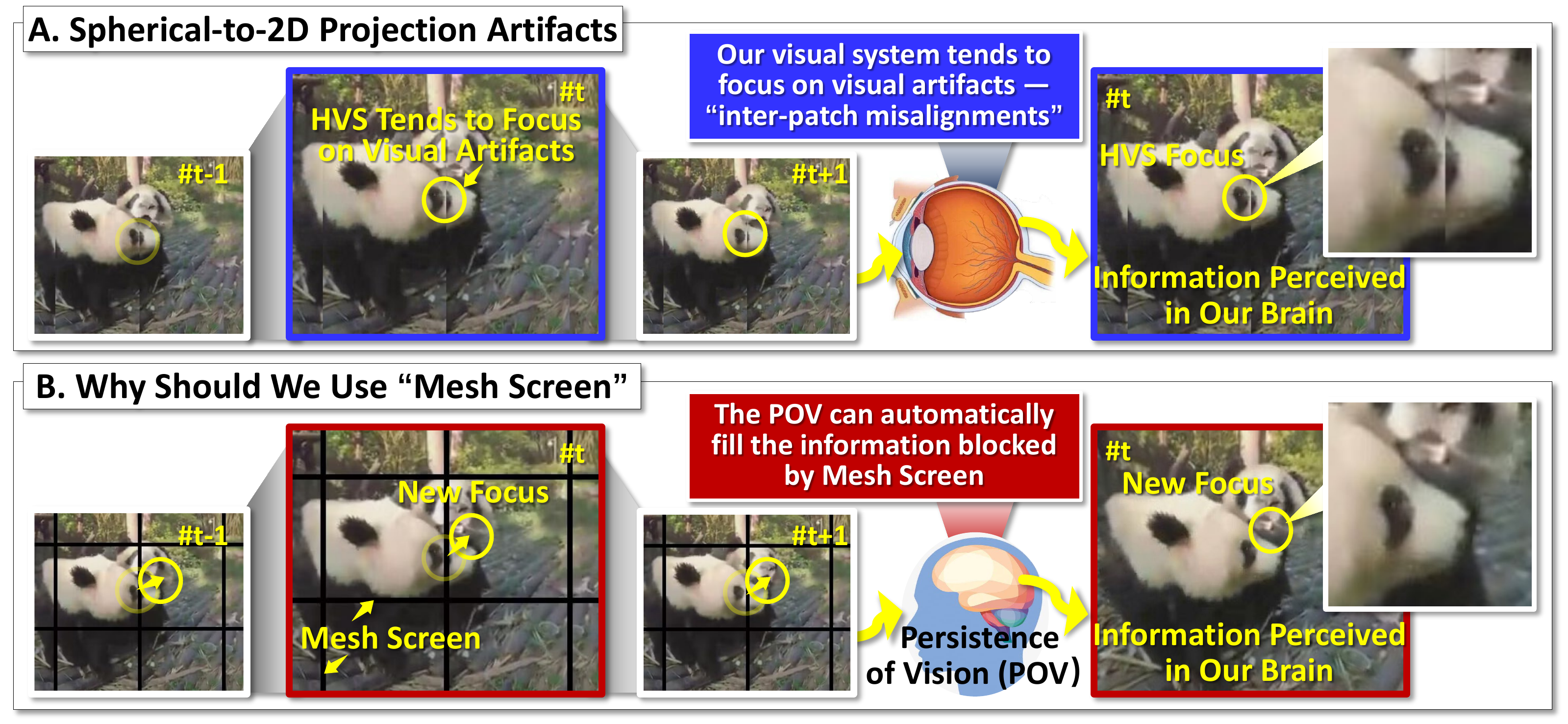}
\vspace{-23pt}
\caption{
Schematic illustration of why we should use ``Mesh Screen". Sub-figure A and B represent the visual information flow without and with a ``mesh screen" in visual input. In the input of \textbf{sub-figure A}, due to inter-patch misalignments, HVS will continue to focus on inter-patch misalignments, and the information the brain perceives will also focus on inter-patch misalignments. In \textbf{sub-figure B}, the inter-patch misalignments are blocked by the mesh window, which will automatically trigger the brain's automatic filling mechanism so that the brain will focus on the event itself. See Sec.~\ref{sub:MS} for details.
}
\label{fig:mesh}
\vspace{-13pt}
\end{figure}

Our ``mesh screen'' consists of two parts, including mesh screen generation (\emph{i.e.}, generate the ``GMask'') and mesh screen deployment (\emph{i.e.}, via $\odot$), which can be detailed as:
\begin{equation}
\begin{aligned}
\label{eq:remapxx}
&\hspace{0.18em}{\rm ERP^{\star\star}}\in\mathbb{R}^{w\times h}={\rm ERP^{\star}}\odot\underset{\Uparrow}{\underline{\rm{GMask}}},\\[-4pt]
&\hspace{2.15em}\ \ \ \ \ \ \ \ \ \overbrace{\rm GMask = \emph{Grids}(ERP, EPats)\in\{0,1\}^{\emph{w}\times \emph{h}}}
\end{aligned}
\end{equation}
where $\rm ERP^\star$ can be obtained via Eq.~\ref{eq:mapping}, ${\rm ERP^{\star\star}}$ is the result with mesh screen solution;
``EPats'' has been defined in Eq.~\ref{eq:mapping} --- the topology information among the ERP sub-patches, \emph{Grids}$(\cdot)$ extracts the grid structure, \emph{i.e.}, the GMask\footnote{We use ``0'' to represent the grid, whose thickness is 5 pixels.} from its input, and $w$, $h$ respectively denotes the ERP image's width and height; $\odot$ is element-wise multiplicative operation.

Using Eq.~\ref{eq:remapxx}, the inter-patch misalignments have been addressed nicely (see ${{\normalsize{\textcircled{\scriptsize{3}}}\normalsize}}$ of Fig.~\ref{fig:pipeline}), and, of course, the ``precondition'' has also reached.
Next, we shall conquer the remaining ``ghost effects''.

\vspace{-9pt}
\subsection{Discriminative Vertical Blur}
\label{sub:BOR}
\vspace{-2pt}
Although we adopted the mesh screen (Sec.~\ref{sub:MS}) to address the inter-patch misalignment problem, we can notice the ``ghost effects'' in ERP$^{\star\star}$ (Eq.~\ref{eq:remapxx}), especially in those sub-patches near the poles.
The main reason for causing the ghost effects has been explained above, \emph{i.e.}, the existence of overlapped regions between two horizontally neighbored ``spherical sub-windows'' (see Fig.~\ref{fig:30deg}-B), and the interested readers shall refer to Sec.~\ref{sub:GSP}.

To handle the ``ghost effects'', our idea is to blur those overlapped regions of ``spherical sub-windows''.
Although this solution looks rude, resulting in some information loss, it effectively alleviates the ghost effects.
The rationale is quite similar to that of the proposed ``mesh screen (Sec.~\ref{sub:MS})'', \emph{i.e.}, the POV mechanism can automatically help our brain restore the primary information of those blurred regions.
Also, the HVS can still possibly be attracted by these blurred regions since the motions, a very critical clue for attracting fixations, remain noticeable after the blur operation\footnote{The HVS is extremely sensitive towards movements~\cite{borst2015common}.}.

Since the ``ghost effects'' are more frequent near the sphere's poles, the ``spherical sub-windows'' near the poles shall have larger regions to be blurred than those near the equator, and that's why we name it as ``discriminative vertical blur (DVB)\footnote{In our implementation, we leave the two rows which are closet to the equator unchanged since they tend to have zero ghost effect.}'', the details of DVB can be given as:
\begin{equation}
\begin{aligned}
\label{eq:cccc}
\hspace{-0.48em}{\rm SWindow}_{i,j} &= {\rm Olap}_{i,j} \cup \big\{{\rm SWindow}_{i,j}-{\rm Olap}_{i,j}\big\},\\
\hspace{-0.48em}{\rm Olap}_{i,j} &= \big\{{\rm SWindow}_{i,j}\cap {\rm SWindow}_{i,j-1}\big\}\\
\hspace{-0.48em}&\ \ \ \ \ \ \ \ \cup \big\{{\rm SWindow}_{i,j}\cap {\rm SWindow}_{i,j+1}\big\},\\
\hspace{-0.48em}{\rm SWindow}_{i,j}^b &\leftarrow \mathcal{B}\Big({\rm Olap}_{i,j}\Big)\cup \Big\{{\rm SWindow}_{i,j}-{\rm Olap}_{i,j}\Big\},
\end{aligned}
\end{equation}
where ``SWindows'' denotes spherical window which has been defined in Eq.~\ref{eq:mapping}; $\cap$ is the intersection operation, and $\cup$ is the union operation; ${\rm Olap}$ denotes the overlapped regions --- the regions containing ghosts effects; $i$ and $j$ respectively denote a SWindow's row index and column index; $\mathcal{B}(\cdot)$ is a typical Gaussian blur\footnote{We empirically set the Gaussian blur as $ksize$=31 and $\sigma$=5.}; ${\rm SWindow}^b$ is the output of our DVB.

Then, the overall of our WinDB can now be changed from Eq.~\ref{eq:mapping} to the following equation:
\begin{equation}
\begin{aligned}
\label{eq:mapping2}
& \hspace{3.95cm}\rm Sphere\leftarrow \mathcal{P}_{\rm E2S}(ERP),\\
& \hspace{0.24cm}\rm \{SSlices, EPats, \mathbb{M}_{E\rightleftharpoons S}\} = \emph{SGrid}(Sphere, ERP),\\
& \hspace{-0.1cm}\rm \{SWindows, \mathbb{M}_{S\rightleftharpoons W}\} = \emph{SWindow}(Sphere, SSlices),\\
& \hspace{0.73cm}\rm ERP^{\star\star \emph{b}} \leftarrow {\rm Mesh}(\underset{\Uparrow}{\underline{\rm ERP^{\star \emph{b}}}}),\\[-4pt]
&\hspace{-0.14cm}\overbrace{\emph{Fill}\Big({\rm EPats},{\rm \mathcal{P}_{S2E}\big(\underset{\Uparrow}{\underline{DVB}}(SWindows)\big)}, \mathbb{M}_{\rm S\rightleftharpoons W}, \mathbb{M}_{\rm E\rightleftharpoons S}\Big)}\\[-4pt]
&\hspace{0.17cm}\overbrace{\rm \color{gray}{Discriminative\ Vertical\ Blur\ (Eq.\ \ref{eq:cccc})}}
\vspace{0.2cm}
\end{aligned}
\end{equation}
where most of the symbols are the same as Eq.~\ref{eq:mapping}; Mesh$(\cdot)$ denotes the proposed mesh screen (Eq.~\ref{eq:remapxx}); ERP$^{\star\star \emph{b}}$ is the output of Eq.~\ref{eq:mapping2}, which can be visualized in ${{\normalsize{\textcircled{\scriptsize{4}}}\normalsize}}$ of Fig.~\ref{fig:pipeline}.
As shown, the ghost effects near the equator have been alleviated significantly, yet some ghost effects are still noticeable near the poles even after applying the proposed DVB. Next, we shall further improve it.
\vspace{6pt}

\vspace{-9pt}
\subsection{Auxiliary Windows}
\label{sub:UAW}
\vspace{-2pt}
To further handle the ghost effects in ERP$^{\star\star \emph{b}}$ (Eq.~\ref{eq:mapping2}), we propose to use ``auxiliary windows''.
Generally, the concept of auxiliary windows is quite simple, \emph{i.e.}, using multiple tailored ``spherical sub-windows'' to facilitate users to watch the panoptic scene's pole regions, see ${{\normalsize{\textcircled{\scriptsize{5}}}\normalsize}}$ of Fig.~\ref{fig:pipeline}.
The primary principles of designing ``auxiliary windows'' include three aspects.
\textbf{First}, for a better watching experience, the adopted auxiliary windows shall have larger coverage and as much as possible while staying distortion-free.
\textbf{Second}, the auxiliary windows can only occupy a small part of the WinDB's main screen to maintain the ``global'' perceptual of ERP$^{\star\star \emph{b}}$.
\textbf{Third}, the auxiliary windows shall have less overlapping to alleviate the ghost effects significantly.

According to these principles, we design the auxiliary windows' covering scope as vertical-45$^{\rm o}$ and horizontal-120$^{\rm o}$.
This design is mainly because: \textbf{1)} the maximum horizontal range of the distortion-free projection $\mathcal{P}_{\rm S2E}$ (Eq.~\ref{eq:mapping}) is about 120$^{\rm o}$~\cite{cheng2018cube} and the HVS's focal range is below 120$^{\rm o}$~\cite{yang2018object} (Eq.~\ref{fig:30deg}-A), and thus we set the auxiliary windows' horizontal covering range to 120$^{\rm o}$; \textbf{2)} since the severest ghost effects are near the poles, we empirically set the horizontal covering range as 45$^{\rm o}$ to strike the tradeoff between retaining global perceptual and suppressing ghost effects.
Thus, we have total $N$ auxiliary windows (see ${{\normalsize{\textcircled{\scriptsize{5}}}\normalsize}}$ of Fig.~\ref{fig:pipeline}), where the auxiliary windows are distortion-free, placed at the top and bottom rows to block the ghost effects, and about 70\% regions of ERP$^{\star\star \emph{b}}$ are still accessible.

The whole process of deploying auxiliary windows (DAW) can be detailed as follows:
\begin{equation}
\begin{aligned}
\label{eq:AW}
% \rm Sphere\leftarrow \mathcal{P}_{\rm E2S}(ERP),\\
&\rm \hspace{0.45cm} WinDB^{-} \rm  = DAW(\underset{\Uparrow}{\underline{AWs}}, {\rm ERP^{\star\star b}}),\\[-4pt]
&\hspace{0.46cm}\overbrace{{\rm AW}_i=\mathcal{P}_{\rm S2W}\big(\underset{\Uparrow}{\underline{{\rm SWindow^+}_i}}\big),i\in\{1,...,N\}}\\[-4pt]
&\hspace{-0.11cm}\overbrace{{\rm SWindow}^+ = \emph{SWindow}\big({\rm Sphere}, {\rm SSlices}^+\big),\ {\rm Eq}.~\ref{eq:mapping2}}\\[0.1cm]
\end{aligned}
\end{equation}
where most of the symbols used have been defined in Eq.~\ref{eq:mapping2}, $N$ represents the total number of auxiliary windows; compared with the previous ``SWindows'' and ``SSlices'', ${\rm SWindows^+}$ and $\rm SSlices^+$ cover more spherical regions, \emph{i.e.}, about 6 times;
$\rm WinDB^{-}$ represents the early version of our WinDB (${{\normalsize{\textcircled{\scriptsize{5}}}\normalsize}}$ of Fig.~\ref{fig:pipeline}) which has multiple advantages, \emph{i.e.}, much alleviated ghost effects with global perception. Yet it still has some disadvantages, \emph{i.e.}, few ghost effects induced by the ``inevitable overlapping\footnote{Such overlapping is mainly induced by the strict ``1-to-many'' correlation between ``SWindow$^+$'' and ``SSlices.''}'' between the auxiliary windows, and, especially, fixations tend to be trapped in auxiliary windows because they tend to be more informative than the smaller ERP sub-patches.
In the next step, we shall handle these disadvantages to further improve our WinDB.

\vspace{-9pt}
\subsection{Dynamic Blurring}
\label{sub:ABT}
\vspace{-2pt}
Here, we aim at two problems of $\rm WinDB^{-}$ (Eq.~\ref{eq:AW}): \textbf{1)} the existence of few ghost effects, and \textbf{2)} trapping HVS fixations.
The collected fixations won't reflect the regional-wise importance degree if these two problems are unsolved.
Thus, we present the ``dynamic blurring'' solution, and the general idea is to blur all auxiliary windows and dynamically clear one that receives fixations.
In this way, the ghost effects can be solved completely, and the fixations won't be trapped in auxiliary windows.
Further, since the movements are still noticeable in blurred auxiliary windows, the blur operation won't cause much information to be lost, and this phenomenon has also been mentioned in Sec.~\ref{sub:MS}.

\begin{figure}[!t]
\centering
\includegraphics[width=1\linewidth]{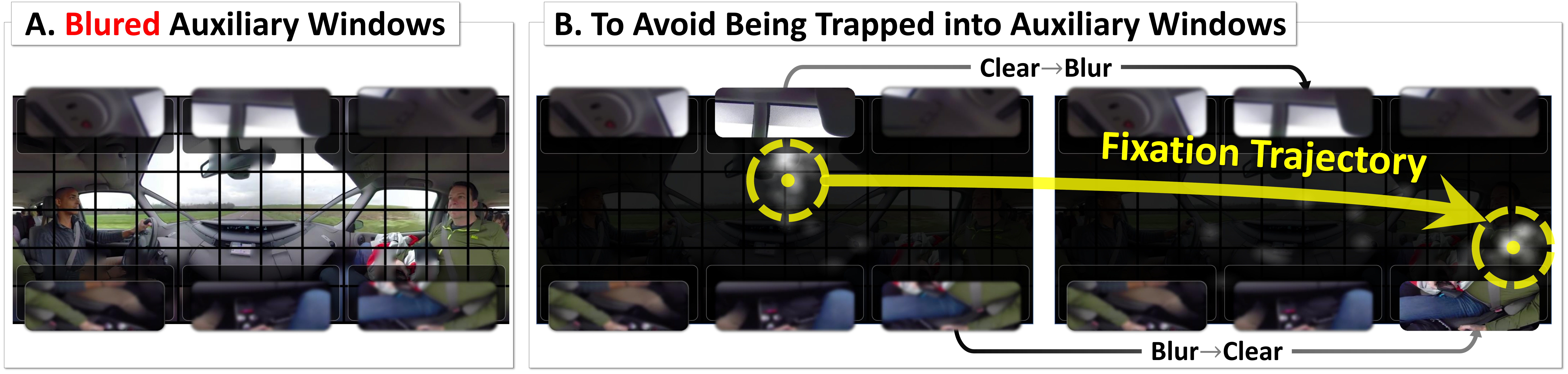}
\vspace{-23pt}
\caption{Technical details of the proposed auxiliary window with a dynamic blurring strategy. The auxiliary window changes (from blurred to clear) when the fixation trajectory sweeps over the auxiliary window. The advantage is that ghost effects can be solved completely, and the fixation won’t be trapped in auxiliary windows. See sec.~\ref{sub:ABT} for details.
}
\label{fig:awblur}
\vspace{-13pt}
\end{figure}

The technical details of the proposed dynamic blurring has been shown in Fig.~\ref{fig:awblur}, which cyclicly repeats three sequential status, \emph{i.e.}, \textbf{B} (blurred), \textbf{C} (clear),  and \textbf{R} (re-blurred).
The \textbf{B} status: at the beginning of the fixation collection, all auxiliary windows are blurred (Gaussian blur\footnote{OpenCV GaussianBlur tool, $ksize$=31 and $\sigma$=5.}).
The \textbf{C} status: a blurred auxiliary window would become clear immediately if the fixation trajectory sweeps over the auxiliary window during the fixation collection process.
The \textbf{R} status: to prevent trapping fixations into auxiliary windows, the ``clear status'' of an auxiliary window won't last long, and our method will ``gradually blur'' (last about 2$\sim$3 seconds) the ``clear'' auxiliary window again).
We can formulate the whole process of dynamic blurring (DB) as follows:
\begin{equation}
\begin{aligned}
\label{eq:remap}
&\rm \hspace{2.7em}WinDB={\rm DB}\normalsize(\underset{\Uparrow}{\underline{WinDB^{-}}}\normalsize),\\[-4pt]
&\rm \hspace{7.0em}\overbrace{\big\{\underset{\Uparrow}{\underline{{{\rm AW}_i}}},i\in{\rm \{1,...,\emph{N}\}}}\\[-4pt]
&\rm \hspace{-0.33cm}\overbrace{
\begin{tikzpicture}
\node (A) at (0,0) {$\textbf{B}$};
\node (B) at (3,0) {$\textbf{C}$};
\node (C) at (6,0) {$\textbf{R}$};
%\draw[->, bend left=10, line width=0.75pt] (C) to node[below, font=\footnotesize ] {Gradual Blurring, 2$\sim$ 3s} (A);
\draw[->, line width=0.75pt] (C) -- node[pos=0.01, below, font=\footnotesize] {\hspace{-6.0cm}Gradual Blurring, 2$\sim$3s} ++(0,-0.7) -| (A);
\draw[->, bend left=0, line width=0.75pt] (A) to node[above, font=\footnotesize ] {Receive Fixations} (B);
\draw[->, bend left=0, line width=0.75pt] (B) to node[above, font=\footnotesize] {Last 2s \& No Fixation} (C);
\end{tikzpicture}
}\\[-4pt]
\end{aligned}
\end{equation}
where
$\rm {DB}(\cdot)$ donates the proposed dynamic blur scheme which deplores on $\rm WinDB^{-}$ (Eq.~\ref{eq:AW}) with \emph{N} auxiliary windows;
${{\rm AW}_i}$ represents the \emph{i}-th auxiliary window; ``WinDB'' presents the final version of the proposed new panoptic fixation collection approach (see ${{\normalsize{\textcircled{\scriptsize{6}}}\normalsize}}$ of Fig.~\ref{fig:pipeline}).

In summary, our WinDB has the following advantages: 1) blind-zoom free, 2) no ghost-effects, 3) no inter-patch misalignment, 4) good global perceptual, 5) almost zero information lost, and 6) user friendly.
Thus, based on WinDB, we can easily collect solid fixations that can correctly reflect the regional-wise importance of the given panoptic scene.
This novel fixation collection tool has built a solid foundation for the panoptic saliency research field.

\vspace{-9pt}
\section{Proposed \ourData~Dataset}
\subsection{Why Should We Build This New Set?}
In the previous HMD-based datasets~\cite{xu2018gaze,cheng2018cube,zhang2018saliency}, we found almost no panoptic video containing ``sudden events'' that could result in ``fixation shifting'' --- driving users' fixations from one place to another with long spherical distance.
Actually, the ``fixation shifting'' is a very common event in our daily life, which could be an important event deserving the HVS's attention.
However, due to the blind-zoom issue, the fixations collected by HMD are fully unaware of such fixation shifting phenomenon.
Thanks to the advantages of our WinDB, we can now handle this problem.
Thus, based on our WinDB, we shall construct a new set, named \ourData, the most challenging and comprehensive panoptic saliency detection dataset, containing various complex scenes with frequent sudden important events.

\vspace{-6pt}
\subsection{Video Clip Collection}
\label{sub:vcc}
To construct the mentioned large dataset, we downloaded almost 400 video clips from YouTube, and nearly 80\% of them contain ``sudden events''.
Then, we removed about 100 low-quality clips (\emph{e.g.}, scenes with plain backgrounds, simple movements, or low-resolutions).
Thus, there is a total of 300 high-quality ones retained.
It is worth mentioning that, in the previous datasets~\cite{zhang2018saliency,zhang2022pav,xu2018gaze,xu2018predicting}, those clips contain almost no ``sudden events'' since their fixation collection is ill-posed. Yet, in facing ``sudden events'', the fixations can be collected by our WinDB correctly.
Fig.~\ref{fig:semanticdis} shows the semantic categories covered by \ourData, \emph{i.e.}, 225 categories.

\begin{figure}[!t]
\centering
\includegraphics[width=0.67\linewidth]{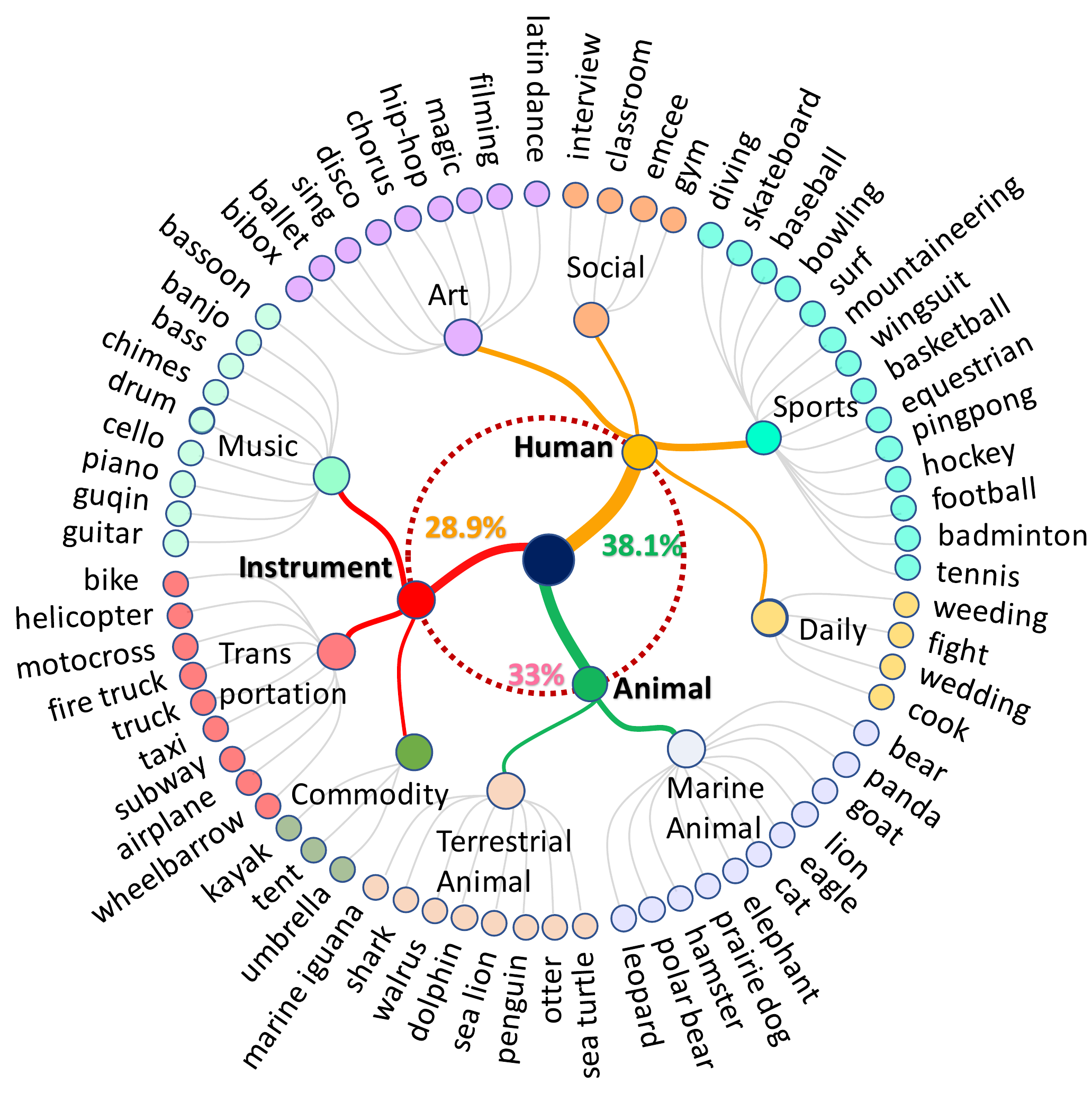}
\vspace{-16pt}
\caption{The semantic categories of \ourData~dataset.}
\label{fig:semanticdis}
\vspace{-13pt}
\end{figure}

\vspace{-6pt}
\subsection{Users Fixation Collection with WinDB}
Based on our newly proposed panoptic fixation collection approach (\emph{i.e.}, WinDB), we have recruited 38 users, including 12 females and 26 males aged between 18$\sim$29.
All users are completely unfamiliar with the fixation collection process; of course, no video clips in our video clip pool have been shown to them before.
Since our WinDB approach is HMD-free, each user only needs to watch the video with a resolution of 1,920$\times$1,080 on the PC, and a typical eye tracker has also been set up. For each user, the entire fixation collection process takes about 50 minutes. It could be actively suspended at any time if the user experienced fatigue or discomfort during fixation collection.
Notice that this HMD-free approach (\emph{i.e.}, WinDB) is more comfortable than the HMD-based one, not to mention that fixations collected by our approach are more consistent with the real regional-wise importance degree in the given panoptic scene.

\begin{figure}[!t]
\centering
\includegraphics[width=1\linewidth]{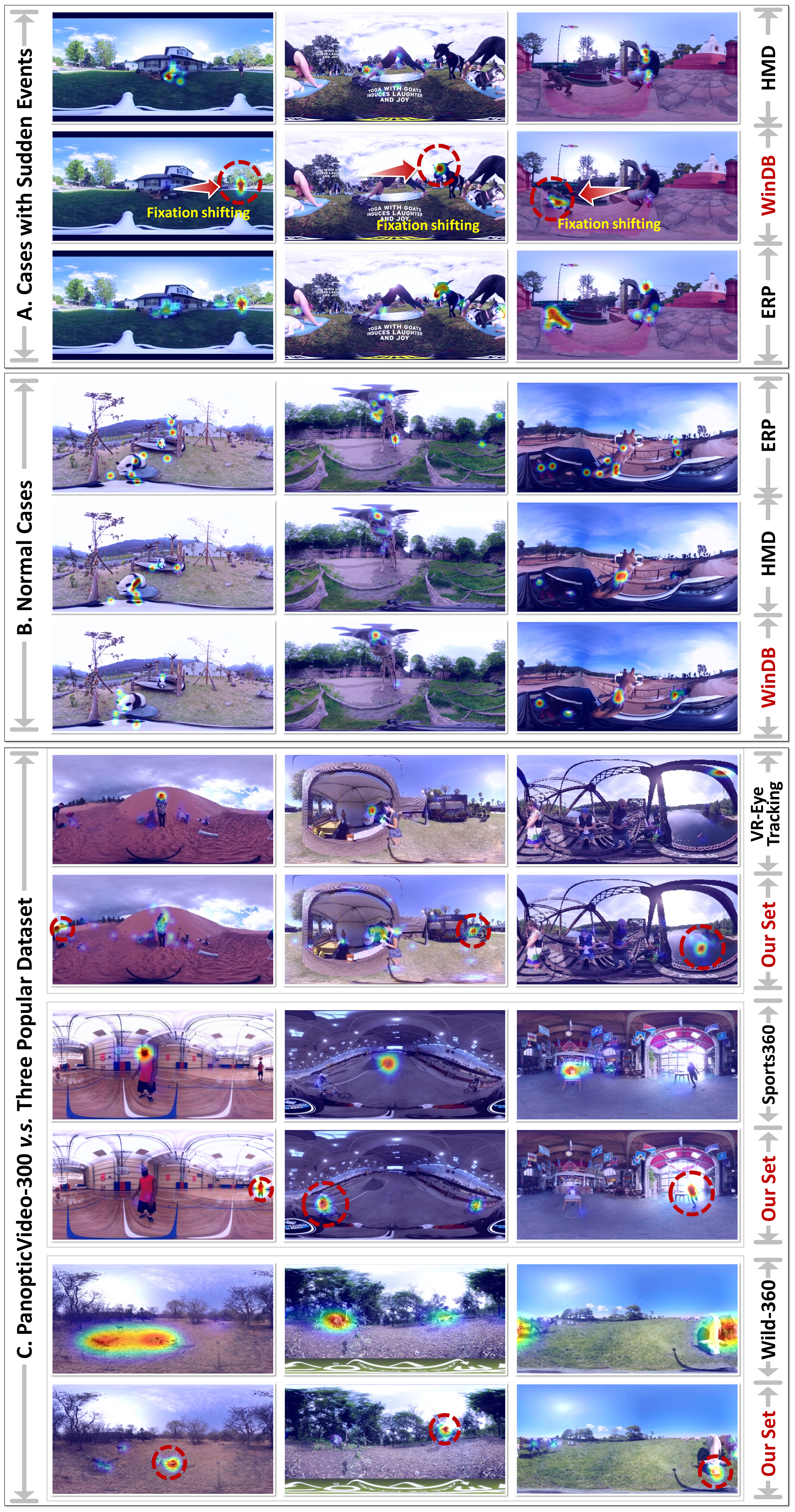}
\vspace{-26pt}
\caption{
Qualitative comparisons between fixation collected by our \ourApproach~approach, conventional fixation collection methods, and existing datasets (\emph{i.e.}, VR-Eye Tracking~\cite{xu2018gaze}, Sports360~\cite{zhang2018saliency}, and Wild-360~\cite{cheng2018cube}). The fixation shifting phenomenon has been highlighted via red cycles. See sec.~\ref{sub:AD} for details.
}
\label{fig:demo}
\vspace{-19pt}
\end{figure}

\vspace{-6pt}
\subsection{Advantage Discussions}
\label{sub:AD}
To demonstrate the advantage of our WinDB against the HMD/ERP-based ones, we provide some demonstrations in Fig.~\ref{fig:demo}, where we have made three in-depth discussions.

\textbf{First}, Fig.~\ref{fig:demo}-A illustrates the cases with ``sudden events'', where important and noticeable events suddenly occur in regions far away from the current main fixations.
By comparing these three rows, the fixations collected by our approach (the 2nd row) are more reasonable than that collected by the HMD-based method (the 1st row), where our approach can capture those ``sudden events'' (highlighted by red cycle) yet the HMD-based method cannot.
The reason is that the sudden events in these cases occurred in the HMD's blind zoom; the users with HMD have missed these events.
Since our WinDB has no blind zoom, the users can fully notice all sudden events, ensuring correct fixation collection.

\textbf{Second}, to verify the advantage of our WinDB against the ERP-based one, we have shown some normal cases in Fig.~\ref{fig:demo}-B.
Without ``sudden events'', the fixations collected by our approach are generally consistent with those collected by the HMD-based method, showing the correctness of our approach.
Also, compared with the ERP-based method, the fixations collected by our WinDB are generally compact, yet the ERP-based fixations are scattered.
The reason is clear that the visual distortions in ERP (especially for those regions near the poles) could easily influence human-eye fixations, drawing them to those distortion-induced visual artifacts. % (see Fig.~\ref{fig:WinDBERP})
Since our WinDB is generally distortion-free, its fixations can be focused on the salient regions.

\textbf{Third}, in the above two discussions, the competitors were ``re-realized" by our team. To further demonstrate the advantage of our PanopticVideo-300 against other real datasets (\emph{i.e.}, \cite{xu2018gaze,zhang2018saliency,cheng2018cube}), we have provided some representative qualitative comparisons in Fig.\ref{fig:demo}-C.
Our fixations are more consistent with the real importance degree of the given panoptic scene, and the reasons have been explained above (\emph{i.e.}, distortion-free and no-blind zoom). %(\emph{i.e.}, VR-Eye Tracking~\cite{xu2018gaze}, Wild-360~\cite{cheng2018cube}, and Sports360~\cite{zhang2018saliency})

%\begin{figure}[!t]
%\centering
%\includegraphics[width=1\linewidth]{SphereERP.pdf}
%\vspace{-26pt}
%\caption{
%Comparison of fixation collected using three methods.
%The ERP-based method often captures multiple fixations that may correspond to just one fixation when projected onto the sphere. In contrast, our WinDB approach is distortion-free and provides fixation results consistent with the HMD-based method.  See sec.~\ref{sub:AD} for details.
%}
%\label{fig:WinDBERP}
%\vspace{-13pt}
%\end{figure}

\begin{figure*}[!t]
\centering
\includegraphics[width=1\linewidth]{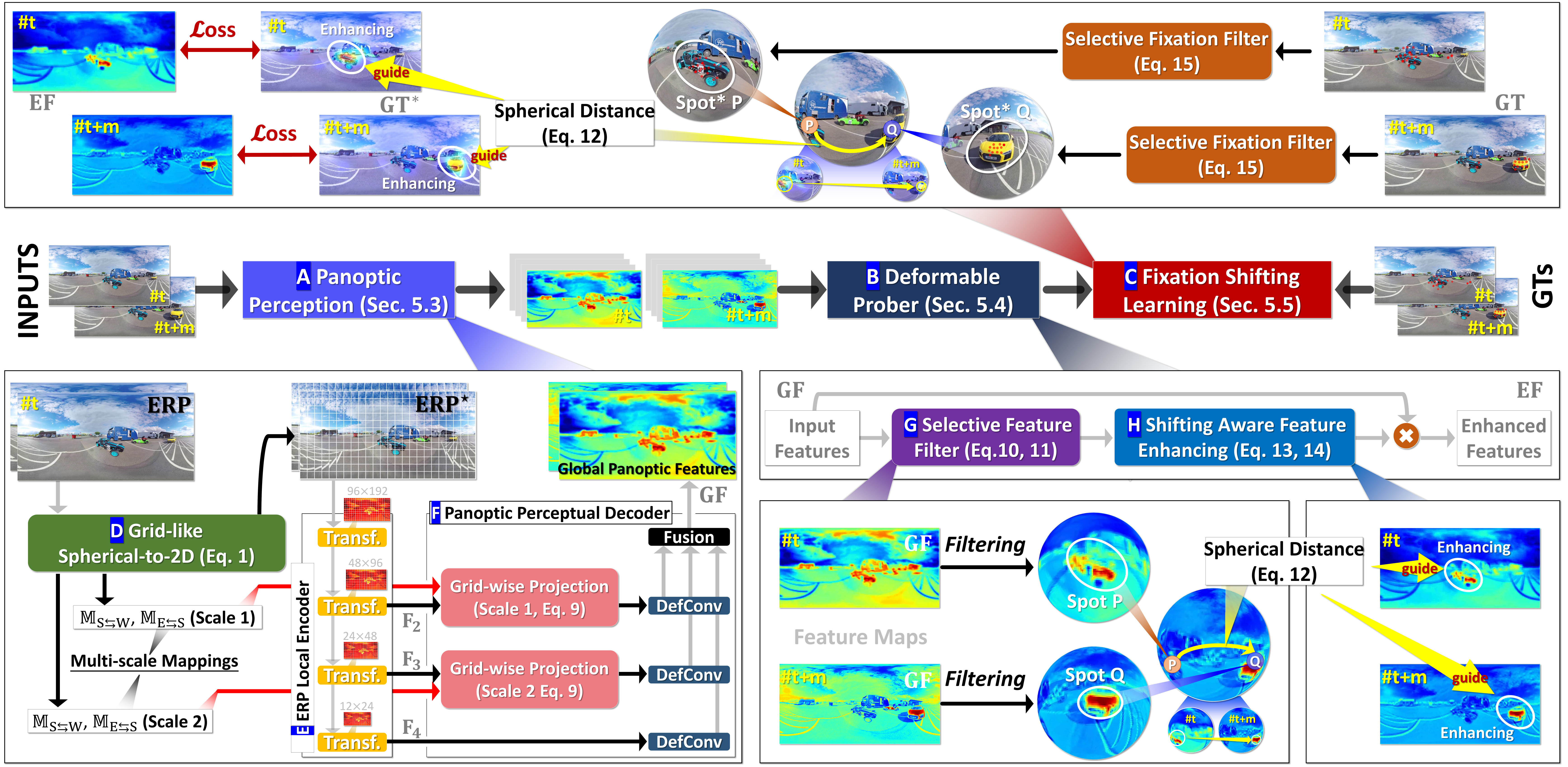}
\vspace{-24pt}
\caption{
The detailed network architecture of our \textbf{Fi}xation \textbf{Sh}ifting \textbf{Net}work (\textbf{FishNet}). Our FishNet has three major components. Component \textbf{A} focuses on performing ERP-based global feature embedding to achieve panoptic perception and avoid visual distortion. \textbf{B} catches fixation shifting in PanopticVideo-300 by refocusing the network to avoid the compression problem of shifted fixations in SOTA models. \textbf{C} makes the network fully aware of and learns the fixation shifting behind mechanism to ensure that the network is sensitive to fixation shifting. See section~\ref{sub:NetO} for details.
}
\label{fig:FishNet}
\vspace{-13pt}
\end{figure*}

\vspace{-9pt}
\section{The Proposed FishNet}
\label{Sec:FishNet}
\vspace{-2pt}
\subsection{Why Should We Need This New Network?}
\label{sec: wswntnn}
\vspace{-2pt}
Now, we have a dataset (\emph{i.e.}, PanopticVideo-300) containing a unique phenomenon --- ``fixation shifting'', making this set very challenging to the existing SOTA fixation prediction methods~\cite{li2023spherical,djilali2021rethinking,yun2022panoramic}.
Here, we shall brief the main reasons causing this challenge.
\textbf{First}, to fully use the spatiotemporal information, a critical aspect for the fixation prediction to suppress false alarms, the SOTA methods~\cite{dahou2021atsal,xu2021spherical,zhang2018saliency} designed their networks to constrain their fixations to stay spatiotemporal smoothness.
However, the ``fixation shifting'' is generally intermittent, where fixations could suddenly jump from one place to another with long spatial distance (see Fig.~\ref{fig:demo}), which is clearly different from ``normal fixations'' --- fixations are spatiotemporally smooth.
This contradiction between these two types of fixations makes the learning process extremely challenging.
So, the ``fixation shifting'' could get compressed or omitted if we directly follow the existing network design that heavily relies on the constraint of staying spatiotemporal smoothness.
\textbf{Second}, almost all previous SOTA methods cannot achieve panoptic global perception, failing to sense sudden events and eventually resulting in their incompetence to catch fixation shifting.
As the first attempt, we present a fancy network design, coined as FishNet, to handle fixation shifting, whose key technical innovations include \textbf{1)} a fancy ``panoptic perception'' to perceive sudden events globally and \textbf{2)} a brand-new ``deformable prober'' and ``fixation shifting learning'' to handle the jumpable fixations.

\vspace{-9pt}
\subsection{Network Overview}
\label{sub:NetO}
\vspace{-2pt}
As shown in Fig.~\ref{fig:FishNet}, our FishNet model mainly contains three major components (\emph{i.e.}, ``\textbf{\textcolor[RGB]{255,255,255}{\sethlcolor{lightblue}{\hl{\textbf{\rm A}}}}} panoptic
perception'' and ``\textbf{\textcolor[RGB]{255,255,255}{\sethlcolor{lightblue}{\hl{\textbf{\rm B}}}}} deformable prober'') with a tailored learning scheme (\emph{i.e.}, ``\textbf{\textcolor[RGB]{255,255,255}{\sethlcolor{lightblue}{\hl{\textbf{\rm C}}}}} fixation shifting learning'').

Taking ERP images as input, the ``sudden events'' could be easily captured by the network because the ERP-based feature embedding is a global process that can let the network's sensing scope cover the entire panoptic scene.
So, the key task for the ``\textbf{\textcolor[RGB]{255,255,255}{\sethlcolor{lightblue}{\hl{\textbf{\rm A}}}}} panoptic perception'' (see Fig.~\ref{fig:FishNet}) is to avoid visual distortions when performing ERP-based global feature embedding, which will be detailed in Sec.~\ref{sub:BAOP}.

The primary objective of ``\textbf{\textcolor[RGB]{255,255,255}{\sethlcolor{lightblue}{\hl{\textbf{\rm B}}}}} deformable prober'' is to let the network being capable of catching ``fixation shifting'' that prevalent in our PanopticVideo-300.
Most of the existing video fixation prediction networks are heavily dependent on ``spatiotemporal'' information, which implies a strong constraint --- the predicted fixations shall stay smooth over spatial and temporal.
So, the shifted fixations are more likely to be compressed by the existing networks.
To improve, our novel ``deformable prober'' provides an inspiring way to ``refocus'' the network on shifted fixations without any side effects, which will be detailed in Sec.~\ref{sub:DFP}.

Moreover, to make our FishNet ``sensitive'' towards the ``fixation shifting,'' we shall also let the training process be fully aware of whether the given GTs have fixation shifting. If fixation shifting exists, the training process shall automatically learn its behind the mechanism.
We realize this objective via the newly devised ``\textbf{\textcolor[RGB]{255,255,255}{\sethlcolor{lightblue}{\hl{\textbf{\rm C}}}}} fixation shifting learning'', which will be detailed in Sec.~\ref{Sub:FSL}.

\vspace{-6pt}
\subsection{Panoptic Perception}
\label{sub:BAOP}
\subsubsection{Technical Rationale}
As shown in the bottom-left of Fig.~\ref{fig:FishNet}, the proposed ``panoptic perception'' of FishNet mainly consists of three parts, \emph{i.e.}, \noindent\textbf{\textcolor[RGB]{255,255,255}{\sethlcolor{lightblue}{\hl{\textbf{\rm D}}}}}~{``grid-like spherical-to-2D''},
\noindent\textbf{\textcolor[RGB]{255,255,255}{\sethlcolor{lightblue}{\hl{\textbf{\rm E}}}}}~{``ERP local encoder''},
and \noindent\textbf{\textcolor[RGB]{255,255,255}{\sethlcolor{lightblue}{\hl{\textbf{\rm F}}}}}~{``panoptic perceptual decoder''}.

The technical rationale of panoptic perception is to achieve distortion-free global feature embedding.
Thus, we first use the ``grid-like spherical-to-2D'' that has been explained and detailed in Sec.~\ref{sub:GSP} to convert a typical ERP to a distortion-free version, \emph{i.e.}, ERP$^\star$ (Eq.~\ref{eq:mapping}).
Since all sub-patches in ERP$^\star$ are distortion-free, we propose to treat them as individual inputs to the ``ERP local encoder'', a typical Transformer-based multi-level encoder, to build the inter-patch relationships.
So, the features generated by the ``ERP local encoder'' are all distortion-free and with global perception.
However, these features are not well aligned with the original ERP containing redundant information, which is mainly induced by the ``ghost effects'' that have been mentioned and explained in Sec.~\ref{sub:BOR}.
Therefore, we have devised the ``panoptic perceptual decoder'', which performs the conventional multi-level feature collection, \emph{i.e.}, we use the newly devised ``grid-wise projection'' to project the encoder's features back to the ERP formation.
As shown in Fig.~\ref{fig:FishNet}, we have performed the ``grid-like spherical-to-2D'' projection two times to span a multi-scale ``grid'' space (\emph{i.e.}, Scale 1 and Scale 2), aiming to be in line with the subsequent multi-level structure of ``ERP local encoder".
After this process, we can obtain two important mapping information, \emph{i.e.}, $\mathbb{M}_{\rm S\rightleftharpoons W}$ and $\mathbb{M}_{\rm E\rightleftharpoons S}$, which will be used as the indicator to guide the feature alignments in the ``grid-wise projection'' part.

\vspace{-6pt}
\subsubsection{Technical Details}
The feature computation process of the ERP local encoder can be briefed as:
\begin{equation}
\begin{aligned}
\label{eq:feature}
&\rm \hspace{1.8cm}\color{gray}{ERP\ Local\ Encoder}\\[-5pt]
&\hspace{-0.32cm}\{\textbf{F}_{1,2,3,4}\}\!=\!\overbrace{{\rm ERPEnC}\big(Split(\underset{\Uparrow}{\rm \underline{ERP^\star}})\big)},\\[-4pt]
&\hspace{0.79cm} \overbrace{\underbrace{\rm GS2E\big(ERP\big)}\!\rightarrow\!\{\rm ERP^\star,\mathbb{M}_{S\rightleftharpoons W}, \mathbb{M}_{E\rightleftharpoons S}, EPats\}} \\[-3pt]
&\rm \hspace{0.2cm} \color{gray}{Grid\!\!-\!\!like\ Spherical\!\!-\!\!to\!\!-\!\!2D\ (Eq.~\ref{eq:mapping})}
\end{aligned}
\end{equation}
where \textbf{F}$_{1,2,3,4}$ denotes four intermediate features obtained via ``ERP local encoder", the operation \emph{Split}$(\cdot)$ splits ERP$^\star$ into individual patches to be fed as the input of encoder (\emph{i.e.}, ERPEnC); $\mathbb{M}_{\rm S\rightleftharpoons W}$, $\mathbb{M}_{\rm E\rightleftharpoons S}$ are the mappings, EPats is the ERP grid's topology information, which all have been defined in Eq.~\ref{eq:mapping} and will be used later (\emph{i.e.}, Eq.~\ref{eq:grid}).

As shown in the Fig.~\ref{fig:FishNet}, the intermediate features have varying sizes with ranges from ${\frac{\rm W}{4}}$$\times$${\frac{\rm H}{4}}$ to ${\frac{\rm W}{32}}$$\times$${\frac{\rm H}{32}}$~, in which the \emph{\rm W} and \emph{\rm H} is the ERP size.
We have abandoned the \textbf{F}$_{1}$ because the features from the shallow level tend to be generally noisy~\cite{gao2023thorough,liu2022poolnet+}.
Thus, only the three of them are actually used (\emph{i.e.}, \textbf{F}$_{2,3,4}$) in our ``panoptic perceptual decoder'', \emph{i.e.}, to align these distortion-free \& global feature representations with the original ERP topology.
Then, when \textbf{F}$_{2,3,4}$ arriving at the ``panoptic perceptual decoder'', \textbf{F}$_{2,3}$ are fed into the ``grid-wise projection'' to realize the feature mentioned above alignments.
Without undergoing the ``grid-wise projection'', the \textbf{F}$_{4}$ shall directly serve as the coarse localizer due to its low resolution.
So, the ``grid-wise projection'' in the ``panoptic perceptual decoder (\textbf{\textcolor[RGB]{255,255,255}{\sethlcolor{lightblue}{\hl{\textbf{\rm F}}}}} in Fig.~\ref{fig:FishNet})'' can be detailed as:
\begin{equation}
\label{eq:grid}
\begin{aligned}
\vspace{0.1cm}
{\rm G}&{\rm F} = {\rm \emph{Fusion}}\big({\rm \emph{Concat}}\big[\\
&\rm {DefConv}\big({\emph{Fill}}(EPats, \mathcal{P}_{E2S}(\textbf{F}_2),\mathbb{M}^1_{S\rightleftharpoons W}, \mathbb{M}^1_{E\rightleftharpoons S})\big), \\
&\ \ \ \ \ \rm {DefConv}\big({\emph{Fill}}(EPats, \mathcal{P}_{E2S}(\textbf{F}_3), \mathbb{M}^2_{S\rightleftharpoons W}, \mathbb{M}^2_{E\rightleftharpoons S})\big), \\
&\ \ \ \ \ \ \ \ \ \ \rm {DefConv}(\textbf{F}_4)\big]\big),
\vspace{0.1cm}
\end{aligned}
\end{equation}
\noindent where GF is the obtained global panoptic features, ``Concat$\left(\cdot\right)$" represents the function for channel-level feature concatenation;
EPats, $\mathbb{M}$ can be obtained from Eq.~\ref{eq:feature}, and the superscripts of $\mathbb{M}$ denote different scales; function \emph{Fill}$(\cdot)$ has been defined in Eq.~\ref{eq:mapping} and detailed in Eq.~\ref{eq:fill}; \emph{Fusion}$(\cdot)$ is a typical collector which includes convolution, batch-normalization, and ReLU; notice that ``DefConv'' denotes deformable convolution~\cite{dai2017deformable}, which can further mitigate the tiny misalignments\footnote{There exist some tiny misalignments when projecting ``spherical sub-windows'' to ``spherical slices'' due to the slight coverage mismatch on the spherical surface.} after the \emph{Fill}$(\cdot)$ operation.

\vspace{-3pt}
\subsubsection{Panoptic Perception v.s. SOTA Solutions}
\label{sub:PPSS}
There are three types of SOTA panoptic networks in our research community; we shall briefly explain our advantages.

1) Bi-stream Networks~\cite{wang2022bifuse,dahou2021atsal,qiao2020viewport,cheng2018cube}. This type of network consists of two research branches (Fig.~\ref{fig:comSOTA}-A), \emph{i.e.}, one branch for the ERP global information with heavy visual distortions, and another branch for the local distortion-free views. These two research branches are combined to make the network's features global and distortion-free at the same time.
The critical problem of this type of network is that the fusion between these research branches is extremely challenging --- they don't have explicit alignments, making the fusion to be an additional feature embedding.
Thus, the features obtained by this approach are inferior to ours.

2) Spherical Convolution-based Networks~\cite{djilali2021rethinking,zhu2019prediction,li2023spherical,su2021learning,xu2021spherical,zhang2018saliency}. Actually, this type of network has a perfect design, which can simultaneously achieve global panoptic perception and stay distortion-free (Fig.~\ref{fig:comSOTA}-B). However, the key problem is that the adopted ``spherical convolution'' differs from the existing CNN, making all 2D feature backbones that could provide strong semantic feature embedding unavailable.
Consequently, these types of networks generally perform poorly without the support of semantic information.

3) Transformer-based Networks~\cite{yun2022panoramic}. This type of network resorts to additional CNN layers after each transformer layer, \emph{i.e.}, the CNN-based local projection, to handle side-effects of ERP's distortions (Fig.~\ref{fig:comSOTA}-C). The major problem is that the intermediate features are generated using ERP's patches with several visual distortions to original ERP, which cannot take full advantage of the pre-trained feature backbones.
%Thus, the visual distortions-induced side-effects still exist even though the adopted ``local projection\footnote{Here, the ``local projection'' refers to the $\mathcal{P}_{\rm S2E}$.} (Eq.~\ref{eq:mapping})'' can alleviate some distortions.

4) Our Panoptic Perception. As shown in Fig.~\ref{fig:comSOTA}-D, our approach is generally ``independent'' to the existing transformers, which can be regarded as a generic plug-in to handle the visual distortions when generating ERP-based global features.
Notice that our solution enables end-to-end training and testing.
Thus, this plug-in nature enables the network to make full use of the pre-trained feature backbones~\cite{wang2021pyramid}.
So, the features obtained by our approach are global, distortion-free, and with strong semantics; all these attributes bring superior performance against the above-mentioned SOTAs.
And these attributes are also very necessary for handling the ``fixation shifting'' phenomenon, which will be detailed next.
In a word, our approach is very inspiring, and the proposed panoptic perception provides a brand-new foundation for feature computation, which is also the first attempt to bridge the gap between the conventional 2D research field and the panoptic research field.

\begin{figure}[!t]
\centering
\includegraphics[width=1\linewidth]{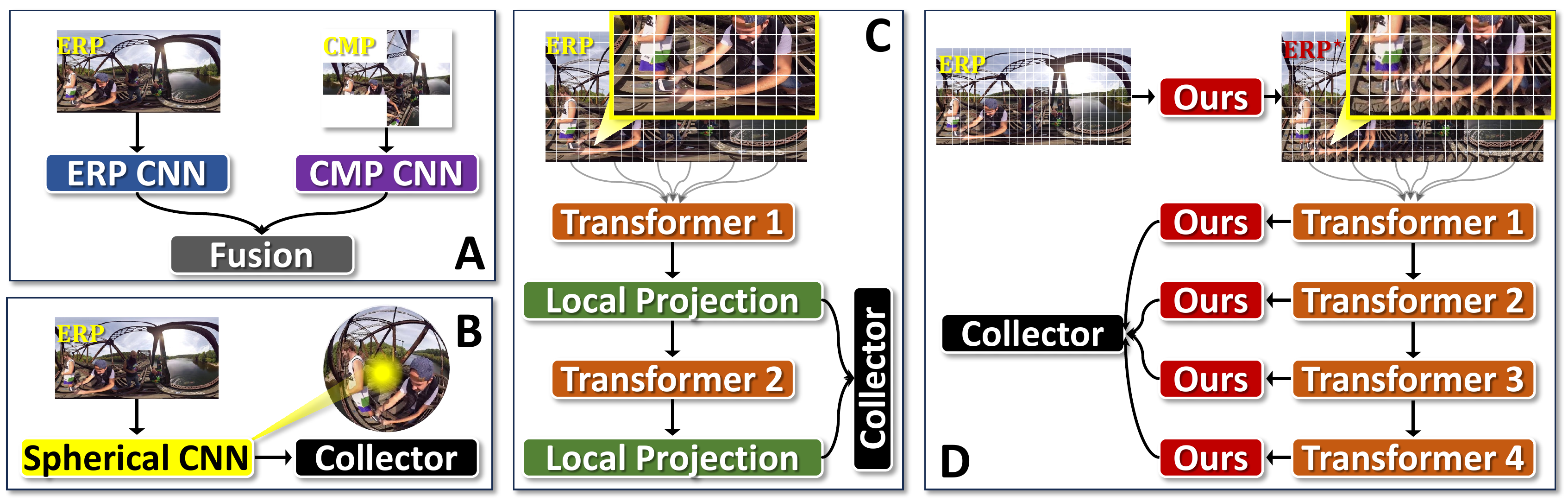}
\vspace{-25pt}
\caption{
The comparison of technical details between our FishNet and the SOTA panoptic video saliency learning methods. Sub-figure A, B, and C represent the SOTA panoptic video saliency learning method, and sub-figure D is our method. ``Ours" stands for our proposed panoptic perception component (\emph{i.e.}, Sec.~\ref{sub:BAOP}). The ``collector" is used to fuse the extracted panoptic ERP features.
%Representative methods of each branch: A-~\cite{wang2022bifuse,dahou2021atsal,qiao2020viewport,cheng2018cube}; B-~\cite{djilali2021rethinking,zhu2019prediction,li2023spherical,su2021learning,xu2021spherical,zhang2018saliency}; C-~\cite{yun2022panoramic}.
See section~\ref{sub:PPSS} for details.
}
\label{fig:comSOTA}
\vspace{-15pt}
\end{figure}

\vspace{-6pt}
\subsection{Deformable Prober}
\label{sub:DFP}
\subsubsection{Technical Rationale}
\vspace{-2pt}
The proposed ``deformable prober'' has been visualized in the bottom-right of Fig.~\ref{fig:FishNet}, consists of two parts, \emph{i.e.}, ``\noindent\textbf{\textcolor[RGB]{255,255,255}{\sethlcolor{lightblue}{\hl{\textbf{\rm G}}}}} selective feature
filter'' and ``\noindent\textbf{\textcolor[RGB]{255,255,255}{\sethlcolor{lightblue}{\hl{\textbf{\rm H}}}}} shifting aware feature
enhancing''.

As we have mentioned in Sec.~\ref{sec: wswntnn}, learning to predict the ``fixation shifting'' is very challenging because these unique fixations generally conflict with the ``normal fixations''.
That is, the normal fixations are spatiotemporally smooth, yet the shifted fixations are not, which tend to jump from one place to another with a long spherical distance.
Since the ``fixation shifting'' phenomenon is quite less than the normal ones\footnote{We designed our PanopticVideo-300 to include frequent ``fixation shifting'' on purpose, primarily because the other datasets almost have zero ``fixation shifting''.} in real works, it tends to be overwhelmed by the normal fixations, then be treated as noises during network training, and get compressed eventually.
So, we shall make the proposed FishNet ``fixation shifting aware'' --- exactly know if there exists fixation shifting and where they are.
Therefore, the key technical rationale of the proposed ``deformable prober'' is to adopt the ``selective feature filter'' to catch the shifted fixations, and then use the ``shifting aware feature enhancing'' to focus on those shifted fixations.

\begin{figure}[!t]
\centering
\includegraphics[width=0.86\linewidth]{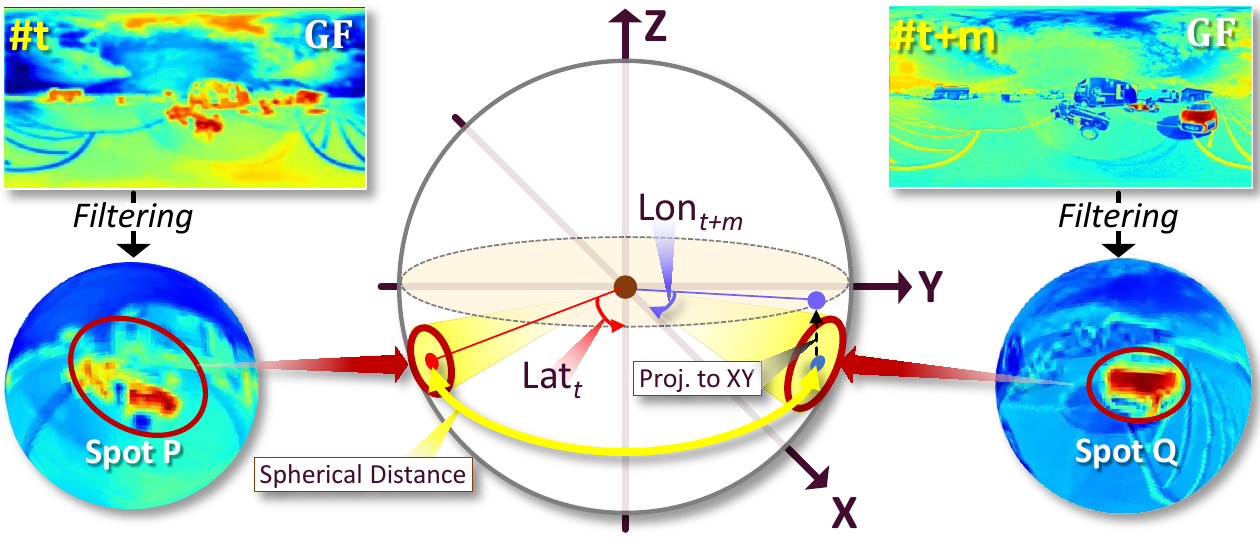}
\vspace{-16pt}
\caption{
Detailed calculation of the spherical distance of ``\textbf{Spot P}" and ``\textbf{Spot Q}". Spot P and Q are from the ``selective feature filter" component of FishNet. The main purpose is to measure the spherical distance between two ``spotlights" that belong to two different frames within a short time span. Lat$_t$ and Lon$_{t+m}$ denote the latitude and longitude of the $\emph{t}$-th and $\emph{\{t+m\}}$-th frame. See section~\ref{Sub:TDSFF} for details.
}
\label{fig:SphereD}
\vspace{-15pt}
\end{figure}

\vspace{-5pt}
\subsubsection{Technical Details of Selective Feature Filter}
\label{Sub:TDSFF}
To catch the shifted fixations, we shall know which attributes differentiate them from the ``normal fixations".
Generally, there are three attributes that make the fixation shifting phenomenon special.
\textbf{First}, the shifted fixations shall have strong feature responses.
\textbf{Second}, the regions correlating to ``fixation shifting'' shall receive the majority of fixations of the scene, \emph{i.e.}, to be the ``spotlight'' of the scenes.
The reason is clear that the ``fixation shifting'' always comes with ``sudden events''; without blind zoom, the users are very likely to be attracted and focused.
\textbf{Third}, there shall be a large spherical distance between temporally neighbored frames' spotlight region.
\emph{W.r.t.} all these attributes, we design the ``\noindent\textbf{\textcolor[RGB]{255,255,255}{\sethlcolor{lightblue}{\hl{\textbf{\rm G}}}}} selective feature filter'' of deformable prober as:
\begin{equation}
\label{eq:Att}
\begin{aligned}
&\hspace{3.2cm}\color{gray}{\rm Dynamic\ Thresholding}\\[-5pt]
&\hspace{-0.2cm}\Big\{\rm \underset{\Downarrow}{\underline{{Fo}_{1},...,{Fo}_\emph{u}}}\Big\}\!\!=\!\mathcal{C}\bigg(\bigg|\overbrace{\mathcal{M}({\rm GF})\!-\!\mathcal{T}_{d}\!\times\!{\rm \emph{max}}\Big\{{\mathcal{M}}({\rm GF})\Big\}}\bigg|_{+}\bigg),\\[-4pt]
&\hspace{-0.15cm}Ms\big\{\rm \cdots\big\}\rightarrow {\rm Spot}
\end{aligned}
\end{equation}
where ``$\mathcal{M}(\cdot)$" returns a matrix that represents the channel-wise mean of its input; GF denotes the features generated via Eq.~\ref{eq:grid}; \emph{max}$(\cdot)$ is the typical maximum function, and $\mathcal{T}_{d}$ is a pre-defined hard threshold; using the ``dynamic thresholding'', following the 1st attribute mentioned above, we can easily obtain those regions with high feature responses, and these regions are very potential to be the ``spotlight'' containing shifted fixations;~\!$|\cdot|_{\rm +}$ stands for keeping only the positive values of the matrix; $\mathcal{C}(\cdot)$ is the connected component analysis~\cite{xia2022tensorized}, which returns $u$ isolated regions (\emph{i.e.}, Fo);
Then, the ``spot" can be localized via the function \emph{Ms}$(\cdot)$, whose rationale correlates to the 2nd attribute mentioned above, which returns an isolated region that has the largest feature response (region's average).

So, the above process has let us meet the first two attributes of ``fixation shifting'', then we resort to ``spherical distance'' to satisfy the 3rd attribute, which measures the spherical distance between two ``spotlights'' that respectively belong to two different frames within a short time span. This process can be formulated as:
\begin{equation}
\begin{aligned}
\label{eq:weight}
\omega_t &= \Big|\Big|\mathcal{P}_{\rm E2S}\big(\digamma(\underset{\Uparrow}{\underline{{\rm Spot}_t}})\big),\mathcal{P}_{\rm E2S}\big(\digamma({\rm Spot}_{t+m})\big)\Big|\Big|_{\rm S},\\[-4pt]
&\hspace{0.39cm}\overbrace{\rm Spot_\emph{t} = \emph{SFF}(GF_\emph{t}),\ Eq.\ \ref{eq:Att}}
\end{aligned}
\end{equation}
where the subscripts $\emph{t}$ and $\emph{t+m}$ denote the $\emph{t}$-th frame and \{$\emph{t+m}$\} frame, and $\emph{m}$ random between \{1,2,...,15\}\footnote{Since the HVS's response limit is 500ms, \emph{i.e.}, about 15 frames.}; GF can be obtained by Eq.~\ref{eq:grid}, \emph{SFF}$(\cdot)$ is the ``selective feature filter (Eq.~\ref{eq:Att})'', and Spot$_t$ is the obtained spotlight in the \emph{t}-th frame; $\digamma(\cdot)$ returns the center coordinates of its input, \emph{i.e.}, $\rm \{Lat_\emph{t}, Lon_\emph{t}\}=\digamma(\rm Spot_\emph{t})$, Lat and Lon respectively denotes the latitude and longitude; $\mathcal{P}_{\rm E2S}$ is a typical projection from ERP to sphere; $||\cdot||_{\rm S}$ measures the spherical distance between its inputs, which can be detailed as:
\begin{equation}
\begin{aligned}
\label{eq:distance}
&\hspace{-0.2cm}||\rm \{Lat_\emph{t},Lon_\emph{t}\},\{Lat_{\emph{t}+\emph{m}},Lon_{\emph{t}+\emph{m}}\}||_{\rm S}=\\[-4pt]
&\hspace{0.3cm}\arccos\!\Big(\sin({\rm Lat}_{t})\!\times\!\sin({\rm Lat}_{t\!+\!m})+\\[-6pt]
&\hspace{1cm}\!\cos({\rm Lat}_{t})\!\times\!\cos({\rm Lat}_{t\!+\!m})\!\times\!\cos({\rm Lon}_{t}\!-\!{\rm Lon}_{t\!+\!m})\Big),
\end{aligned}
\end{equation}
where we have provided a vivid demonstration regarding the computation of spherical distance, see Fig.~\ref{fig:SphereD}.

The output of Eq.~\ref{eq:weight}, $\omega$, can reflect the degree of how the 3rd attribute is satisfied --- a larger $\omega$ indicates the higher probability of Spot$_t$ to be a region containing shifted fixations.

\begin{figure}[!t]
\centering
\includegraphics[width=1\linewidth]{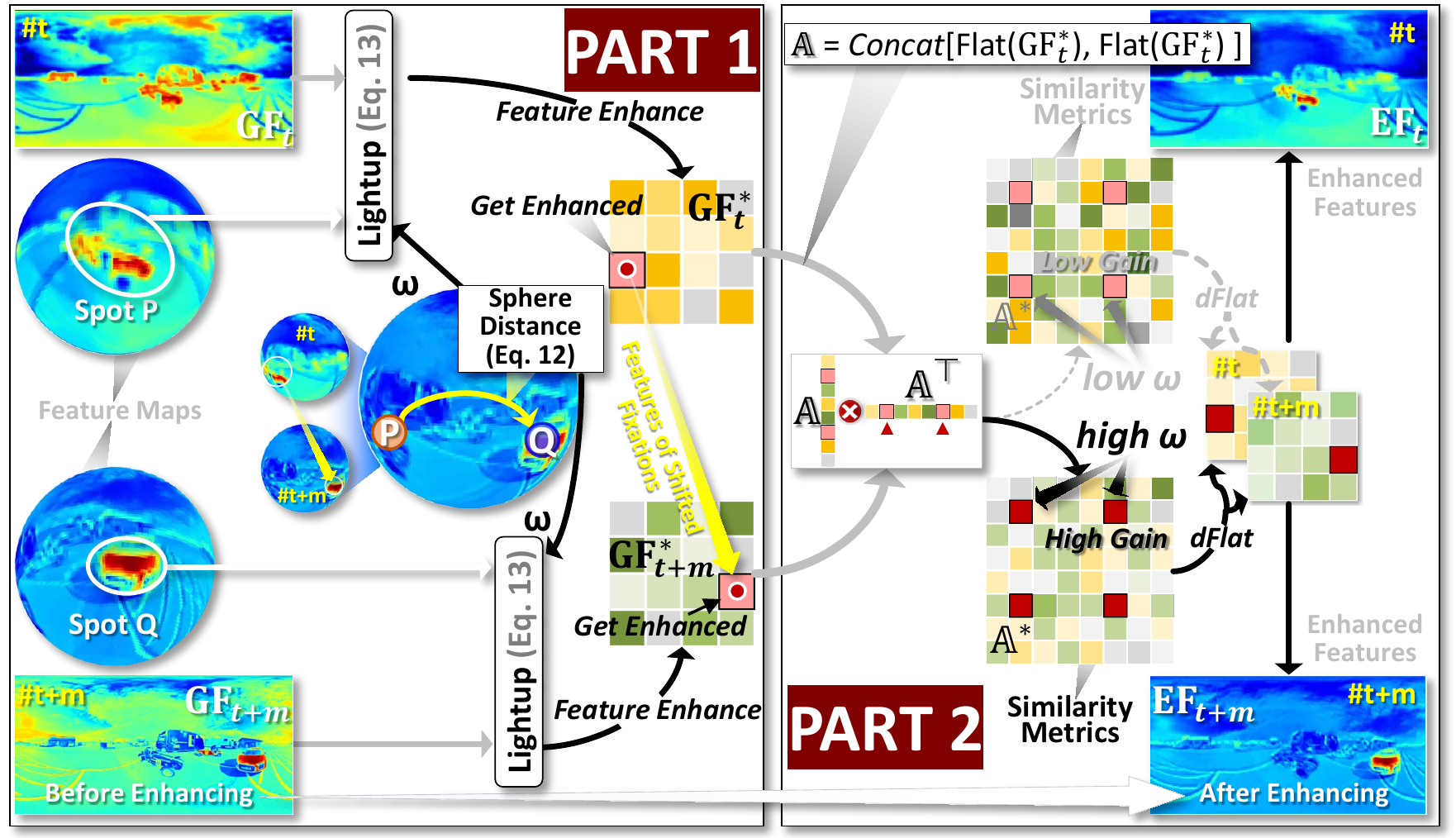}
\vspace{-23pt}
\caption{
Visualizing the ``shifting-aware feature enhancing'' consists of two sequential parts: \textbf{PART 1}, enhancing features associated with shifted fixations by increasing their feature values (\emph{i.e.}, \emph{Lightup}), and \textbf{PART 2}, ensuring these modified features facilitate the network's understanding of the actual ``process" of fixation shifting.
%, \emph{i.e.}, spotlight focus shifts from one spherical position to another rather than learning independently from them.
See section~\ref{sub:SAFE} for details.
}
\vspace{-12pt}
\label{fig:STAtt}
\end{figure}

\vspace{-11pt}
\subsubsection{Shifting-aware Feature Enhancing}
\label{sub:SAFE}
\vspace{-2pt}
In Fig.~\ref{fig:FishNet}, using the proposed ``selective feature filter'', we have already known which regions in the panoptic scene contain ``fixation shifting'', \emph{i.e.}, the Spot (Eq.~\ref{eq:Att}).
To let the network focus on these regions, we present the ``\noindent\textbf{\textcolor[RGB]{255,255,255}{\sethlcolor{lightblue}{\hl{\textbf{\rm H}}}}} shifting-aware feature enhancing'', whose key idea is to highlight those features correlating to the shifted fixations.
Our idea is somewhat tricky, which consists of two sequential parts: \textbf{1)} ``light up'' all features that possibly contain shifted fixations, then \textbf{2)} make these modified features trainable.

In \textbf{PART 1}, we simply use a very crude scheme to realize the ``light up'' process --- simply increase the feature values according to $\omega$ (Eq.~\ref{eq:weight}).
That is, given a larger $\omega$, the ``Spot's'' feature values shall be increased more.
The rationale of this process is very straightforward, the features that are crudely increased are, of course, more distinct from others, and thus, the network could pay more attention to them since they take a large portion of the overall training loss.
However, simply lighting up features as above has a critical problem, \emph{i.e.}, though the network already knows where the shifted fixations are, the network could still be ``fixation shifting-unaware''.
That is, from the network's perspective, the modified features are just the same to the normal ones except having some regions with large values, saying the networks can learn from these features, yet what they learned is completely independent of the ``fixation shifting process''.
So, what we shall let the network learn is the ``process'' of fixation shifting, \emph{i.e.}, spotlight focus shifts from one spherical position to another.
Thus, the primary objective of \textbf{PART 2} is to achieve this goal.
Next, we shall respectively detail parts 1) and 2), which have also been visualized in Fig.~\ref{fig:STAtt}.

The GF feature light up process (\emph{i.e.}, the \textbf{PART 1}) can be formulated as:
\begin{equation}
\begin{aligned}
\\[-20pt]
\rm GF^*\gets \emph{Lightup}(GF, Spot, \omega),
\\[-3pt]
\end{aligned}
\label{eq:ligu}
\end{equation}
where the function \emph{Lightup}$(\cdot)$ positively increases the feature values according to the spherical distance $\omega$ (Eq.~\ref{eq:weight}) by $\times (1+\omega)$, and the increased features are indicated by Spot (Eq.~\ref{eq:Att}).
This process has been visually demonstrated in PART 1 of Fig.~\ref{fig:STAtt}.
Then, the \textbf{PART 2} can be represented as:
\begin{equation}
\begin{aligned}
&\hspace{-0.27cm}{\rm \emph{dFlat}}(\underset{\Uparrow}{\underline{\mathbb{A}^*}}) \rightarrow \Big\{{\rm EF}_{t},{\rm EF}_{t\!+\!m}\Big\},\\[-4pt]
&\hspace{0.59cm}\mathbb{A}^* =  \underset{\Uparrow}{\mathbb{A}}\odot\sigma\big({\rm \emph{Softmax}}(\mathbb{A}\times \mathbb{A}^{\top})\times \mathbb{A}\big)\\[-4pt]
&\hspace{-1.18cm}\overbrace{\rm \mathbb{A} = \emph{Concat}\big(\emph{Flat}(GF^*_\emph{t}), \emph{Flat}(GF^*_{\emph{t}+\emph{m}})\big)}
\end{aligned}
\label{eq:dffff}
\end{equation}
where \emph{Flat}$(\cdot)$ flattens its input matrix to a single column, and \emph{dFlat}$(\cdot)$ divides its input back to two matrixes;
$\odot$ denotes element-wise matrix multiplicative operation;
\emph{Concat}$(\cdot)$ represents the concatenation;
${\sigma}(\cdot)$ denotes the sigmoid function;
$\mathbb{A}^{\top}$\!\! denotes the transpose of matrix $\mathbb{A}$; EF$_t$ denotes the enhanced features of the \emph{t}-th frame.
It is worth mentioning that the key of Eq.~\ref{eq:dffff} is the computation of $\mathbb{A}^*$, which follows the typical co-attention~\cite{vaswani2017attention} to convert the individual Spots of two different frames into ``a unified shifting process''.
The whole Eq.~\ref{eq:dffff} has been shown in the PART 2 of Fig.~\ref{fig:STAtt}.

\vspace{-6pt}
\subsection{Fixation Shifting Learning}
\label{Sub:FSL}
\vspace{-2pt}
In the above subsections, we have presented a brand-new network tailored for the ``fixation shifting'' problem, and the fixation shifting-aware features (\emph{i.e.}, EF) can be obtained via Eq.~\ref{eq:dffff}.
However, from the perspective of network training, using these fixation shifting-aware features alone cannot ensure good learning towards the fixation shifting phenomenon.
Here, we shall further explain this problem.

Generally, the overall training loss in video-related tasks (\emph{i.e.}, video saliency~\cite{cornia2018predicting}) usually covers multiple consecutive 15 frames; the ``fixation shifting'' is a sudden process that could be done in just two frames. In this case, the frames with fixation shifting (\emph{i.e.}, the two frames) only take a small part of the entire loss, which overlooks the ``fixation shifting'' process during training. % \emph{e.g.}, fifteen frames. However,

Therefore, we present the ``fixation shifting learning'', which targets focusing the loss backpropagation process on those frames with fixation shifting.
The key idea is in line with the ``shifting-aware feature enhancing''.
Instead of performing on features, our fixation shifting learning directly modifies the back propagated training loss, \emph{i.e.}, amplifies the losses of those frames with shifted fixations.
The whole process of ``fixation shifting learning'' has been shown at the top of Fig.~\ref{fig:FishNet}, and the loss function can be expressed as:
\begin{equation}
\begin{aligned}
\mathcal{L}oss =
&\sum_t{\mathcal{L}_{\rm KL}({\rm EF}_{t}, \underset{\Uparrow}{\underline{{\rm GT}_{t}^*}})}\!+\!\lambda\!\times\!\sum{\mathcal{L}_{\text{MSE}}\left({\omega}_{t},{\omega}^{\star}_{t}\right)},\\[-4pt]
&\hspace{-0.47cm}\overbrace{\rm GT^*\gets \emph{Lightup}(GT, \underset{\Uparrow}{\underline{Spot^*}}, \omega^{\star}),\ Eq.~\ref{eq:ligu}}\\[-4pt]
&\hspace{1.75cm}\overbrace{{\rm Spot^*} = MS\{\underset{\Uparrow}{\cdots}\}}\\[-4pt]
&\hspace{0.29cm}{\emph{Clustering}({\rm GT})}\rightarrow \Big\{{\overbrace{{\rm Fo}_{1},...,{\rm Fo}_\emph{u}}}\Big\}
\vspace{-2pt}
\end{aligned}
\label{eq:loss}
\end{equation}
where GT is the real fixations, $\mathcal{L}_{\rm KL}$ is the KL divergence loss~\cite{wang2018revisiting}, $\mathcal{L}_{\rm MSE}$ is the MSE loss~\cite{xu2021spherical}, $\rm EF_\emph{t}$ can be obtained via Eq.~\ref{eq:dffff}, ${\omega}$ and ${\omega}^{\star}$ denote the spherical distances respectively on the features and the GT;
\{Fo\} denotes the fixation clusters generated via \emph{Clustering}$\cdot$ (\emph{i.e.}, the classic DBSCAN~\cite{ester1996density}), and the function \emph{MS} select one fixation cluster from \{Fo\}, and the selected cluster shall have the largest number of fixations.

\vspace{-6pt}
\section{Experiments}
\vspace{-2pt}
This experimental section covers two aspects.
\textbf{Firstly}, we provide relevant details about the \ourApproach~approach and the \ourData~dataset.
We also conduct comparative experiments using fixation data collected with HMD to validate our method.
Moreover, we performed a user study to demonstrate the superiority of our \ourApproach.
We also provided the generic dataset analysis to demonstrate the field boost of our WinDB approach and \ourData~dataset.
\textbf{Secondly}, we present the benchmark results of our fixation shifting network, \OurNet, on the \ourData~dataset.
We evaluated its performance through quantitative and qualitative comparisons with SOTA methods.
Additionally, we conducted an ablation study on different components to verify the effectiveness.
We summarized key parameters mentioned in the previous technical details, shown in Table~\ref{tab:parameters}.

\begin{table}[!h]
\vspace{-10pt}
  \centering
    \caption{Details of key parameters used in our WinDB and FishNet.}
  \vspace{-12pt}
    \begin{tabular}{c}
    \begin{minipage}{1\linewidth}
     \hspace{-8pt}
      \includegraphics[width=\linewidth]{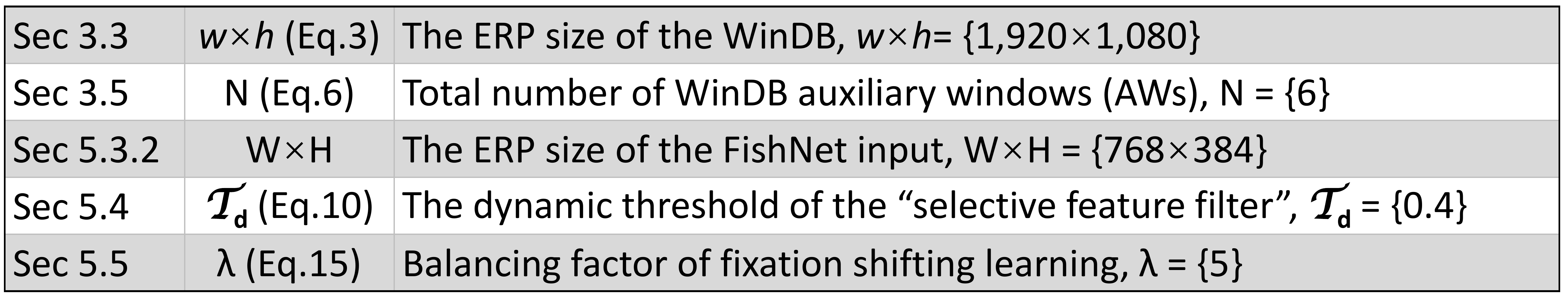}
    \end{minipage}
    \end{tabular}
\label{tab:parameters}
\vspace{-9pt}
\end{table}

\vspace{-9pt}
\subsection{Experiments of \ourApproach~and \ourData}
\vspace{-2pt}
\subsubsection{Platform and Hardware}
\vspace{-2pt}
Our WinDB uses Tobii Eye Tracker (v2) to collect panoptic fixations. To verify the effectiveness of our fixation collection method, we also use HMD to collect fixations as the references, where we use the HTC VIVE PRO EYE with 7invensun a-Glass eye tracker. We use a PC with an NVIDIA RTX 3090 GPU for the representative models' training and testing.

\begin{figure}[!t]
\centering
\includegraphics[width=0.80\linewidth]{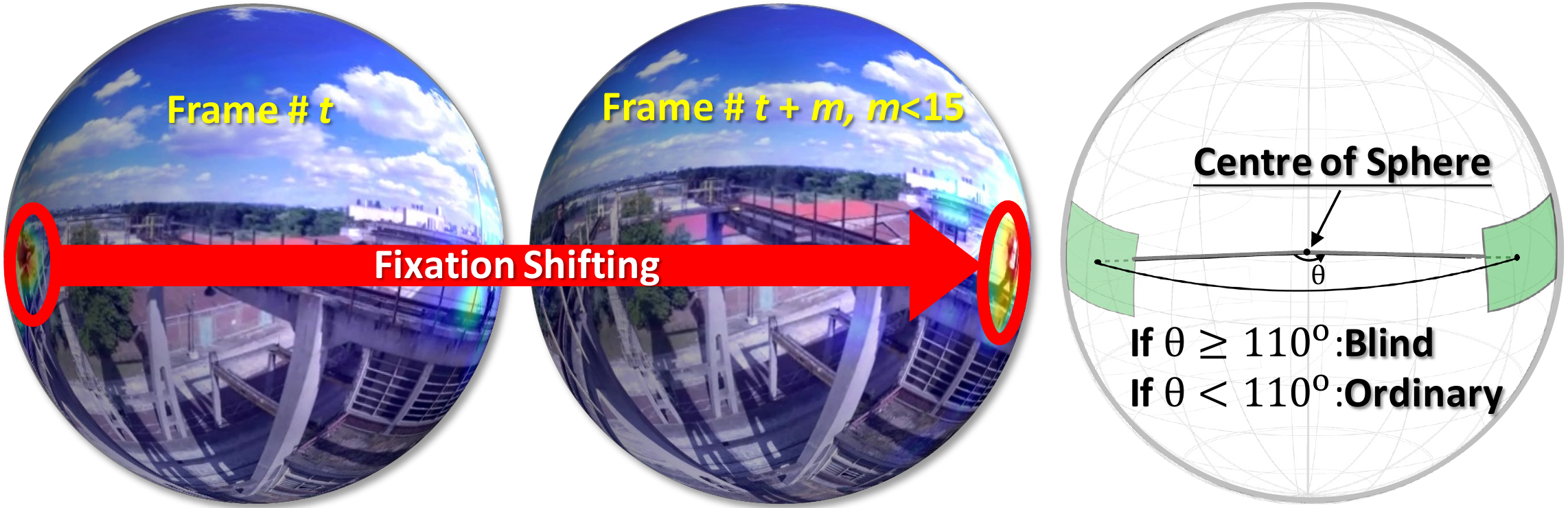}
\vspace{-13pt}
\caption{
Technical details of dividing our \ourData~into ``blind group'' and ``ordinary group''. See Sec.~\ref{Sub:Split} for details.
}
\vspace{-13pt}
\label{fig:setdivide}
\end{figure}

\vspace{-6pt}
\subsubsection{Dataset Split}
\vspace{-2pt}
\label{Sub:Split}
Since our data set contains ``blind zoom" and ``ordinary" scenes, for the convenience of subsequent experimental ``user study" and ``generic analysis", we divide the video clips in our \ourData~into two groups: \textbf{1)} clips with fixation shifting, also called the ``blind group'', and \textbf{2)} clips without fixation shifting, named as ``ordinary group''.
We measure the maximum fixation shifting distance for every 15 frames\footnote{Since the HVS's response limit is 500 ms, \emph{i.e.}, about 15 frames.} to determine if a given clip contains fixation shifting; for every 15 frames, we compute the spherical distance for any two frames. We can obtain a 15$\times$15 diagonal distance matrix for each clip. If the maximum fixation shifting distance is below a pre-defined threshold (we set this threshold to $\rm 110^o$ according to the maximum FOV of our HVS~\cite{hu2017deep,chou2018self}), we will regard this clip as a blind one
\footnote{We compute the spherical distance for any two frames in every 15 frames. We can obtain a 15$\times$15 diagonal spherical distance matrix for each clip, and the maximum element is considered. Each one in the blind zoom group MUST contain at least one shifting clip.};
otherwise, we label it as an ordinary one. This detailed process has been shown in Fig.~\ref{fig:setdivide}.
Thus, all 300 clips in our set can be divided into 195 ``blind zoom'' clips and 105 ``ordinary'' clips.

\begin{table}[!t]
%\vspace{-10pt}
  \centering
    \caption{Components quantitative evaluations in Fig.~\ref{fig:pipeline} to verify if all parts adopted in WinDB are effective. See Sec.~\ref{sub:Corr} for details.}
  \vspace{-12pt}
    \begin{tabular}{c}
    \begin{minipage}{1\linewidth}
    \hspace{-9pt}
      \includegraphics[width=\linewidth]{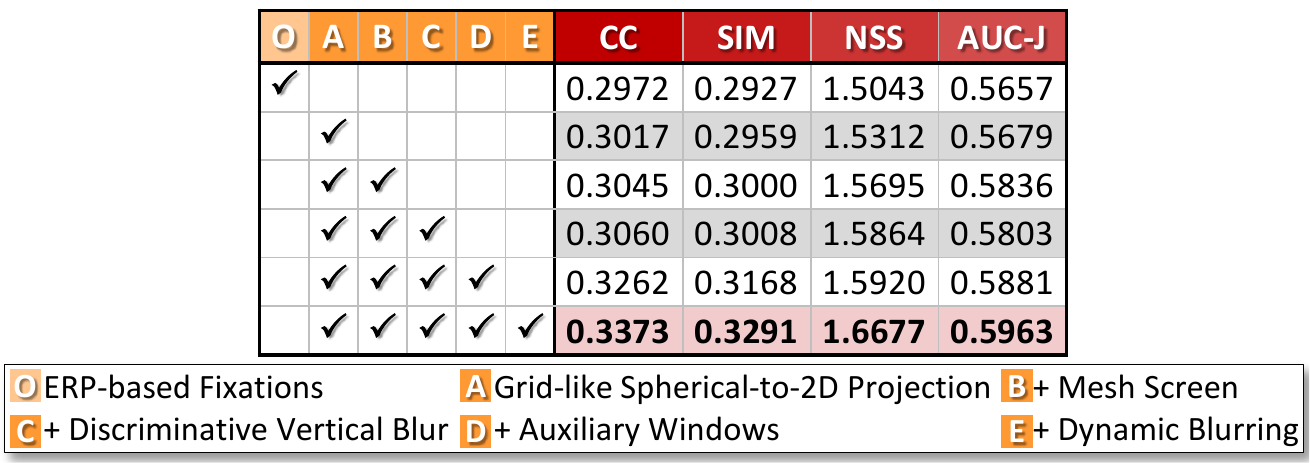}
    \end{minipage}
    \end{tabular}
\label{tab:pipeline}
\vspace{-14pt}
\end{table}

\vspace{-3pt}
\subsubsection{Correctness of Our \ourApproach~Approach}
\label{sub:Corr}
\vspace{-2pt}
The main advantage of our \ourApproach~collection approach against the classic HMD-based one is to handle the blind zoom limitation.
Thus, for those ordinary panoptic scenes (\emph{i.e.}, \emph{without} blind zooms), the fixations collected by our approach should stay consistent with those fixations collected by the HMD-based method, and this aspect can verify the soundness of our WinDB.

To demonstrate the effectiveness of each technical step used in our \ourApproach~(\emph{i.e.}, \textbf{\textcolor[RGB]{255,255,255}{\sethlcolor{red}{\hl{\textbf{\rm A}}}}} \!-\! \textbf{\textcolor[RGB]{255,255,255}{\sethlcolor{red}{\hl{\textbf{\rm E}}}}} in Fig.~\ref{fig:pipeline}), we conducted a component evaluation, where each component recruited no overlap of 10 users to provide fixations and was quantitatively tested by measuring four widely-used metrics (AUC-J, SIM, CC, NSS
~\cite{zhu2021viewing,dahou2021atsal,borji2019saliency,bylinskii2018different}).
The ERP-based and HMD-based fixations are newly collected here to serve as the references\footnote{Since the fixation shifting phenomenon is very rare in the existing HMD-based datasets, we have to perform HMD-based fixation collection in our new PanopticVideo-300 for the quantitative verification.}.
Notice that we have randomly selected ten clips from the ``ordinary'' group (\emph{i.e.}, all the panoptic scenes have no blind zoom) as a small validation set.

The quantitative results have been shown in Table~\ref{tab:pipeline}.
As can be seen, compared with the ERP-based fixations collection approach (marked by \textbf{\textcolor[RGB]{255,255,255}{\sethlcolor{O1yellow}{\hl{\textbf{\rm O}}}}}), our \ourApproach~method (marked by \textbf{\textcolor[RGB]{255,255,255}{\sethlcolor{A1yellow}{\hl{\textbf{\rm E}}}}}) can achieve significant performance improvement.
We can also notice that the overall performance can get promoted once a critical component has been applied, showing the necessity of each technical component.
\emph{B.t.w.}, we may notice that some quantitative numerics are not very high.
The main reason is that the HMD-based fixations are not the perfect ``ground truths'', \emph{i.e.}, the ``blind zoom group'' and ``ordinary group'' split is not perfect since it is just based on an empirical threshold (110$^{\circ}$), resulting limited numeric scores.

\vspace{-6pt}
\subsubsection{User Study}
\label{Sub:US}
\vspace{-2pt}
We have conducted a user study to further verify the overall quality of fixation maps collected by our \ourApproach~method against those collected by the HMD-based methods. The overall experiment setting is shown in Fig.~\ref{fig:subject}.
In this experiment, we have randomly selected 16 video clips from the ``blind group'' (Fig.~\ref{fig:setdivide}).
To conduct this user study, we recruited 30 users equally divided into three groups (see the groups A, B, and C in Fig.~\ref{fig:subject}).
Subject groups A and B are the experimental groups that provide HMD-based fixations and our \ourApproach-based fixations.
Then, we show the local salient views of each clip, in which salient views are automatically selected using the same scheme mentioned above, to the subject group C, and each subject in group C will provide an overall score ranging between 0$\sim$9 to each clip.
Notice that each clip will be shown to users in group C three times.
The 1st time is the plain ERP version to let the users become familiar with the overall contents in advance.
The 2nd and 3rd are randomly shown clips with salient views selected by either the HMD-based fixations or our \ourApproach~fixations.
Meanwhile, when showing clips (with all regions blurred except the salient view) to users in group C, we collect each subject's fixations since a higher quality of salient views should receive more fixations when users watch it.
Thus, we can obtain two indicators after this user study, \emph{i.e.}, \textbf{1)} the subjective quality scores, and \textbf{2)} the fixation point numbers in salient views.
Notice that a good view can attract more subject's fixations and receive a higher quality score.
These two indicators' results are shown in Fig.~\ref{fig:score}, where the right part is the subjective quality scores, and the left is the fixation point numbers.
As shown in the figure, our method can significantly outperform the HMD-based method in both indicators, where salient views determined by our method can receive more fixations and higher quality scores, verifying the superiority of our WinDB approach.

\begin{figure}[!t]
\centering
\vspace{-2pt}
\includegraphics[width=1\linewidth]{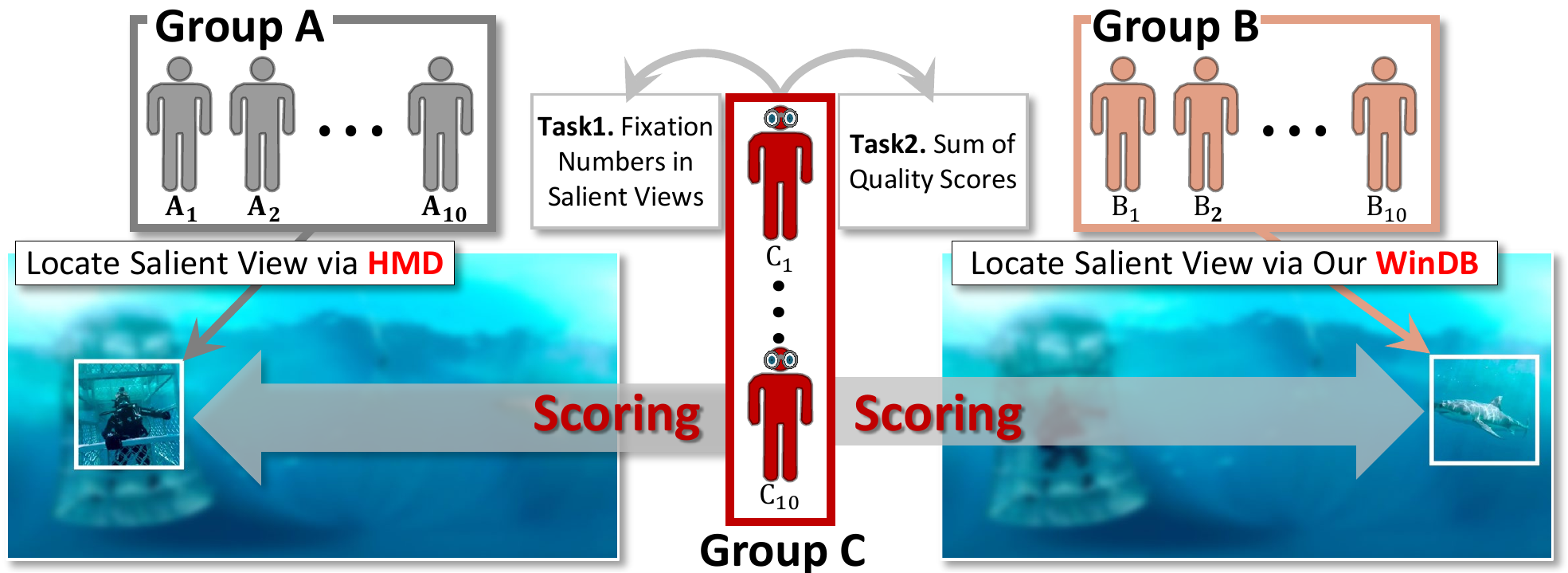}
\vspace{-23pt}
\caption{
Details of the subjective user study.
The user study is divided into three groups (\emph{i.e.}, A, B, and C), with ten users in each group.
Group A collects gaze data and locates salient views using HMD.
Group B collects gaze data and locates salient views using our WinDB.
Group C scores the salient view generated by Groups A and B (Task 1).
During the scoring process of Task 1, Group C's gaze data was collected without their informed consent (Task 2).
See Sec.~\ref{Sub:US} for details.
}
\label{fig:subject}
\vspace{-16pt}
\end{figure}

\begin{figure}[!b]
\centering
\vspace{-16pt}
\includegraphics[width=1\linewidth]{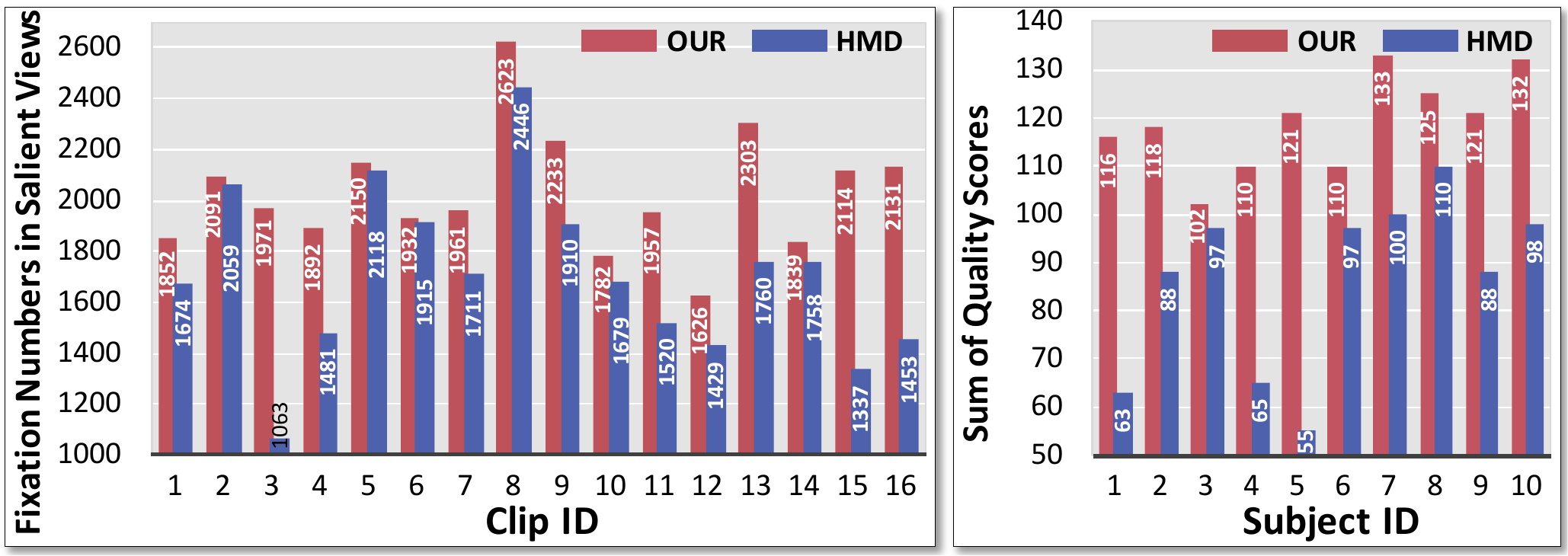}
\vspace{-24pt}
\caption{
User study results.
The left part shows the difference in fixation numbers in salient views selected by our \ourApproach~and the HMD-based approach.
The right part illustrates the sum of scores assigned by users after respectively experiencing our \ourApproach~approach and the HMD-based approach.
These two results suggest that our approach is more favorable than the HMD-based one.
See Sec.~\ref{Sub:US} for details.
}
\label{fig:score}
%\vspace{-16pt}
\end{figure}

\vspace{-5pt}
\subsubsection{Generic Analysis}
\label{sec:GA}
\vspace{-2pt}
This experiment targets to verify if fixations collected by our \ourApproach~approach can promote the existing panoptic fixation prediction models, where we choose the three most representative models (SpCNN~\cite{martin2020panoramic}, SalGAN~\cite{SalGAN}, and SalEMA~\cite{Linardos2019}).
Our rationales are two-fold.
\textbf{(1)} In ``ordinary'' panoptic scenes, if the fixations collected by our WinDB approach are adaptive to the fixations collected by the HMD-based method, a panoptic fixation prediction model trained on the HMD-based fixations only could get significant performance improvement once including our fixations into the training set.
\textbf{(2)} The fixations collected by our WinDB approach shall be able to let a panoptic fixation prediction model handle ``blind'' panoptic scenes well.
To verify these two aspects, we have conducted the experiments in Table~\ref{tab:Table1}.

We choose the existing HMD-based VR-EyeTracking dataset~\cite{xu2018gaze} as the baseline.
Then, we randomly select 50 clips with ``ordinary'' scenes as \textbf{A1} and randomly select 50 clips with ``blind'' scenes as \textbf{A2}.
Similarly, from our \ourData, we randomly select 50 clips with ``ordinary'' scenes as \textbf{B1} and 50 clips with ``blind'' scenes as \textbf{B2}.
Meanwhile, from our set, we randomly select 30 clips with ``ordinary'' scenes as \textbf{C1} and 30 clips with ``blind'' scenes as \textbf{C2}.
The formulated \textbf{A1}, \textbf{A2}, \textbf{B1}, and \textbf{B2} will serve as the training set to train the adopted three SOTA models; then these models will be tested on \textbf{C1} and \textbf{C2} respectively.
Notice that there exists no intersection between these splits.

\begin{table}[!t]
  \centering
    \caption{Quantitative evaluation to verify if \ourData~is adaptive to the existing HMD-based fixations~\cite{xu2018gaze}, and please details in corresponding paper context (\textbf{Generic Analysis}). The B1+B2 split denotes the combination of the ordinary group and the blind group; the corresponding tendency can be seen in Table~\ref{tab:CompareTable}. See Sec.~\ref{sec:GA} for details.}
  \vspace{-12pt}
    \begin{tabular}{c}
    \begin{minipage}{1\linewidth}
    \hspace{-8pt}
      \includegraphics[width=1\linewidth]{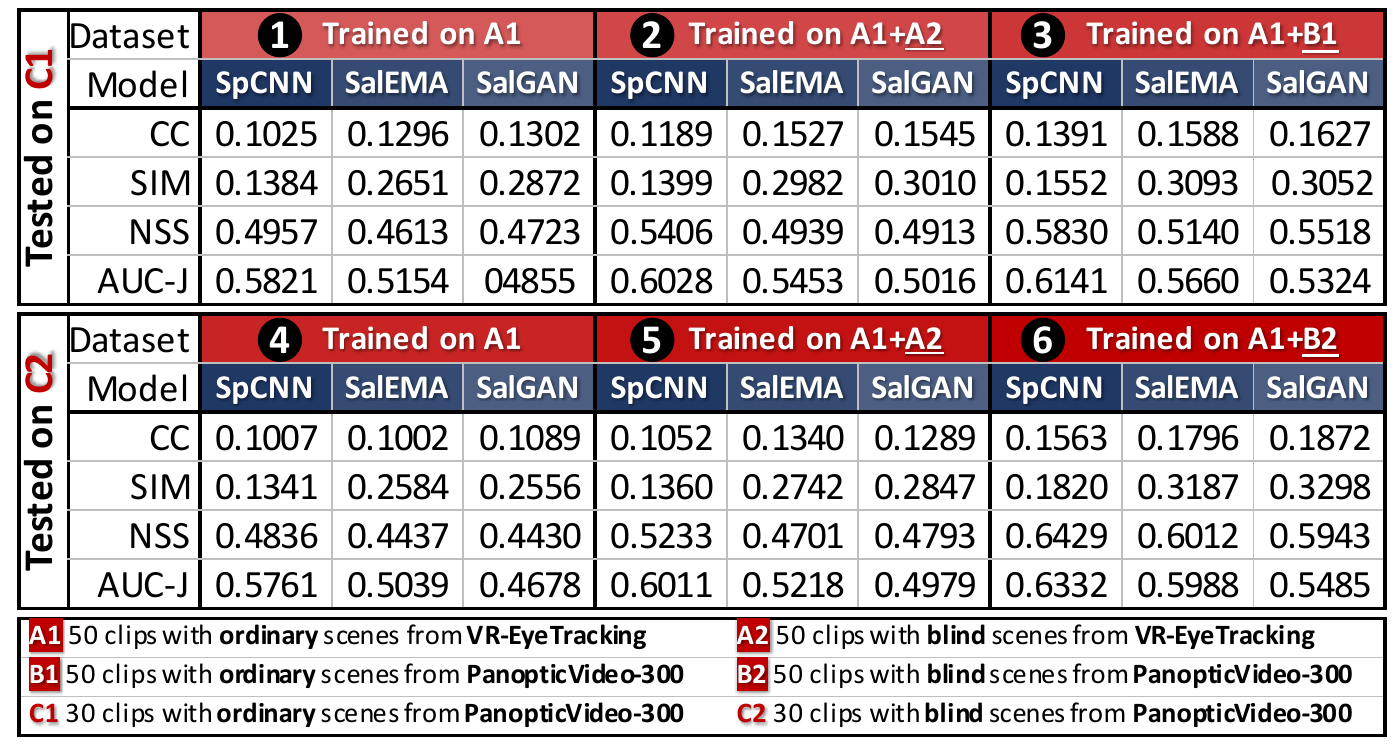}
    \end{minipage}
    \end{tabular}
\label{tab:Table1}
\vspace{-16pt}
\end{table}

By comparing mark {\large{\ding{202}}}
with mark {\large{\ding{205}}} in the table, we can easily notice that models trained on ``ordinary'' scenes could generally perform worse on ``blind'' scenes.
This is because these models have never learned the fixation shifting phenomenon when taking \textbf{A1} as the training set only.
Then, we have tested the three models by using \textbf{A1}+\textbf{A2} as the training set, and the testing results on \textbf{C1} and \textbf{C2} are still the same tendency (see mark {\large{\ding{202}}} \emph{vs.} {\large{\ding{203}}}, and {\large{\ding{205}}} \emph{vs.} {\large{\ding{206}}}), \emph{i.e.}, numerics in \textbf{C2} are generally worse than that in \textbf{C1}.
Also, as expected, we have noticed that the SOTA models' performances on both \textbf{C1} and \textbf{C2} have been promoted after using the additional training data.
The reason is that the initial 50 clips are far from enough for models to achieve good performances, and frames in \textbf{C2} do not completely belong to the fixation shifting case.
Finally, we have tested the models' performances by respectively adding \textbf{B1} and \textbf{B2} to the training set \textbf{A1}.
By comparing {\large{\ding{204}}} and {\large{\ding{207}}}, we can easily notice a significant performance improvement, implying that the added \textbf{B2} is very effective in enabling models to tackle the ``blind'' panoptic scenes.

%Also, as shown by the residuals of \{{\large{\ding{203}}}, {\large{\ding{204}}}\} and \{{\large{\ding{206}}}, {\large{\ding{207}}}\}, we can further confirm that the performance gain is exactly brought by the available of ``blind'' clips in the training set.
%For example, by increasing the training set size only, SalGAN's NSS metric can be improved from 0.4913$\rightarrow$0.5518 on the \textbf{C1} split. However, the performance gain could be significant when the increased data containing ``blind'' clips, \emph{i.e.}, the SalEMA's NSS metric has increased from 0.4701$\rightarrow$0.6012 on the \textbf{C2} split.

In summary, the experiment conducted above can ensure: 1) our \ourData~is adaptive to the existing HMD-based sets, and 2) models trained on set without containing any ``blind'' scenes cannot perform well in scenes with fixation shifting --- a very common phenomenon in real works. Thus, our \ourData~is a necessary complementary part to the existing ones.

\vspace{-11pt}
\subsection{Experiments of \OurNet}
\vspace{-2pt}
\subsubsection{Implementation details of FishNet}
\vspace{-2pt}
Our FishNet model uses Transformer~\cite{wang2021pyramid} as ``ERP local encoder'' (Sec.~\ref{sub:BAOP}), implemented with SGD optimizers in the PyTorch framework.
The learning rate was set to 1e-4, and a batch size of three was used for 11 epochs.
Each video frame is resized to a resolution of W$\times$H (768$\times$384) and divided into patches of size lon$\times$lat (15$^{\rm o}$$\times$15$^{\rm o}$).
During training, we construct each training instance consisting of two frames, \emph{e.g.}, the \emph{t}-th frame and the \{\emph{t}+\emph{m}\}-th frame in Fig.~\ref{fig:FishNet}, where \emph{m} random between \{1,2,...,15\}.
\begin{figure*}[!t]
\centering
\includegraphics[width=1\linewidth]{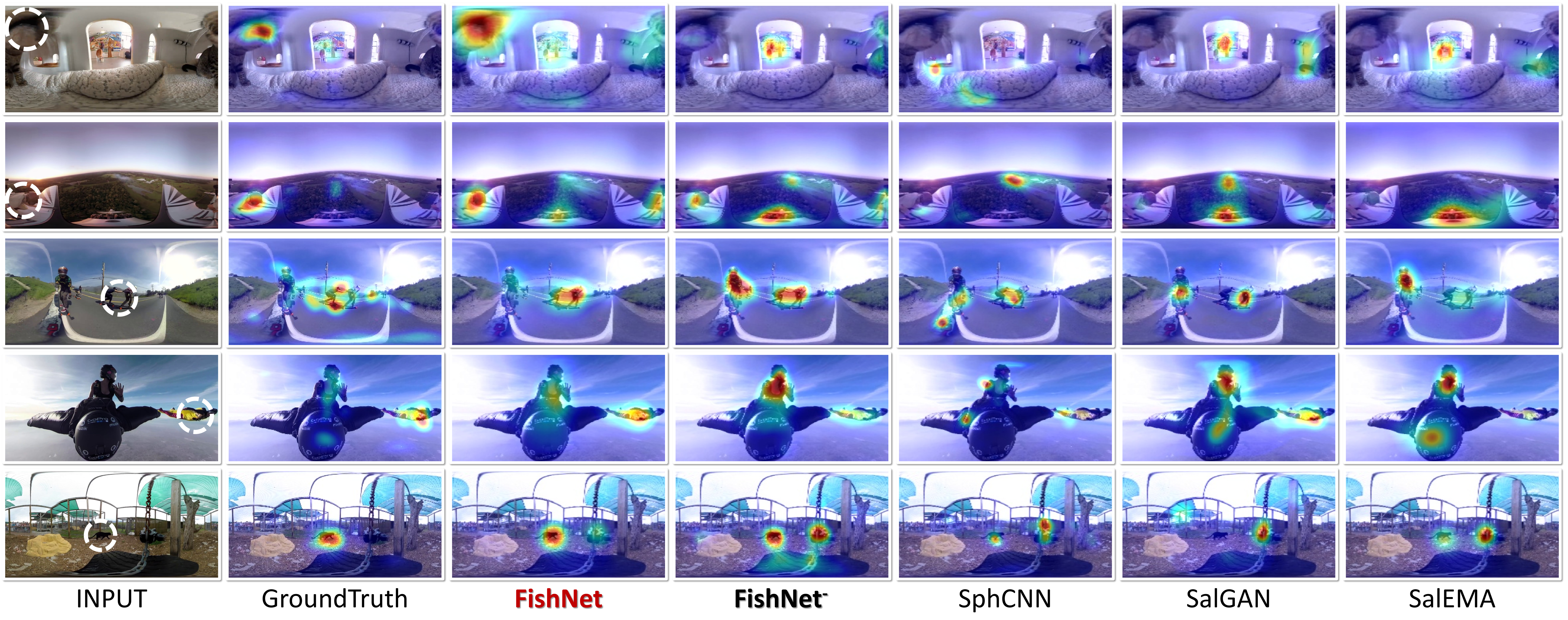}
\vspace{-24pt}
\caption{
Qualitative comparisons between our model (\emph{i.e.}, \textbf{FishNet} and \textbf{FishNet$^-$}) and SOTA models (\emph{i.e.}, SalEMA~\cite{Linardos2019}, SalGAN~\cite{SalGAN}, and SpCNN~\cite{martin2020panoramic}) on the \ourData. \textbf{FishNet$^-$} denotes a version of FishNet that includes only the ``panorama perception" component, excluding the ``deformable prober" and ``fixation shifting learning" components. All models were retrained using the training set of \ourData and tested on the corresponding testing set.
See Sec.~\ref{sub:QC} for details.
}
\label{fig:Compare}
\vspace{-15pt}
\end{figure*}

\begin{table}[!t]
  \centering
    \caption{Quantitative comparisons between FishNet and SOTA models on our set. +p/–p: with/without equatorial prior; ATsalAtt-I/V, ATsalImg/ATsalVideo: ATSal attention/expert model trained on image/video set~\cite{rai2017dataset,xu2018gaze}; \textbf{TE}, \textbf{RT}, \textbf{TR}: tested/retrained/trained the model on our set. `*' indicates that the method does not load a pre-trained model, while `\#' indicates that the method is a salient object detection method. See Sec.~\ref{sub:QC} for details.}
\vspace{-12pt}
    \begin{tabular}{c}
    \begin{minipage}{1\linewidth}
    \hspace{11pt}
      \includegraphics[width=0.83\linewidth]{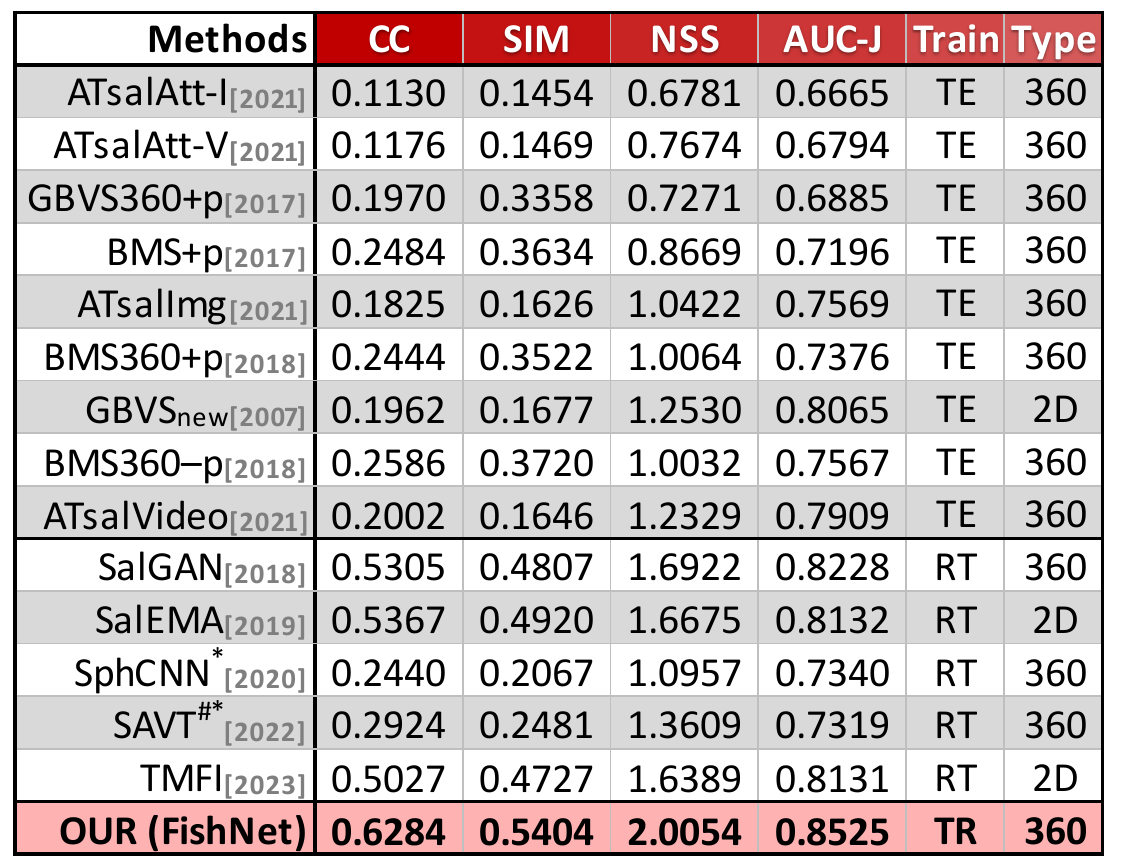}
    \end{minipage}
    \end{tabular}
\label{tab:CompareTable}
\vspace{-15pt}
\end{table}

\vspace{-6pt}
\subsubsection{Quantitative Comparisons}
Table~\ref{tab:CompareTable} presents our \OurNet~(see Fig.~\ref{fig:FishNet}) and some SOTA panoptic fixation prediction models' performances on our \ourData.
We have divided our set into training and testing sets, where the training set contains 240 clips and the rest are testing set\footnote{The detailed training and testing split can be found on our GitHub \url{https://github.com/guotaowang/PanopticVideo-300}.}.

The compared methods here include GBVS360, BMS360~\cite{lebreton2018gbvs360}, GBVS\footnote{\url{https://github.com/Pinoshino/gbvs}.}, ATsal~\cite{dahou2021atsal}, SalEMA~\cite{Linardos2019}, SalGAN~\cite{SalGAN}, SpCNN~\cite{martin2020panoramic}, SAVT~\cite{wu2022view}, and TMFI~\cite{zhou2023transformer}.
%Notice that,
Different from conventional 2D fixation prediction community or 2D salient object detection community, which tend to release their codes publicly, only two panoptic fixation prediction models' codes are made publicly trainable, \emph{i.e.}, SalGAN~\cite{SalGAN}, SpCNN~\cite{martin2020panoramic}.
Also, we have reached to the authors of SAVT~\cite{wu2022view}, and they helped us to retrain the model.
Others' codes are neither available nor retrainable.
To ensure a fair comparison, those models with available codes are all retrained on our training set, marked by ``RT or TR'' in the ``Train'' column.
As shown in Table~\ref{tab:CompareTable}, our FishNet model has achieved the best performance, indicating that the adopted ``deformable prober" and ``fixation shifting learning" is useful in handling the fixation shifting problem.
%\emph{W.r.t.} the relatively lower result in terms of SIM, we believe this phenomenon is quite normal since the compared SOTA models have also been re-trained on our training set.
Because our set can be adaptive to the existing HMD-based sets, which have been verified above (Sec.~\ref{sec:GA}), it is still possible for other 360 models to achieve good performance gain, demonstrating the generic advantage of our \ourData~again.

\vspace{-3pt}
\subsubsection{Qualitative Comparisons}
\label{sub:QC}
\vspace{-2pt}
Fig.\ref{fig:Compare} illustrates the visual results of \OurNet, \OurNet$^{-}$ (\emph{i.e.}, a version of FishNet that includes only the ``panorama perception" component, excluding the ``deformable prober" and ``fixation shifting learning" components), and three representative SOTA models (\emph{i.e.}, SalEMA~\cite{Linardos2019}, SalGAN~\cite{SalGAN}, SpCNN~\cite{martin2020panoramic}) on \ourData.
As shown in the figure, compared to the \OurNet$^{-}$ and other SOTA models, our proposed \OurNet~effectively focuses on objects that fixation shifted, providing evidence for the effectiveness of our ``deformable prober" (Sec.~\ref{sub:DFP}) and ``fixation shifting learning" (Sec.~\ref{Sub:FSL}) component of FishNet.
For example, in line 1 and line 3 of Fig.\ref{fig:Compare}, our \OurNet~can accurately focus on the sudden appearance of ``cat" and ``wingsuit flying", while other models without the capability of perceiving fixation shifting fail to change the fixations to the sudden events.
In addition, our \OurNet$^{-}$, with its powerful global perception and local distortion-free ability, can catch potential salient events in panoptic videos, \emph{i.e.}, the ``black leopard" in line 5 and the ``aircraft pilots" in line 2 of Fig.\ref{fig:Compare}.

\vspace{-9pt}
\subsection{Effectiveness Evaluation on Different Components}
\label{sub:EEDC}
\subsubsection{Effectiveness of the Panoptic Perception of \OurNet}
\vspace{-2pt}
To verify the effectiveness of the proposed ``panoptic perception" component of~\OurNet, we designed three different implementations, as detailed in Table~\ref{tab:component}.

\textbf{First}, \textbf{\textcolor[RGB]{255,255,255}{\sethlcolor{O1blue}{\hl{\textbf{\rm O}}}}} ``No pre-trained parameters" corresponds to the spherical convolution network~\cite{li2023spherical,su2021learning,xu2021spherical,zhang2018saliency}, where the convolution kernel is defined on the sphere to achieve distortion-free processing. However, due to the absence of pre-trained spherical convolution model parameters, loading pre-trained parameters is impossible. Consequently, for the sake of fairness in comparison, our proposed ``panoptic perception" component does not utilize pre-trained models and is labeled as \textbf{\textcolor[RGB]{255,255,255}{\sethlcolor{O1blue}{\hl{\textbf{\rm O}}}}}.

\textbf{Second}, \textbf{\textcolor[RGB]{255,255,255}{\sethlcolor{A1blue}{\hl{\textbf{\rm A}}}}} ``Transformer with local projection"~\cite{yun2022panoramic} incorporates CNN-based local projection after each transformer layer to handle the ERP's distortions.
Consequently, we made similar modifications and retrained our proposed ``panoptic perception" component accordingly.

\textbf{Third}, \textbf{\textcolor[RGB]{255,255,255}{\sethlcolor{A1blue}{\hl{\textbf{\rm B}}}}} ``Panoptic Perception" represents a completely transformer-independent approach that enables global panoptic perception and local distortion-free. This approach can effectively leverage the pre-trained transformer model~\cite{wang2021pyramid} in the 2D domain. To maintain consistency with the previous two implementations (\emph{i.e.}, \textbf{\textcolor[RGB]{255,255,255}{\sethlcolor{O1blue}{\hl{\textbf{\rm O}}}}} and \textbf{\textcolor[RGB]{255,255,255}{\sethlcolor{A1blue}{\hl{\textbf{\rm A}}}}}), we retained only the ``panoptic perception" component in FishNet (\emph{i.e.}, \textbf{\textcolor[RGB]{255,255,255}{\sethlcolor{A1blue}{\hl{\textbf{\rm B}}}}}). It is important to note that we re-trained on our training set and re-tested on our test set for all three implementations.

\textbf{1)} When comparing line 1 and line 3 in Table~\ref{tab:component},
we can observe that loading pre-trained parameters (\emph{i.e.}, \textbf{\textcolor[RGB]{255,255,255}{\sethlcolor{A1blue}{\hl{\textbf{\rm B}}}}} panoptic perception) leads to an approximate 3\%$\sim$10\% performance improvement compared to \textbf{\textcolor[RGB]{255,255,255}{\sethlcolor{O1blue}{\hl{\textbf{\rm O}}}}}.
This is because loading pre-trained weights can essentially be regarded as introducing additional training data, resulting in a more powerful semantic feature representation, \emph{e.g.}, the pre-trained Transformer~\cite{wang2021pyramid} was trained on the ImageNet dataset, which consists of 1.28 million training images from 1,000 categories with strong distinctive semantics.

\textbf{2)} By comparing line 1 with line 2 in Table~\ref{tab:component}, we can find that, by leveraging the pre-trained parameters and ``local projection ($\mathcal{P}_{\rm S2E}$)", \textbf{\textcolor[RGB]{255,255,255}{\sethlcolor{A1blue}{\hl{\textbf{\rm A}}}}} transformer with local projection achieves a significant performance improvement (+4\%) compared to \textbf{\textcolor[RGB]{255,255,255}{\sethlcolor{O1blue}{\hl{\textbf{\rm O}}}}}. However, when comparing line 2 and line 3, there is an 8\% gap in CC between \textbf{\textcolor[RGB]{255,255,255}{\sethlcolor{A1blue}{\hl{\textbf{\rm A}}}}} transformer-based local projection and \textbf{\textcolor[RGB]{255,255,255}{\sethlcolor{A1blue}{\hl{\textbf{\rm B}}}}} our proposed panoptic perception.
This is mainly because the adopted plain $\mathcal{P}_{\rm S2E}$ of \textbf{\textcolor[RGB]{255,255,255}{\sethlcolor{A1blue}{\hl{\textbf{\rm A}}}}} can alleviate the side effects induced by the ERP's visual distortions.
However, there is another clear side-effect, \emph{i.e.}, the pre-trained Transformer network parameters are not fully used.% not applicable anymore.

\textbf{3)} In line 3 in the Table~\ref{tab:component}, we observe a significant performance boost in \textbf{\textcolor[RGB]{255,255,255}{\sethlcolor{A1blue}{\hl{\textbf{\rm B}}}}} panoptic perception.
This improvement is attributed to its independence from the transformer, making it a universal plug-in. Thus, it can achieve global perception without any network modifications. Thus, the pre-trained Transformer network parameters~\cite{wang2021pyramid} can be fully used.

\vspace{-9pt}
\subsubsection{Effectiveness of Deformable Prober}
\label{sec:edpp}
\vspace{-2pt}
In Sec.~\ref{sub:DFP}, we proposed a ``deformable prober'' to enhance the FishNet's learning ability for fixation shifting. It can capture and enhance the shifted fixation feature and prevent the fixation shifting feature from being compressed as noise during training.
As shown in Fig.~\ref{fig:FishNet}, it contains two parts: \textbf{1)} ``selective feature filter'' and \textbf{2)} ``fixation shifting aware feature enhancement''.
%The former captures the fixation-shifted features, and the latter focuses on the model learning on the fixation-shifted features.
As shown in Fig.~\ref{fig:STAtt}, the ``fixation shifting aware feature enhancement'' also contains two sequential parts: PART 1 --- ``light up features that possibly contain shifted fixations by increasing their feature values", and PART 2 --- ``ensuring these modified features are trainable''.

\begin{table}[!t]
%\vspace{-10pt}
  \centering
    \caption{Quantitative evidence of component studies for \OurNet. (See Sec.~\ref{sub:EEDC})}
  \vspace{-12pt}
    \begin{tabular}{c}
    \begin{minipage}{1\linewidth}
    \hspace{-8pt}
      \includegraphics[width=\linewidth]{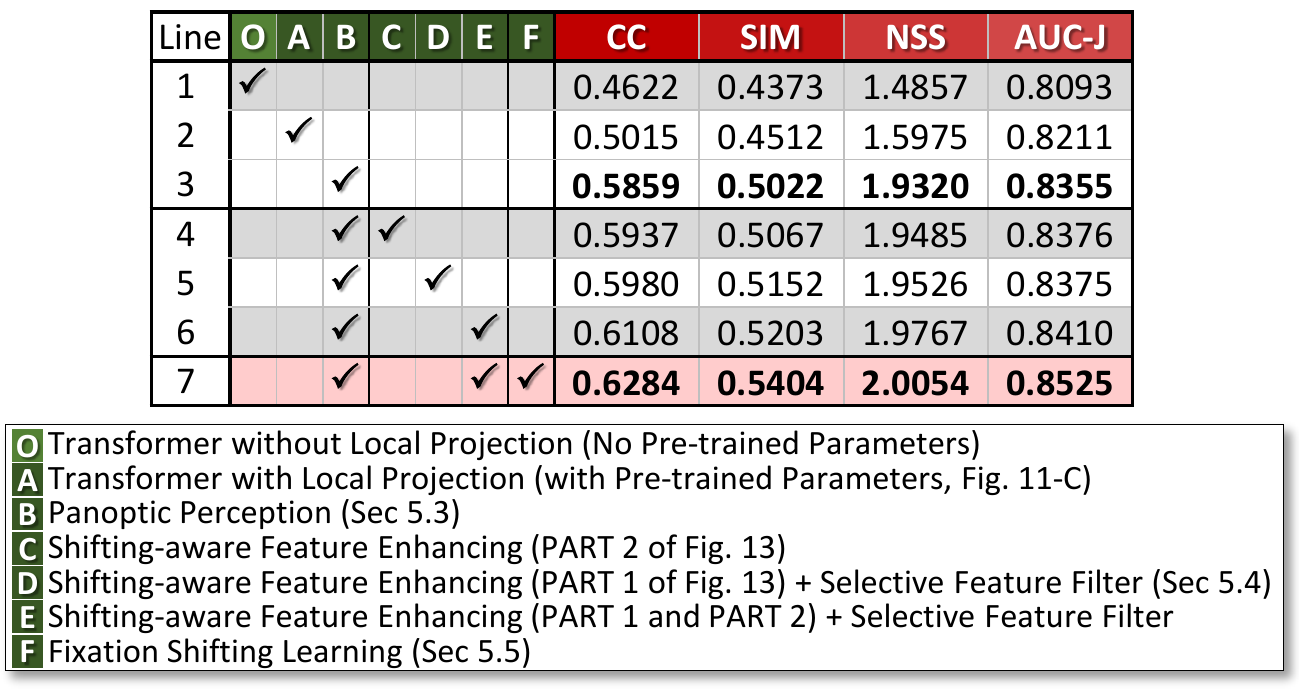}
    \end{minipage}
    \end{tabular}
\label{tab:component}
\vspace{-18pt}
\end{table}

As shown in Table~\ref{tab:component}, by comparing line 3 to line 4, we observe that \textbf{\textcolor[RGB]{255,255,255}{\sethlcolor{A1blue}{\hl{\textbf{\rm C}}}}} shifting-aware feature enhancing without \emph{Lightup} leads to an approximate 1\% improvement compared to \textbf{\textcolor[RGB]{255,255,255}{\sethlcolor{A1blue}{\hl{\textbf{\rm B}}}}} panoptic perception.
In other words, the \textbf{\textcolor[RGB]{255,255,255}{\sethlcolor{A1blue}{\hl{\textbf{\rm C}}}}} shifting-aware feature enhancing without \emph{Lightup} can also be used for fixation learning.
Although there is no specific component designed for perceptual fixation shifting, fixation learning can still be achieved with the assistance of the inter-frame similarity relation matrix, albeit with less ideal performance.

Furthermore, when comparing line 3 to line 5 in Table~\ref{tab:component}, we notice that our designed \textbf{\textcolor[RGB]{255,255,255}{\sethlcolor{A1blue}{\hl{\textbf{\rm D}}}}} selective feature filter with \emph{Lightup} yields an approximate 2\% performance improvement compared to \textbf{\textcolor[RGB]{255,255,255}{\sethlcolor{A1blue}{\hl{\textbf{\rm B}}}}} panoptic perception.
The primary reason for this improvement is that the introduction of \emph{Lightup} can enhance the feature response associated with fixation shifting, thereby enhancing the network's capability to perceive fixation shifting.
However, the enhanced feature remains isolated without establishing inter-frame correlation.

Finally, when comparing line 3 and line 6 in Table~\ref{tab:component}, we can find that our proposed \textbf{\textcolor[RGB]{255,255,255}{\sethlcolor{A1blue}{\hl{\textbf{\rm E}}}}} deformable prober (comprising selective feature filter and fixation shifting aware feature enhancement) can bring about a performance improvement of about 3\%$\sim$4\% compared to \textbf{\textcolor[RGB]{255,255,255}{\sethlcolor{A1blue}{\hl{\textbf{\rm B}}}}} panoptic perception.
The key reason behind this enhancement is that the deformable prober enables our network to perceive fixation shifting, enhances the fixation-shifted features, and establishes learnable inter-frame feature enhancement at the network level.
%Therefore, our proposed deformable prober can significantly enhance the perception and learning ability of the network for fixation shifting.

\begin{figure*}
    \begin{minipage}{\textwidth}
            \begin{minipage}[t]{0.25\textwidth}
        \centering\small
        \makeatletter\def\@captype{table}
        \renewcommand\table{6}
                        \caption{Ablation study on the FishNet's \protect\\ input size, \emph{i.e.}, the input \protect\\ ERP's size (Sec.~\ref{sub:BAOP}).}
            \vspace{-13pt}
            \resizebox{1.\linewidth}{!}{
            \begin{tabular}{c}
                \includegraphics[width=1\linewidth]{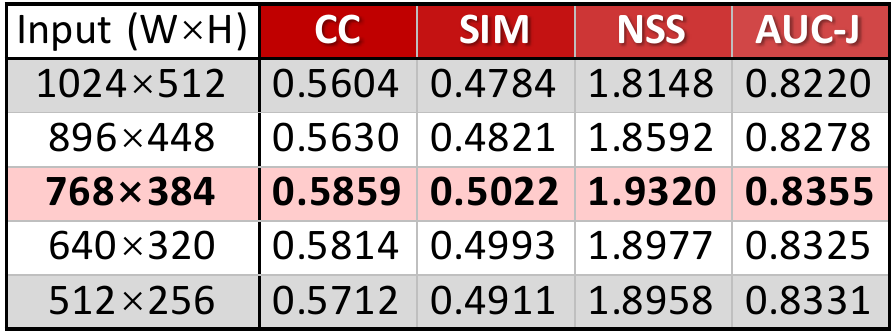}
            \end{tabular}
            }
            \label{tab:size}
        \end{minipage}
        \begin{minipage}[t]{0.25\textwidth}
        \centering\small
        \makeatletter\def\@captype{table}
        \renewcommand\table{7}
                        \caption{\hspace{0.3em}Ablation study on the FisNet's \protect\\ sub-patch size of \protect\\ ERP$^\star$
                        (Sec.~\ref{sub:BAOP}).}
            \vspace{-13pt}
            \resizebox{1.\linewidth}{!}{
            \begin{tabular}{c}
                \includegraphics[width=1\linewidth]{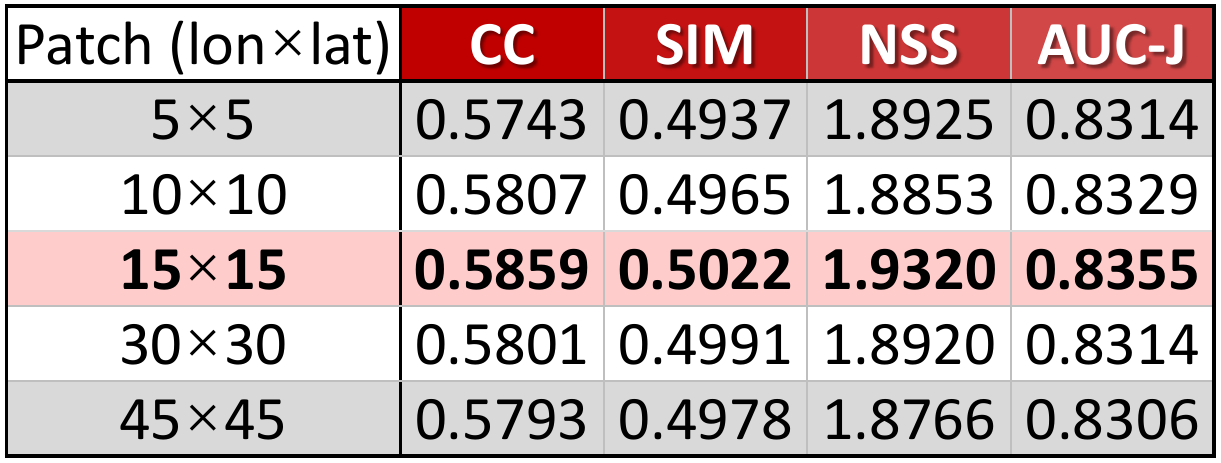}
            \end{tabular}
            }
            \label{tab:patchSize}
        \end{minipage}
        %\hspace{0.1ex}
        \begin{minipage}[t]{0.25\textwidth}
        \centering\small
        \makeatletter\def\@captype{table}
        \renewcommand\table{8}
                        \caption{Ablation study on the balancing factor used in fixation shifting learning (Sec.~\ref{Sub:FSL}).}
            \vspace{-13pt}
            \resizebox{1.\linewidth}{!}{
            \begin{tabular}{c}
                \includegraphics[width=1\linewidth]{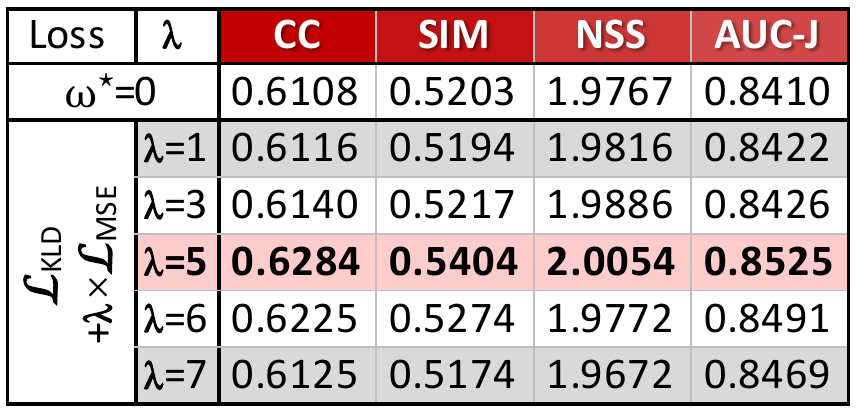}
            \end{tabular}
            }
            \label{tab:lambda}
        \end{minipage}
                \begin{minipage}[t]{0.23\textwidth}
        \centering\small
        \makeatletter\def\@captype{table}
        \renewcommand\table{9}
         \caption{Ablation study on $\mathcal{T}_d$ used in Technical Details of Selective Feature Filter (Sec.~\ref{Sub:TDSFF}).}
            \vspace{-13pt}
            \resizebox{1.\linewidth}{!}{
            \begin{tabular}{c}
                \includegraphics[width=1\linewidth]{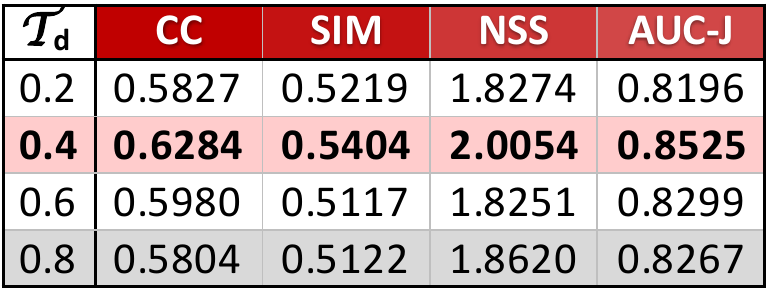}
            \end{tabular}
            }
            \label{tab:Thold}
        \end{minipage}
    \end{minipage}
    \vspace{-17pt}
\end{figure*}

\vspace{-6pt}
\subsubsection{Effectiveness of Fixation Shifting Learning}
In Sec.~\ref{Sub:FSL}, we presented the ``fixation shifting learning'', which aims to focus the network training process towards fixation shifting by emphasizing the loss associated with frames containing shifted fixations.

As demonstrated in the lines 6 and 7 of Table~\ref{tab:component}, we found that our proposed \textbf{\textcolor[RGB]{255,255,255}{\sethlcolor{A1blue}{\hl{\textbf{\rm F}}}}} ``fixation shifting learning'' leads to an approximate 2\% improvement compared to \textbf{\textcolor[RGB]{255,255,255}{\sethlcolor{A1blue}{\hl{\textbf{\rm E}}}}} deformable prober.
Furthermore, when comparing line 3 and line 7 of Table~\ref{tab:component}, we observe that our designed fixation shifting related component (\emph{i.e.}, deformable prober and fixation shifting learning) leads to an approximate 4\% performance improvement compared to the proposed panoptic perception component.
Since most video clips in the \ourData~contain fixation shifting scenes, achieving a substantial improvement depends on effectively addressing the primary challenge — fixation shifting.
Therefore, the substantial performance improvement reported in the table demonstrates the effectiveness of our fixation shifting learning.
This approach enables the FishNet network to focus better on fixation shifting learning during the training process.

\vspace{-6pt}
\subsection{Ablation Study}
\subsubsection{The Size of Panoptic Perception Input}
In Sec.~\ref{sub:BAOP}, we introduced the ``grid-like spherical-to-2D (Eq.~\ref{eq:mapping})" to address ERP distortion (see \textbf{\textcolor[RGB]{255,255,255}{\sethlcolor{lightblue}{\hl{\textbf{\rm D}}}}} in Fig.~\ref{fig:FishNet}) and convert ERP to ERP$^{\star}$, where the visual distortions can be solved and the global information has been well retained.

During this process, there are two ``sizes'' that could influence the overall network performance of FishNet, \emph{i.e.}, the FishNet's input ERP's size, and the FishNet's sub-patch size in ERP$^\star$.
So, here we perform two ablation studies to determine the optimal choices of these two sizes, and the quantitative results have been reported in Table~\ref{tab:size} and Table~\ref{tab:patchSize}.

As shown in Table~\ref{tab:size}, the performance improves (+1\%) when the input ERP width increases from 512 to 768. This is mainly because the larger ERP input can capture richer details and contextual information, which helps to understand the panoptic content. However, further increasing the input width from 768 to 1024 results in model performance degradation (-2\%). This is because larger network input implies additional information to be considered during the learning, complicating the learning process and degenerating the overall performance.

We also tested different patch sizes, and the corresponding results are reported in Table \ref{tab:patchSize}.
The table shows that the best choice is 15$^{\rm o}$$\times$15$^{\rm o}$.
Other choices could lead to some performance degradation.
The reason is that all other parameters are selected based on our default sub-patch size --- 15$^{\rm o}$$\times$15$^{\rm o}$.
Though using other choices could hurt the overall performance, the decreasing degree is marginal, further showing our approach's robustness.

\vspace{-6pt}
\subsubsection{Balancing Factor of Fixation Shifting Learning}
In Eq.~\ref{eq:loss}, we introduced a factor $\lambda$ to balance the impact of fixation shifting loss $\mathcal{L}_{\text{MSE}}$ and fixation prediction loss $\mathcal{L}_{\text{KLD}}$. Gradually increasing the value of $\lambda$ makes the FishNet pay more attention to the fixation shifting.
According to Table \ref{tab:lambda}, we observe an improvement (+2\%) in model performance when gradually increasing $\lambda$ from 1 to 5.
However, as $\lambda$ increases (5$\rightarrow$7), the performance degrades (-3\%).
This phenomenon is quite reasonable because a very large $\lambda$ could make the network over-emphasize learning the fixation shifting, and a small $\lambda$ could do the opposite. And the assign $\lambda=5$ exactly strikes the optimal trade-off.

\vspace{-3pt}
\subsubsection{$\mathcal{T}_d$ Used in Selective Fixation Filter}
In the Sec.~\ref{Sub:TDSFF} of ``selective feature filter", we utilize a dynamic threshold ($\mathcal{T}_d$) to extract ``Spot" (\emph{i.e.}, Eq. ~\ref{eq:Att}). The dynamic threshold is crucial for the quality of the Spot, because it directly determines whether the resulting Spot can represent the area where fixation shifting is most likely to occur in the current frame.
We have tested multiple choices of $\mathcal{T}_d$.
In Table~\ref{tab:Thold}, we observe that increasing $\mathcal{T}_d$ (\emph{i.e.}, 0.2$\rightarrow$0.4) improves the performance (+2\%). This is because the higher threshold accurately identifies the regions where fixation shifting is most likely to occur.
However, the performance degrades (-4\%) as $\mathcal{T}_d$ increases (\emph{i.e.}, 0.4$\rightarrow$0.8) since some regions containing shifted fixations could get filtered.

\vspace{-8pt}
\section{Limitations}
\label{sec:limition}
\vspace{-2pt}
Our WinDB approach effectively addresses the ``blind zoom" issue in HMD fixation data collection.
However, the human-designed ``dynamic auxiliary window" (Sec.~\ref{sub:ABT}) in WinDB may introduce minor biases to user attention.
For example, as shown in Fig.~\ref{fig:limitation}-(a), the salient event consistently remains within the yellow auxiliary window, often in a static state. As a result, the user tends to continuously fixate on the salient event, leading to the dynamic blur of the auxiliary window. Although the auxiliary window quickly becomes clear when it detects user interest, it may slightly affect fixation data collection.
Additionally, our FishNet ensures the perception of fixation shifting but focuses on a single salient event of each frame. In extreme cases, as shown in Fig.~\ref{fig:limitation}-(b), limited discrimination between multiple potential spotlights of each frame can lead to fixation shifting misjudgment. This is because various regions exhibit high feature responses; these regions can potentially be the ``spotlight", and minor changes between them can influence the generated ``spot" (Eq.~\ref{eq:Att}). In this case, FishNet may inaccurately perceive fixation shifts, limiting its performance in fixation shifting learning.

\begin{figure}[!t]
\centering
\includegraphics[width=1\linewidth]{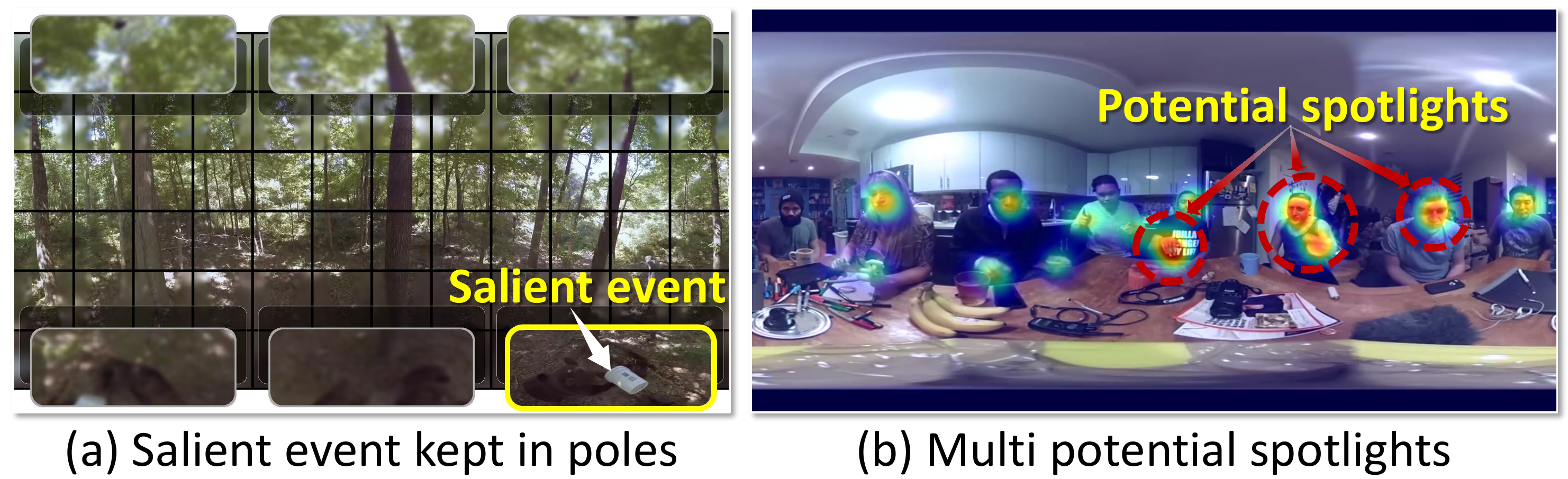}
\vspace{-23pt}
\caption{
The limitations of the proposed WinDB and FishNet, see Sec.~\ref{sec:limition} for more details.
}
\vspace{-10pt}
\label{fig:limitation}
\end{figure}

\vspace{-8pt}
\section{Conclusion}
\vspace{-2pt}
In this paper, we have made three distinguished innovations.
First, we have pointed out one critical limitation of the widely-used HMD-based panoptic fixation collection method, \emph{i.e.}~\cite{xu2018gaze,zhang2018saliency,xu2018predicting,zhang2022pav}, it suffers from ``blind zooms'', making the collected fixations not suitable in real works.
Thus, we have presented an HMD-free panoptic fixation collection approach named \textbf{\ourApproach}, which is more economical, comfortable, and, more importantly, technically correct.
Then, using our \ourApproach~approach,
we constructed a large set \textbf{\ourData}, containing 300 clips with over 225 semantic categories.
Compared to the existing set~\cite{xu2018gaze,cheng2018cube,zhang2018saliency}, our \ourData~contains 195 clips with fixation shifting.
The proposed \ourApproach~approach and the \ourData~have a large potential to promote the panoptic fixation prediction tool to a new era.
Finally, we have devised a simple yet effective \textbf{\OurNet}~model to handle the fixation shifting issue.
We verified the effectiveness and necessity of all technical components adopted in the paper through extensive objective and subjective experiments.

\bibliographystyle{IEEEtran}
\vspace{-8pt}
\bibliography{paper}{}

\begin{thebibliography}{10}
\providecommand{\url}[1]{#1}
\csname url@samestyle\endcsname
\providecommand{\newblock}{\relax}
\providecommand{\bibinfo}[2]{#2}
\providecommand{\BIBentrySTDinterwordspacing}{\spaceskip=0pt\relax}
\providecommand{\BIBentryALTinterwordstretchfactor}{4}
\providecommand{\BIBentryALTinterwordspacing}{\spaceskip=\fontdimen2\font plus
\BIBentryALTinterwordstretchfactor\fontdimen3\font minus
  \fontdimen4\font\relax}
\providecommand{\BIBforeignlanguage}[2]{{%
\expandafter\ifx\csname l@#1\endcsname\relax
\typeout{** WARNING: IEEEtran.bst: No hyphenation pattern has been}%
\typeout{** loaded for the language `#1'. Using the pattern for}%
\typeout{** the default language instead.}%
\else
\language=\csname l@#1\endcsname
\fi
#2}}
\providecommand{\BIBdecl}{\relax}
\BIBdecl

\bibitem{hu2017deep}
H.-N. Hu, Y.-C. Lin, M.-Y. Liu, H.-T. Cheng, Y.-J. Chang, and M.~Sun, ``Deep
  360 pilot: Learning a deep agent for piloting through 360 sports videos,'' in
  \emph{CVPR}, 2017.

\bibitem{sitzmann2018saliency}
V.~Sitzmann, A.~Serrano, A.~Pavel, M.~Agrawala, D.~Gutierrez, B.~Masia, and
  G.~Wetzstein, ``Saliency in vr: How do people explore virtual environments?''
  \emph{TVCG}, vol.~24, no.~4, pp. 1633--1642, 2018.

\bibitem{zhang2018saliency}
Z.~Zhang, Y.~Xu, J.~Yu, and S.~Gao, ``Saliency detection in 360 videos,'' in
  \emph{ECCV}, 2018, pp. 488--503.

\bibitem{zhang2022pav}
Y.~Zhang, F.-Y. Chao, W.~Hamidouche, and O.~Deforges, ``Pav-sod: A new task
  towards panoramic audiovisual saliency detection,'' \emph{TOMM}, 2022.

\bibitem{xu2018gaze}
Y.~Xu, Y.~Dong, J.~Wu, Z.~Sun, Z.~Shi, J.~Yu, and S.~Gao, ``Gaze prediction in
  dynamic 360 immersive videos,'' in \emph{CVPR}, 2018.

\bibitem{xu2018predicting}
M.~Xu, Y.~Song, J.~Wang, M.~Qiao, L.~Huo, and Z.~Wang, ``Predicting head
  movement in panoramic video: A deep reinforcement learning approach,''
  \emph{TPAMI}, 2018.

\bibitem{Aberman_2022_CVPR}
K.~Aberman, J.~He, Y.~Gandelsman, I.~Mosseri, D.~Jacobs, K.~Kohlhoff,
  Y.~Pritch, and M.~Rubinstein, ``Deep saliency prior for reducing visual
  distraction,'' in \emph{CVPR}, 2022.

\bibitem{Jiang_2022_CVPR}
L.~Jiang, Y.~Li, S.~Li, M.~Xu, S.~Lei, Y.~Guo, and B.~Huang, ``Does text
  attract attention on e-commerce images: A novel saliency prediction dataset
  and method,'' in \emph{CVPR}, 2022.

\bibitem{wang2021semantic}
G.~Wang, C.~Chen, D.-P. Fan, A.~Hao, and H.~Qin, ``From semantic categories to
  fixations: A novel weakly-supervised visual-auditory saliency detection
  approach,'' in \emph{CVPR}, 2021.

\bibitem{tsiami2020stavis}
A.~Tsiami, P.~Koutras, and P.~Maragos, ``Stavis: Spatio-temporal audiovisual
  saliency network,'' in \emph{CVPR}, 2020.

\bibitem{djilali2021rethinking}
Y.~Djilali, T.~Krishna, K.~McGuinness, and N.~OConnor, ``Rethinking 360deg
  image visual attention modelling with unsupervised learning.'' in
  \emph{ICCV}, 2021.

\bibitem{zhu2021viewing}
Y.~Zhu, G.~Zhai, Y.~Yang, H.~Duan, X.~Min, and X.~Yang, ``Viewing behavior
  supported visual saliency predictor for 360 degree videos,'' \emph{TCSVT},
  vol.~32, no.~7, pp. 4188--4201, 2021.

\bibitem{ECCV2022}
H.~Yun, S.~Lee, and G.~Kim, ``Panoramic vision transformer for saliency
  detection in 360 videos,'' in \emph{ECCV}, 2018.

\bibitem{nguyen2018your}
A.~Nguyen, Z.~Yan, and K.~Nahrstedt, ``Your attention is unique: Detecting
  360-degree video saliency in head-mounted display for head movement
  prediction,'' in \emph{ACM MM}, 2018.

\bibitem{zhu2019prediction}
Y.~Zhu, G.~Zhai, X.~Min, and J.~Zhou, ``The prediction of saliency map for head
  and eye movements in 360 degree images,'' \emph{TMM}, vol.~22, no.~9, pp.
  2331--2344, 2019.

\bibitem{cohen2018spherical}
T.~Cohen, M.~Geiger, J.~Kohler, and M.~Welling, ``Spherical cnns,''
  \emph{ICLR}, 2018.

\bibitem{jiang2019spherical}
C.~Jiang, J.~Huang, K.~Kashinath, P.~Marcus, and M.~Niessner, ``Spherical cnns
  on unstructured grids,'' \emph{ICLR}, 2019.

\bibitem{su2019kernel}
Y.-C. Su and K.~Grauman, ``Kernel transformer networks for compact spherical
  convolution,'' in \emph{CVPR}, 2019.

\bibitem{TCSVT1}
D.~Chen, C.~Qing, X.~Lin, M.~Ye, X.~Xu, and P.~Dickinson, ``Intra- and
  inter-reasoning graph convolutional network for saliency prediction on 360
  images,'' \emph{TCSVT}, 2022.

\bibitem{li2022spherical}
J.~Li, L.~Han, C.~Zhang, Q.~Li, and Z.~Liu, ``Spherical convolution empowered
  viewport prediction in 360 video multicast with limited fov feedback,''
  \emph{TOMM}, 2022.

\bibitem{lee2020spherephd}
Y.~Lee, J.~Jeong, J.~Yun, W.~Cho, and K.-J. Yoon, ``Spherephd: Applying cnns on
  360$^\circ$ images with non-euclidean spherical polyhedron representation,''
  \emph{TPAMI}, 2020.

\bibitem{esteves2018learning}
C.~Esteves, C.~Allen-Blanchette, A.~Makadia, and K.~Daniilidis, ``Learning so
  (3) equivariant representations with spherical cnns,'' in \emph{ECCV}, 2018.

\bibitem{weiler2018learning}
M.~Weiler, F.~A. Hamprecht, and M.~Storath, ``Learning steerable filters for
  rotation equivariant cnns,'' in \emph{CVPR}, 2018.

\bibitem{qiao2020viewport}
M.~Qiao, M.~Xu, Z.~Wang, and A.~Borji, ``Viewport-dependent saliency prediction
  in 360 video,'' \emph{TMM}, vol.~23, pp. 748--760, 2020.

\bibitem{su2017learning}
Y.-C. Su and K.~Grauman, ``Learning spherical convolution for fast features
  from 360 imagery,'' \emph{NeurIPS}, 2017.

\bibitem{lee2018memory}
S.~Lee, J.~Sung, Y.~Yu, and G.~Kim, ``A memory network approach for story-based
  temporal summarization of 360 videos,'' in \emph{CVPR}, 2018, pp. 1410--1419.

\bibitem{su2017making}
Y.-C. Su and K.~Grauman, ``Making 360 video watchable in 2d: Learning
  videography for click free viewing,'' in \emph{CVPR}, 2017.

\bibitem{xu2020viewport}
M.~Xu, L.~Jiang, C.~Li, Z.~Wang, and X.~Tao, ``Viewport-based cnn: A multi-task
  approach for assessing 360$^\circ$ video quality,'' \emph{TPAMI}, vol.~44,
  no.~4, pp. 2198--2215, 2020.

\bibitem{Tome_2019_ICCV}
D.~Tome, P.~Peluse, L.~Agapito, and H.~Badino, ``xr-egopose: Egocentric 3d
  human pose from an hmd camera,'' in \emph{ICCV}, 2019.

\bibitem{li2018bridge}
C.~Li, M.~Xu, X.~Du, and Z.~Wang, ``Bridge the gap between vqa and human
  behavior on omnidirectional video: A large-scale dataset and a deep learning
  model,'' in \emph{ACM MM}, 2018, pp. 932--940.

\bibitem{rondon2021track}
M.~F.~R. Rondon, L.~Sassatelli, R.~Aparicio-Pardo, and F.~Precioso, ``Track: A
  new method from a re-examination of deep architectures for head motion
  prediction in 360$^\circ$ videos,'' \emph{TPAMI}, vol.~44, no.~9, pp.
  5681--5699, 2021.

\bibitem{kim2018vrsa}
H.~Kim, H.-T. Lim, S.~Lee, and Y.~M. Ro, ``Vrsa net: Vr sickness assessment
  considering exceptional motion for 360 vr video,'' \emph{TIP}, vol.~28,
  no.~4, pp. 1646--1660, 2018.

\bibitem{Sun_2021_CVPR}
C.~Sun, M.~Sun, and H.-T. Chen, ``Hohonet: 360 indoor holistic understanding
  with latent horizontal features,'' in \emph{CVPR}, 2021.

\bibitem{Xu_2021_CVPR}
J.~Xu, J.~Zheng, Y.~Xu, R.~Tang, and S.~Gao, ``Layout-guided novel view
  synthesis from a single indoor panorama,'' in \emph{CVPR}, 2021.

\bibitem{yang2018automatic}
Y.~Yang, S.~Jin, R.~Liu, S.~B. Kang, and J.~Yu, ``Automatic 3d indoor scene
  modeling fr om single panorama,'' in \emph{CVPR}, 2018.

\bibitem{Zhang_2019_ICCV}
C.~Zhang, S.~Liwicki, W.~Smith, and R.~Cipolla, ``Orientation-aware semantic
  segmentation on icosahedron spheres,'' in \emph{ICCV}, 2019.

\bibitem{su2016pano2vid}
Y.-C. Su, D.~Jayaraman, and K.~Grauman, ``Pano2vid: Automatic cinematography
  for watching 360$^\circ$ videos,'' in \emph{ACCV}, 2016.

\bibitem{yu2018deep}
Y.~Yu, S.~Lee, J.~Na, J.~Kang, and G.~Kim, ``A deep ranking model for
  spatio-temporal highlight detection from a 360$^\circ$ video,'' in
  \emph{AAAI}, 2018.

\bibitem{zhuang2023spdet}
C.~Zhuang, Z.~Lu, Y.~Wang, J.~Xiao, and Y.~Wang, ``Spdet: Edge-aware
  self-supervised panoramic depth estimation transformer with spherical
  geometry,'' \emph{TPAMI}, 2023.

\bibitem{taneja2015geometric}
A.~Taneja, L.~Ballan, and M.~Pollefeys, ``Geometric change detection in urban
  environments using images,'' \emph{TPAMI}, vol.~37, no.~11, pp. 2193--2206,
  2015.

\bibitem{jin2020geometric}
L.~Jin, Y.~Xu, J.~Zheng, J.~Zhang, R.~Tang, S.~Xu, J.~Yu, and S.~Gao,
  ``Geometric structure based and regularized depth estimation from 360 indoor
  imagery,'' in \emph{CVPR}, 2020.

\bibitem{Pintore_2021_CVPR}
G.~Pintore, M.~Agus, E.~Almansa, J.~Schneider, and E.~Gobbetti, ``Slicenet:
  Deep dense depth estimation from a single indoor panorama using a slice-based
  representation,'' in \emph{CVPR}, 2021.

\bibitem{song20233d}
R.~Song, W.~Zhang, Y.~Zhao, Y.~Liu, and P.~L. Rosin, ``3d visual saliency: an
  independent perceptual measure or a derivative of 2d image saliency?''
  \emph{TPAMI}, 2023.

\bibitem{cheng2018cube}
H.-T. Cheng, C.-H. Chao, J.-D. Dong, H.-K. Wen, T.-L. Liu, and M.~Sun, ``Cube
  padding for weakly-supervised saliency prediction in 360 videos,'' in
  \emph{CVPR}, 2018.

\bibitem{xiong2018snap}
B.~Xiong and K.~Grauman, ``Snap angle prediction for 360 panoramas,'' in
  \emph{ECCV}, 2018.

\bibitem{ma2020stage}
G.~Ma, S.~Li, C.~Chen, A.~Hao, and H.~Qin, ``Stage-wise salient object
  detection in 360 omnidirectional image via object-level semantical saliency
  ranking,'' \emph{TVCG}, vol.~26, no.~12, pp. 3535--3545, 2020.

\bibitem{rana2019towards}
A.~Rana, C.~Ozcinar, and A.~Smolic, ``Towards generating ambisonics using
  audio-visual cue for virtual reality,'' in \emph{ICASSP}, 2019.

\bibitem{wang2022bifuse}
F.-E. Wang, Y.-H. Yeh, Y.-H. Tsai, W.-C. Chiu, and M.~Sun, ``Bifuse++:
  Self-supervised and efficient bi-projection fusion for 360 depth
  estimation,'' \emph{TPAMI}, vol.~45, no.~5, pp. 5448--5460, 2022.

\bibitem{dahou2021atsal}
Y.~Dahou, M.~Tliba, K.~McGuinness, and N.~OConnor, ``Atsal: An attention based
  architecture for saliency prediction in 360 videos,'' in \emph{ICPR}, 2021.

\bibitem{cong2023multi}
R.~Cong, K.~Huang, J.~Lei, Y.~Zhao, Q.~Huang, and S.~Kwong, ``Multi-projection
  fusion and refinement network for salient object detection in 360$^\circ$
  omnidirectional image,'' \emph{TNNLS}, 2023.

\bibitem{li2023spherical}
J.~Li, L.~Han, C.~Zhang, Q.~Li, and Z.~Liu, ``Spherical convolution empowered
  viewport prediction in 360 video multicast with limited fov feedback,''
  \emph{ACM TMCCA}, vol.~19, no.~1, pp. 1--23, 2023.

\bibitem{su2021learning}
Y.-C. Su and K.~Grauman, ``Learning spherical convolution for 360$^\circ$
  recognition,'' \emph{TPAMI}, vol.~44, no.~11, pp. 8371--8386, 2021.

\bibitem{xu2021spherical}
Y.~Xu, Z.~Zhang, and S.~Gao, ``Spherical dnns and their applications in 360
  images and videos,'' \emph{TPAMI}, vol.~44, no.~10, pp. 7235--7252, 2021.

\bibitem{yun2022panoramic}
H.~Yun, S.~Lee, and G.~Kim, ``Panoramic vision transformer for saliency
  detection in 360$^\circ$ videos,'' in \emph{ECCV}, 2022, pp. 422--439.

\bibitem{liu2021visual}
N.~Liu, N.~Zhang, K.~Wan, L.~Shao, and J.~Han, ``Visual saliency transformer,''
  in \emph{ICCV}, 2021.

\bibitem{wang2021pyramid}
W.~Wang, E.~Xie, X.~Li, D.-P. Fan, K.~Song, D.~Liang, T.~Lu, P.~Luo, and
  L.~Shao, ``Pyramid vision transformer: A versatile backbone for dense
  prediction without convolutions,'' in \emph{ICCV}, 2021, pp. 568--578.

\bibitem{dai2017deformable}
J.~Dai, H.~Qi, Y.~Xiong, Y.~Li, G.~Zhang, H.~Hu, and Y.~Wei, ``Deformable
  convolutional networks,'' in \emph{ICCV}, 2017, pp. 764--773.

\bibitem{anderson1978myth}
J.~Anderson and B.~Fisher, ``The myth of persistence of vision,'' \emph{JUFA},
  vol.~30, no.~4, pp. 3--8, 1978.

\bibitem{Yoon_2022_CVPR}
Y.~Yoon, I.~Chung, L.~Wang, and K.-J. Yoon, ``Spheresr: 360deg image
  super-resolution with arbitrary projection via continuous spherical image
  representation,'' in \emph{CVPR}, 2022.

\bibitem{AR}
O.~Younis, W.~Al-Nuaimy, F.~Rowe \emph{et~al.}, ``A hazard detection and
  tracking system for people with peripheral vision loss using smart glasses
  and augmented reality,'' \emph{IJACSA}, vol.~10, no.~2, 2019.

\bibitem{borst2015common}
A.~Borst and M.~Helmstaedter, ``Common circuit design in fly and mammalian
  motion vision,'' \emph{Nature neuroscience}, vol.~18, no.~8, pp. 1067--1076,
  2015.

\bibitem{yang2018object}
W.~Yang, Y.~Qian, J.-K. Kamarainen, F.~Cricri, and L.~Fan, ``Object detection
  in equirectangular panorama,'' in \emph{ICPR}, 2018, pp. 2190--2195.

\bibitem{gao2023thorough}
W.~Gao, S.~Fan, G.~Li, and W.~Lin, ``A thorough benchmark and a new model for
  light field saliency detection,'' \emph{TPAMI}, 2023.

\bibitem{liu2022poolnet+}
J.-J. Liu, Q.~Hou, Z.-A. Liu, and M.-M. Cheng, ``Poolnet+: Exploring the
  potential of pooling for salient object detection,'' \emph{TPAMI}, vol.~45,
  no.~1, pp. 887--904, 2022.

\bibitem{xia2022tensorized}
W.~Xia, Q.~Gao, Q.~Wang, X.~Gao, C.~Ding, and D.~Tao, ``Tensorized bipartite
  graph learning for multi-view clustering,'' \emph{TPAMI}, vol.~45, no.~4, pp.
  5187--5202, 2022.

\bibitem{vaswani2017attention}
A.~Vaswani, N.~Shazeer, N.~Parmar, J.~Uszkoreit, L.~Jones, A.~Gomez, L.~Kaiser,
  and I.~Polosukhin, ``Attention is all you need,'' \emph{NeurIPS}, 2017.

\bibitem{cornia2018predicting}
M.~Cornia, L.~Baraldi, G.~Serra, and R.~Cucchiara, ``Predicting human eye
  fixations via an lstm-based saliency attentive model,'' \emph{TIP}, vol.~27,
  no.~10, pp. 5142--5154, 2018.

\bibitem{wang2018revisiting}
W.~Wang, J.~Shen, F.~Guo, M.-M. Cheng, and A.~Borji, ``Revisiting video
  saliency: A large-scale benchmark and a new model,'' in \emph{CVPR}, 2018.

\bibitem{ester1996density}
M.~Ester, H.-P. Kriegel, J.~Sander, X.~Xu \emph{et~al.}, ``A density-based
  algorithm for discovering clusters in large spatial databases with noise,''
  in \emph{AAAI}, vol.~96, no.~34, 1996, pp. 226--231.

\bibitem{chou2018self}
S.-H. Chou, Y.-C. Chen, K.-H. Zeng, H.-N. Hu, J.~Fu, and M.~Sun, ``Self-view
  grounding given a narrated 360 video,'' in \emph{AAAI}, 2018.

\bibitem{borji2019saliency}
A.~Borji, ``Saliency prediction in the deep learning era: Successes and
  limitations,'' \emph{TPAMI}, 2019.

\bibitem{bylinskii2018different}
Z.~Bylinskii, T.~Judd, A.~Oliva, A.~Torralba, and F.~Durand, ``What do
  different evaluation metrics tell us about saliency models?'' \emph{TPAMI},
  vol.~41, no.~3, pp. 740--757, 2018.

\bibitem{martin2020panoramic}
D.~Martin, A.~Serrano, and B.~Masia, ``Panoramic convolutions for 360
  single-image saliency prediction,'' in \emph{CVPRW}, 2020.

\bibitem{SalGAN}
F.-Y. Chao, L.~Zhang, W.~Hamidouche, and O.~Deforges, ``Salgan360: Visual
  saliency prediction on 360 degree images with generative adversarial
  networks,'' in \emph{ICMEW}, 2018.

\bibitem{Linardos2019}
P.~Linardos, E.~Mohedano, J.~Nieto, K.~McGuinness, X.~GiroiNieto, and
  N.~OConnor, ``Simple vs complex temporal recurrences for video saliency
  prediction,'' in \emph{BMVC}, 2019.

\bibitem{rai2017dataset}
Y.~Rai, J.~Gutierrez, and P.~LeCallet, ``A dataset of head and eye movements
  for 360 degree images,'' in \emph{MMSys}, 2017.

\bibitem{lebreton2018gbvs360}
P.~Lebreton and A.~Raake, ``Gbvs360, bms360, prosal: Extending existing
  saliency prediction models from 2d to omnidirectional images,'' \emph{SP:IP},
  vol.~69, pp. 69--78, 2018.

\bibitem{wu2022view}
J.~Wu, C.~Xia, T.~Yu, and J.~Li, ``View-aware salient object detection for
  360$^\circ$ omnidirectional image,'' \emph{TMM}, pp. 1--15, 2022.

\bibitem{zhou2023transformer}
X.~Zhou, S.~Wu, R.~Shi, B.~Zheng, S.~Wang, H.~Yin, J.~Zhang, and C.~Yan,
  ``Transformer-based multi-scale feature integration network for video
  saliency prediction,'' \emph{TCSVT}, 2023.

\end{thebibliography}
\end{document}